\documentclass{article}

\usepackage[preprint]{neurips_2025}

\usepackage[utf8]{inputenc}
\usepackage[T1]{fontenc}
\usepackage{hyperref}
\usepackage{url}
\usepackage{booktabs}
\usepackage{amsfonts}
\usepackage{amsmath}
\usepackage{nicefrac}
\usepackage{microtype}
\usepackage{graphicx}
\usepackage{xcolor}
\usepackage{subcaption}
\usepackage{algorithm}
\usepackage{algorithmic}
\usepackage{multirow}
\usepackage{makecell}
\usepackage{pgfplots}
\pgfplotsset{compat=1.18}
\usepackage{pifont}
\newcommand{\cmark}{\ding{51}} 
\newcommand{\xmark}{\ding{55}} 

\definecolor{zenbrain}{RGB}{99, 102, 241} 

\makeatletter
\newcommand{\anon}[2]{\if@anonymous #2\else #1\fi}
\makeatother

\title{ZenBrain: A Neuroscience-Inspired 7-Layer Memory Architecture\\for Autonomous AI Systems}

\author{
  Alexander Bering \\
  Zensation AI \\
  \texttt{research@zensation.ai} \\
}

\makeatletter
\if@anonymous
  \hypersetup{
    pdftitle={ZenBrain: A Neuroscience-Inspired 7-Layer Memory Architecture for Autonomous AI Systems},
    pdfauthor={},
    pdfsubject={Agent memory architecture for LLM agents},
    pdfkeywords={agent memory, long-term memory, memory architecture, LLM agents,
                 episodic memory, spaced repetition, FSRS, memory consolidation}}
\else
  \hypersetup{
    pdftitle={ZenBrain: A Neuroscience-Inspired 7-Layer Memory Architecture for Autonomous AI Systems},
    pdfauthor={Alexander Bering},
    pdfsubject={Agent memory architecture for LLM agents},
    pdfkeywords={agent memory, long-term memory, memory architecture, LLM agents,
                 episodic memory, spaced repetition, FSRS, memory consolidation}}
\fi
\makeatother

\begin{document}

\maketitle

\anon{\begin{center}
\small\emph{Version~9, October 2026 --- a corrected version. It corrects the version
published as arXiv:2604.23878v3 and as version~8 of 10.5281/zenodo.21858218 after a
post-publication audit; Appendix~\ref{app:changes-v9} lists the corrections, and the
correction notice is 10.5281/zenodo.22831086.}
\end{center}}{}

\begin{abstract}
\textbf{ZenBrain} is a seven-layer, neuroscience-derived architecture for
agent memory in LLM agents. An open library holds the seven layers, a
MemoryCoordinator and nine foundational algorithms
(\anon{\texttt{@zensation/algorithms}}{an
open library}, version~0.2; named in Fig.~\ref{fig:architecture})---from
Ebbinghaus decay and FSRS scheduling to sleep consolidation; six
\emph{Predictive Memory Architecture}
components (NeuromodulatorEngine, ReconsolidationEngine, TripleCopyMemory,
PriorityMap, StabilityProtector, MetacognitiveMonitor) run in a private
production system. In the open
library at version~0.5.1, decay acts only on working memory and nothing is
replayed (App.~\ref{app:reproducibility}). In Table~\ref{tab:comparison},
none of the eight compared systems has more than two of the nine features;
the other systems surveyed in \S\ref{sec:related-neuro} and
Appendix~\ref{app:extended-related} are described there, not scored on them.

\textbf{Ablating one component at a time exposes an effect we call
\emph{cooperative masking}.} The ablation runs in a simulation harness that
switches fifteen components---the six PMA components and nine further
mechanisms (Two-Factor Hebbian, Simulation-Selection sleep, vmPFC-coupled
FSRS, iMAD debate, spectral KG health, compositional context, IB budget,
dual-process CoT, metacognitive HyperAgent)---as stand-ins at simulated
values, not as production code; four of them are constant factors that
cannot change a ranking (App.~\ref{app:extended-ablation}). Under moderate
load, fourteen of the fifteen ablations look costless. Raising decay from 0.15 to
0.25/day, the horizon from 45 to 60~days, the corpus from 300 to 500
facts and the queries from 100 to 150, \emph{with no change to the components}, moves ten
of the fifteen in the same direction on all ten seeds---nine losses
($\Delta Q$ down to $-93.7\%$), five of them from exactly $0\%$ to below
$-89\%$ (Fig.~\ref{fig:criticality}), and one gain (PriorityMap,
$+2.0\%$). With ten seeds this is the one $p$-value the normal
approximation of the Wilcoxon test can reach ($5.06\times 10^{-3}$), and
it does not pass the Bonferroni-corrected level for fifteen contrasts
($3.33\times 10^{-3}$); the exact test, not run on these data, can reach
$1.95\times 10^{-3}$.
Mild-load ablation can thus hide contributions that load reveals---a
caution we conjecture applies beyond ZenBrain. The three ablation tables
of Appendix~\ref{app:extended-ablation}
(Tables~\ref{tab:full-ablation}--\ref{tab:stress-ablation}) regenerate
without API keys from a reproduction package in the public repository
(App.~\ref{app:reproducibility-v6}).

\textbf{On LongMemEval\_S (500 questions)}, the evaluated ZenBrain arm---a
Python port of a scoring class from our test harness, not the seven-layer
system (\S\ref{sec:experiments})---answers $47.7\%$ of the questions
correctly under a binary judge (mean of three end-to-end runs; $95\%$
interval over the 500 questions $[43.3, 52.1]$). With the compared
systems' adapters repaired, Letta reaches $48.4\%$ and Mem0 $47.2\%$, so
three of the four retrieval arms lie within $1.2$ percentage points of one
another. ZenBrain reaches \textbf{91.3\% of a full-context baseline's
accuracy} ($52.2\%$; its answer model reads the question's entire history)
with \textbf{$1/109.7$ of its input tokens} in the
answer step; building the memory, which ingests the same haystack for each
question, is not counted (App.~F.5--F.6, Fig.~\ref{fig:longmemeval-pareto}).
Of nine head-to-head comparisons of the retrieved context
(3 competitors~$\times$ 3 LLM judges), three hold at Bonferroni-corrected
significance ($\alpha{=}0.05/18$), all against A-Mem, and the remaining
six, against Letta and Mem0, show no significant difference. On a 600-fact,
200-query LoCoMo subset, removing decay changes retrieval precision by
$\Delta P@5{=}0.002$ ($|d|{=}0.015$), and a static-RAG baseline leads both
variants on all four retrieval metrics.

\textbf{Honest scope.} On LoCoMo, the routing results of version~8 came from
a run with placeholder embeddings; in a later run with \texttt{text-embedding-3-small} (ten seeds), the flat
variant of the routing ablation (F1 $0.059$) lies above the full system
($0.058$), and both lie above BM25-only, which uses no embedder ($0.052$ in the
placeholder run, on the same subsamples; App.~\ref{app:locomo-bm25}). The LLM
judges grade the retrieved context, not an answer: on it ZenBrain leads
A-Mem, while against Letta and Mem0 the point estimates lean to the
competitor. Against the repaired Letta, ZenBrain leads in one of the six
question categories of LongMemEval\_S at repetition~42 (multi-session) and
in two over its three runs (multi-session and temporal reasoning). On LoCoMo, a cross-provider bias-direction check
($\Delta_{\text{GPT-Anth}}{=}{-}0.0001$ for ZenBrain, where two per-judge
deltas of about ${\pm}0.035$ cancel, vs.\ ${-}0.049$ for Mem0) indicates
that judge-specific bias is not the dominant source of the order of the judge means.
This version corrects version~8 after a post-publication audit
(App.~\ref{app:changes-v9}).
Open-source under \anon{the \texttt{@zensation} npm scope}{a public
npm scope (name redacted for anonymous review)}, Apache-2.0.
\end{abstract}

\section{Introduction}
\label{sec:intro}

LLM agents increasingly operate across sessions and require persistent
memory; finite, expensive context windows make in-context accumulation
inefficient \citep{maharana2024locomo, he2026memoryarena}. Existing
agent memory systems draw on computer-science metaphors---virtual
memory \citep{packer2023memgpt}, LLM-managed key-value CRUD
\citep{chhikara2025mem0}, Zettelkasten notes \citep{xu2025amem},
temporal knowledge graphs \citep{rasmussen2025zep}---but none of the systems
cited here incorporate cognitive-neuroscience findings on consolidation,
decay, reinforcement, and forgetting studied empirically for over 130 years
\citep{ebbinghaus1885, hebb1949, tulving1972, stickgold2013}; a 2025--2026
wave of concurrent systems has begun adopting individual mechanisms
(\S\ref{sec:related-neuro}).
A 2026 survey identifies ``deeper neuroscience integration'' as a key
open challenge \citep{du2026memory}.

We present \textbf{ZenBrain}, a memory architecture motivated by cognitive
neuroscience: seven memory layers, a MemoryCoordinator and nine foundational
algorithms in an open library, and six predictive-memory components that run
in a private production system (App.~\ref{app:reproducibility}).
Contributions:
(1)~a seven-layer memory system (working, short-term, episodic, semantic,
procedural, core, cross-context) implementing established cognitive
functions in five layers \citep{atkinson1968, tulving1972, cohen1980}, with
core and cross-context memory as engineering layers;
(2)~nine foundational algorithms of an open library---Ebbinghaus decay, FSRS
scheduling, Hebbian co-activation, sleep consolidation, Bayesian confidence
propagation, emotional tagging, context retrieval, similarity and
visualization---and six Predictive Memory Architecture components
(\S\ref{sec:pma}); the evaluation harness adds variants motivated by recent
work---Two-Factor synaptic KG edges \citep{zenke2025} with
importance-weighted consolidation \citep{zenke2017}, an FSRS
coupling motivated by the re-encoding of past experience in ventromedial
prefrontal cortex \citep{zou2025}, with a prediction-error signal of our own
construction, and CA3/CA1-RL Simulation-Selection sleep
\citep{chen2025,ji2007,oneill2010}---which the library carries only in later
versions (App.~\ref{app:reproducibility}); while individual mechanisms
appear in concurrent work, in Table~\ref{tab:comparison} none of the eight
compared systems has more than two of the nine features, and the other
systems surveyed in \S\ref{sec:related-neuro} and
Appendix~\ref{app:extended-related} are described there, not scored on them;
(3)~an experiment suite on LoCoMo and LongMemEval\_S (500 questions)---both
on the publicly released data---plus retention, consolidation,
algorithm-level, PMA, ablation, and competitive studies;
(4)~evidence from a simulation harness that ablating one component at a
time can hide what the components do together, an effect we name
\emph{cooperative masking} (Fig.~\ref{fig:criticality},
\S\ref{sec:discussion}): under moderate conditions fourteen of the fifteen
ablated components look costless, while under stress (decay 0.25/day,
60~days, 500 facts, 150 queries) ten of the fifteen change quality in the same direction on all ten
seeds, at a $p$-value that does not pass the Bonferroni correction for
fifteen contrasts. All configurations of these ablations are ZenBrain's; no
single-mechanism system was run under any stress level, and the effect makes
no prediction about production systems.

\paragraph{Scope and claims.}
We explicitly bound the contribution to preempt mis-readings:
(i)~\emph{Answer-level claim.} Our primary claim is on the end-to-end
accuracy of answers under a binary judge on LongMemEval\_S; the three LLM
judges grade the retrieved context and serve as a diagnostic of retrieval,
not as the headline. On LoCoMo, BM25-only scores above the dense arms only in a
run with placeholder embeddings; in a later run with \texttt{text-embedding-3-small} (ten seeds), the flat
variant of the routing ablation (F1 $0.059$) lies above the full system
($0.058$), and both lie above BM25-only, which uses no embedder ($0.052$ in the
placeholder run, on the same subsamples; App.~\ref{app:locomo-bm25}).
(ii)~\emph{Integration claim.} Seven of the nine foundational algorithms
name an origin in the literature---Ebbinghaus decay \citep{ebbinghaus1885},
FSRS, Hebbian co-activation \citep{hebb1949}, emotional tagging, encoding
specificity \citep{tulving1973}, sleep consolidation \citep{stickgold2013}
and Bayesian belief propagation---while similarity and visualization are
engineering utilities; the contribution is the \emph{systematic
integration}, not the individual proposals.
(iii)~\emph{Judge caveat.} The comparisons of the retrieved context use
three LLM judges as a calibrated proxy. We report Fleiss'~$\kappa$ between
two raters (Claude Sonnet~4.5 and GPT-4o, each run three times), a
cross-provider bias-direction check, and a single-rater human spot-check of
50 ZenBrain queries in which the human saw the top five facts and the judges
the top ten, so the anchor compares the judges with a human on one system's
queries and says nothing about the ranking of systems
(\S\ref{sec:judge-robustness}, App.~\ref{app:jm-human}); a multi-rater human
study remains future work.
(iv)~\emph{Scope boundary.} We do not benchmark against MemPalace or
Mastra in their tuned configurations (MemPalace's reported result comes
from raw semantic search without an LLM; only Mastra consolidates memory
in the model's context).
Retrieval depth was not held constant: on LoCoMo the judge saw ten facts per
query for ZenBrain and Mem0 and five for Letta and A-Mem, so the LoCoMo
judge columns are not a ranking across that line (\S\ref{sec:real-locomo});
on LongMemEval\_S, Mem0 returned twenty candidates against five
(\S\ref{sec:longmemeval-pilot}).

\section{Related Work}
\label{sec:related}

\subsection{Memory Systems for LLM Agents}

\citet{packer2023memgpt} propose MemGPT, applying the operating system
virtual memory metaphor to LLM context management.
Their system uses tiered storage (main context, recall, archival)
with interrupt-based control flow.
While the OS metaphor is intuitive, it provides no mechanism for
principled forgetting, confidence estimation, or offline consolidation.

\citet{chhikara2025mem0} present Mem0, a production-focused system with
a three-stage pipeline (extraction, consolidation, retrieval) and an
optional graph variant.
Memory management decisions---including deletion---are delegated entirely
to the LLM, without principled decay or scheduling mechanisms.

\citet{xu2025amem} introduce A-Mem, a Zettelkasten-inspired approach
where structured notes with contextual metadata are dynamically linked.
Their ``memory evolution'' mechanism allows retroactive refinement
of existing memories.
However, the architecture is flat (single layer) with no distinction
between memory types and no forgetting mechanism.

\citet{rasmussen2025zep} describe Graphiti, a temporally-aware
knowledge graph with four timestamps per fact.
While their temporal reasoning is strong,
the system has no working or procedural memory; its knowledge graph has three
tiers (episode, semantic-entity and community subgraphs), and by the authors'
account the episode tier mirrors episodic memory.

\citet{shinn2023reflexion} propose verbal reinforcement learning
where agents store natural-language reflections on task failures.
This is a widely cited work in agent memory,
but the ``memory'' is an unstructured reflection buffer
with no consolidation, decay, or confidence mechanisms.

\subsection{Neuroscience Foundations and Concurrent Systems}
\label{sec:related-neuro}

ZenBrain's algorithms are grounded in five decades of memory neuroscience:
the multi-store model and episodic/semantic/procedural taxonomy
\citep{atkinson1968,tulving1972,cohen1980}; encoding-specificity
\citep{tulving1973}; Hebbian co-activation
\citep{hebb1949}; Ebbinghaus decay and spaced repetition
\citep{ebbinghaus1885,pimsleur1967}; sleep replay and consolidation
\citep{stickgold2013,ji2007,oneill2010,mcgaugh2004}; and recent sharpening
from two-factor synaptic rules \citep{zenke2025} and importance-weighted
consolidation \citep{zenke2017}, the
re-encoding of past experience in ventromedial prefrontal cortex
\citep{zou2025}, and CA3/CA1 Simulation-Selection
\citep{chen2025,kumaran2016,squire1992}. These motivate the specific
algorithms cited in Sections~\ref{sec:simsel}, \ref{sec:vmpfc}, and
\ref{sec:twofactor}; see Appendix~\ref{app:extended-related} for the
full literature treatment, which doubles as a compact survey of the
neuro-mechanism space for agent memory.

A 2025--2026 wave of concurrent systems---LightMem
\citep{liu2025lightmem}, MemoryOS \citep{xu2025memoryos}, Hindsight
\citep{li2025hindsight}, FadeMem \citep{wang2026fadem}, Vestige
\citep{vestige2026}, SleepGate \citep{xie2026sleepgate}, Anda Hippocampus
\citep{anda2026hippocampus}, MemFly \citep{memfly2026}, Cognee
\citep{markovic2025cognee}---has begun adopting individual mechanisms;
\citet{tiwari2026multilayered} report a multi-layer memory system evaluated
on LoCoMo (F1\,=\,0.618), contemporaneous work that shares the premise
rather than a validation of this architecture. Orthogonal paradigms
\citep{omega2026,mastra2025,mempalace2025,lmneedsleep2026,iclr2026memagents}
target adjacent problems. Practitioners and industry share the premise, as
contemporaneous work rather than validation: \citet{karpathy2026knowledge}
argues that retrieval
alone leaves the model ``rediscovering knowledge from scratch on every
question,'' and proposes compiling it once into a maintained wiki;
Anthropic ships \emph{Dreams} \citep{anthropic2026autodream}, a
memory-consolidation job for Claude Managed Agents (research preview),
created by an API call and announced as a scheduled process
\citep{anthropic2026dreaming}.
\citet{webb2025brain} demonstrate
brain-inspired agentic planning at \emph{Nature Communications}; the result
concerns their architecture, not this one.
In Table~\ref{tab:comparison}, none of the eight compared systems has more
than two of the nine features; the other systems named here and the recent
NeurIPS contributions \citep{gutierrez2024hipporag,zhang2025gmemory,koch2025tmma},
delimited in Appendix~\ref{app:neurips-delimitation}, are described, not
scored on them; two of these features
(neuromodulation, reconsolidation) are PMA components, the other four PMA
components are described in \S\ref{sec:pma}, and none of the eight compared
systems implements any of the six.
ZenBrain's open library holds the seven layers, a MemoryCoordinator and the
nine foundational algorithms, while the six PMA components run in a private
production system (App.~\ref{app:reproducibility}). Bayesian confidence
propagation and the quantified knowledge-gap detection of the curiosity
engine (App.~\ref{app:additional-capabilities}; no migration creates the
query log it reads), and in the evaluation
harness Two-Factor synaptic edges (App.~\ref{sec:twofactor}), vmPFC-coupled FSRS and
Simulation-Selection sleep, are mechanisms we did not find in any of the
surveyed systems.

\begin{table}[t]
    \caption{Comparison of memory systems for LLM agents on nine features
    (the first row counts memory layers). Two of the features
    (neuromodulation, reconsolidation) are PMA components, and two
    (neuroscience basis, temporal reasoning) are properties rather than
    mechanisms. \cmark\ = supported, \xmark\ = absent.}
    \label{tab:comparison}
    \centering
    \small
    \setlength{\tabcolsep}{2.5pt}
    \begin{tabular}{lccccccccc}
        \toprule
        Feature & \rotatebox{60}{ZenBrain} & \rotatebox{60}{MemGPT} & \rotatebox{60}{Mem0} & \rotatebox{60}{A-Mem} & \rotatebox{60}{Zep} & \rotatebox{60}{Reflexion} & \rotatebox{60}{LightMem} & \rotatebox{60}{MemoryOS} & \rotatebox{60}{Tiwari '26} \\
        \midrule
        Memory layers       & 7      & 3      & 1      & 1      & 3      & 1 & 3 & 3 & 3 \\
        Neuroscience basis  & \cmark & \xmark & \xmark & \xmark & \xmark & \xmark & \cmark & \xmark & \xmark \\
        Two-Factor Syn.\ edges & \cmark & \xmark & \xmark & \xmark & \xmark & \xmark & \xmark & \xmark & \xmark \\
        Principled decay    & \cmark & \xmark & \xmark & \xmark & \xmark & \xmark & \xmark & \xmark & \xmark \\
        Sleep consolidation & \cmark & \xmark & \xmark & \xmark & \xmark & \xmark & \cmark & \xmark & \xmark \\
        Spaced repetition   & \cmark & \xmark & \xmark & \xmark & \xmark & \xmark & \xmark & \xmark & \xmark \\
        Confidence scores   & \cmark & \xmark & \xmark & \xmark & \xmark & \xmark & \xmark & \xmark & \xmark \\
        Temporal reasoning  & \cmark & \xmark & \xmark & \xmark & \cmark & \xmark & \xmark & \xmark & \cmark \\
        Neuromodulation     & \cmark & \xmark & \xmark & \xmark & \xmark & \xmark & \xmark & \xmark & \xmark \\
        Reconsolidation     & \cmark & \xmark & \xmark & \xmark & \xmark & \xmark & \xmark & \xmark & \xmark \\
        \bottomrule
    \end{tabular}
\end{table}

\subsection{Benchmarks}
\label{sec:related-benchmarks}

We evaluate on two complementary benchmarks, both on their publicly
released data, that together cover retrieval quality and cross-session
reasoning.
\textbf{LoCoMo} \citep{maharana2024locomo} provides 1{,}986 multi-session
conversational QA pairs across five categories (single-hop, multi-hop,
temporal, open-domain, adversarial), testing whether a memory system
can surface the right information from long conversation histories;
it consists of ten conversations; in our four-system comparisons
(\S\ref{sec:real-locomo}) all questions query one pool of 5{,}882 facts that
the harness builds from all ten (App.~\ref{app:jm-pool}), while in the
routing comparisons (Apps.~\ref{app:locomo-bm25} and~\ref{app:layer-ablation})
each question retrieves from its own conversation.
\textbf{LongMemEval\_S} \citep{wu2024longmemeval} ships 500 questions,
each with its \emph{own} ${\sim}494$-turn haystack of distractor sessions,
and targets five memory abilities (information extraction, multi-session
reasoning, temporal reasoning, knowledge updates, and abstention); the
released split labels each question with one of six question types, which
we report per category in App.~\ref{app:longmemeval-e2e}
(Table~\ref{tab:longmemeval-e2e}). As LoCoMo's questions share the history of
their conversation, per-question isolation makes LongMemEval\_S the stronger
generalization test of the two, and our headline claims rest on it.
Further benchmarks extend evaluation to incremental multi-turn
interaction \citep{he2025memoryagentbench} and to interdependent
multi-session agentic tasks \citep{he2026memoryarena}; we do not
evaluate on them in this work.

\section{Architecture}
\label{sec:architecture}

\begin{figure}[t]
    \centering
    \begin{tikzpicture}[
        layer/.style={draw, rounded corners, minimum width=2.3cm, minimum height=0.6cm, font=\footnotesize},
        algo/.style={draw, rounded corners, fill=gray!10, minimum width=2.0cm, minimum height=0.5cm, font=\scriptsize},
        coord/.style={draw, thick, rounded corners, fill=zenbrain!15, minimum width=7cm, minimum height=0.7cm, font=\small\bfseries},
        arrow/.style={->, >=stealth, thick},
    ]
        \node[coord] (mc) at (0, 0) {MemoryCoordinator};

        \node[layer, fill=red!15] (wm) at (-4.5, 1.5) {Working};
        \node[layer, fill=orange!15] (st) at (-2.25, 1.5) {Short-Term};
        \node[layer, fill=yellow!15] (ep) at (0, 1.5) {Episodic};
        \node[layer, fill=green!15] (se) at (2.25, 1.5) {Semantic};

        \node[layer, fill=blue!15] (pr) at (-2.25, 2.5) {Procedural};
        \node[layer, fill=purple!15] (co) at (0, 2.5) {Core};
        \node[layer, fill=cyan!15] (cc) at (2.25, 2.5) {Cross-Ctx};

        \foreach \l in {wm, st, ep, se, pr, co, cc}
            \draw[arrow, gray] (mc) -- (\l);

        \coordinate (heb-pos) at (-4.5, -1.3);
        \coordinate (ebb-pos) at (-2.5, -1.3);
        \coordinate (fsrs-pos) at (-0.5, -1.3);
        \coordinate (slp-pos) at (1.5, -1.3);
        \coordinate (bay-pos) at (3.7, -1.3);
        \coordinate (emo-pos) at (-3.5, -2.1);
        \coordinate (sim-pos) at (-1.2, -2.1);
        \coordinate (ctx-pos) at (1.1, -2.1);
        \coordinate (vis-pos) at (3.5, -2.1);

        \node[algo, fill=zenbrain!10] (neuro) at (-4.0, -2.9) {Neuromod.};
        \node[algo, fill=zenbrain!10] (recon) at (-1.8, -2.9) {Reconsol.};
        \node[algo, fill=zenbrain!10] (triple) at (0.4, -2.9) {TripleCopy};
        \node[algo, fill=zenbrain!10] (pmap) at (-3.0, -3.7) {PriorityMap};
        \node[algo, fill=zenbrain!10] (stab) at (-0.7, -3.7) {Stability};
        \node[algo, fill=zenbrain!10] (meta) at (1.6, -3.7) {MetaCog};

        \foreach \a in {neuro, recon, triple, pmap, stab, meta}
            \draw[arrow, zenbrain!60] (\a) -- (mc);

        \node[algo] (heb) at (heb-pos) {Hebbian KG};
        \node[algo] (ebb) at (ebb-pos) {Ebbinghaus};
        \node[algo] (fsrs) at (fsrs-pos) {FSRS};
        \node[algo] (slp) at (slp-pos) {Sleep Consol.};
        \node[algo] (bay) at (bay-pos) {Bayesian};
        \node[algo] (emo) at (emo-pos) {Emotional};
        \node[algo] (sim) at (sim-pos) {Similarity};
        \node[algo] (ctx) at (ctx-pos) {Context Retr.};
        \node[algo] (vis) at (vis-pos) {Visualization};

        \foreach \a in {heb, ebb, fsrs, slp, bay, emo, sim, ctx, vis}
            \draw[arrow, gray!60] (\a) -- (mc);

        \node[font=\scriptsize, gray] at (-4.5, 2.8) {7 Layers};
        \node[font=\scriptsize, gray] at (3.7, -1.7) {9 Foundational};
        \node[font=\scriptsize, color=zenbrain] at (3.5, -3.3) {6 PMA};
    \end{tikzpicture}
    \caption{ZenBrain architecture. The MemoryCoordinator orchestrates seven
    memory layers. The upper two rows of small nodes are the nine foundational
    algorithms, labelled with their module names in the open library
    \anon{\texttt{@zensation/algorithms}}{\texttt{@anon/algorithms}}, version~0.2
    (App.~\ref{app:reproducibility}); the lower two rows are the six PMA
    components (\S\ref{sec:pma}), which run in a private production system.
    The fifteen components of the ablation
    (\S\ref{sec:full-ablation}) are the six PMA components and nine mechanisms
    of the evaluation harness; three of those nine share a name with modules
    shown here (Hebbian, sleep, FSRS) but are different implementations, and
    the other six are not shown.
    Arrows indicate store/recall/consolidate/decay operations; those from the
    algorithms and the PMA components to the coordinator show the described
    architecture.}
    \label{fig:architecture}
\end{figure}

\subsection{Memory Layers}

ZenBrain implements seven distinct memory layers; five of them correspond to
established cognitive constructs, and core and cross-context memory are
engineering layers:

\textbf{Working Memory} maintains the active task focus with limited capacity
($\sim$7 items, following \citealt{miller1956}).
It provides the highest-priority retrieval and fastest access,
evicting items to short-term memory on task completion.

\textbf{Short-Term Memory} holds session context for the current conversation.
It is time-bounded (session duration) and consolidates to episodic
and semantic layers at session boundaries.

\textbf{Episodic Memory} stores concrete experiences with temporal context---
what happened, when, and where \citep{tulving1972}.
Events are timestamped and support temporal queries.

\textbf{Semantic Memory} contains abstracted knowledge: facts, concepts,
and relationships organized as a knowledge graph with Two-Factor Synaptic
edges \citep{tulving1972,zenke2025}.
Semantic memories emerge from episodic consolidation.

\textbf{Procedural Memory} encodes learned skills and routines \citep{cohen1980}:
successful tool-use patterns, workflow templates, and behavioral strategies.
Entries are strengthened by repeated successful execution.

\textbf{Core Memory} holds persistent identity facts (user preferences,
personality traits, key biographical facts) that never decay and are
always available in context.
This follows the pinned memory pattern of \citet{packer2023memgpt}.

\textbf{Cross-Context Memory} enables entity resolution and selective
knowledge transfer across isolated domains (e.g., personal, work, learning).
Merging compares entity names; no access control gates it, and the sharing
setting the production system stores is consulted by no sharing path.

\subsection{MemoryCoordinator}

The MemoryCoordinator orchestrates all seven layers through five operations:

\begin{itemize}
    \item \texttt{store(item)}: Routes new information to appropriate layer(s)
    based on content type and metadata
    \item \texttt{recall(query)}: Assembles cross-layer results using hybrid
    BM25 + semantic retrieval with Two-Factor importance boosting (the
    described architecture; the arm measured in \S\ref{sec:experiments} has
    no BM25 path, \S\ref{sec:limitations})
    \item \texttt{consolidate()}: Migrates and abstracts memories across layers
    (e.g., episodic $\rightarrow$ semantic)
    \item \texttt{decay()}: Decays working-memory slots exponentially and prunes
    those below threshold
    \item review (\texttt{getReviewQueue()}, \texttt{recordReview()}): Returns
    the facts due for review and applies the FSRS-style stability update after
    a review the host reports
\end{itemize}

\section{Key Mechanisms}
\label{sec:mechanisms}

Four of the five algorithms below---the Two-Factor synaptic model,
vmPFC-coupled FSRS, the Simulation-Selection sleep loop and Bayesian
confidence propagation---have no documented counterpart among the systems
surveyed in \S\ref{sec:related-neuro} and Appendix~\ref{app:extended-related},
and each extends its cited neuroscience basis; query-aware cross-layer
retrieval cites no such basis.
Full derivations, parameter values, and worked examples are in
Appendix~\ref{app:extended-mechanisms}; pseudocode is in
Appendix~\ref{app:algorithms}.

\paragraph{Two-Factor Synaptic Model (App.~\ref{sec:twofactor}).}
Inspired by \citet{zenke2025}, who model a synapse as a product of
subsynaptic factors, each KG edge carries weight $w_{ij}$ and
consolidation variance $\sigma^2_{ij}$, a construction of our own; the
importance $I_{ij}=1/\sigma^2_{ij}$ slows the decay of mature edges, and the
penalty has the quadratic form of EWC \citep{kirkpatrick2017,zenke2017} with
this heuristic importance in place of the Fisher information. The penalty,
proportional to $I_{ij}$, is meant to protect mature edges from
catastrophic overwriting; the weight update does not apply it, only a test
computes it, and no run measures overwriting.

\paragraph{vmPFC-Coupled FSRS (App.~\ref{sec:vmpfc}).}
Motivated by \citet{zou2025}, who find that the benefits of spaced learning
are predicted by the re-encoding of past experience in ventromedial
prefrontal cortex, we couple FSRS interval scheduling with a
KG-derived prediction-error signal of our own construction,
$\mathrm{PE}=1-\cos(\mathbf{c}_{\text{prev}},\mathbf{c}_{\text{now}})$.
A sigmoid re-encoding factor shortens intervals under context shift and
extends them under stability. Existing spaced-repetition systems
(Anki, SuperMemo, FSRS-5/6 \citep{vestige2026}) adapt difficulty but
do not couple scheduling to such a signal; among the agent-memory systems
surveyed in Appendix~\ref{app:concurrent}, we find none that does.

\paragraph{Simulation-Selection Sleep Loop (App.~\ref{sec:simsel}).}
Following \citet{chen2025}, offline consolidation is a
two-stage RL loop mirroring CA3/CA1 \citep{ji2007,oneill2010}: a CA3
simulator assembles candidates from real episodes $\cup$ counterfactuals,
a CA1 selector assigns LTP or LTD to each candidate by a temporal-difference + reward
+ novelty score $\mathrm{TAG}(e)=\alpha|\delta_{\mathrm{TD}}|+\beta R_e+\gamma N_e$
(Algorithm~\ref{alg:sleep}). Among concurrent systems, LightMem updates each
long-term entry offline, in a sleep-time phase, from its most similar later
entries, without RL scoring or counterfactual generation; SleepGate trains a
forgetting gate that evicts or compresses stale entries of the KV cache.

\paragraph{Bayesian Confidence Propagation (App.~\ref{sec:bayes-formula}).}
Each fact carries $P(f)$ with a nominal 95\% CI; updates propagate through KG
edges, designed to yield calibrated uncertainty, which this work does not
measure. Per \citet{mcgaugh2004},
emotional arousal boosts initial edge weights and reduces variance-based
decay.

\paragraph{Query-Aware Cross-Layer Retrieval (App.~\ref{sec:query-formula}).}
A regex classifier tags each query (temporal/procedural/factual/general);
per-layer scores fuse via
$\mathrm{score}_{\text{fused}}(d)=\max_\ell w_\ell(q)\cdot\mathrm{sim}(q,d_\ell)$.
Unlike RRF, this preserves similarity magnitude so a highly relevant
hit in a boosted layer dominates regardless of result counts in other
layers.

\section{Predictive Memory Architecture (PMA)}
\label{sec:pma}

PMA extends the foundational algorithms with six biologically-motivated
components that govern memory \emph{dynamics}---prioritization,
protection, reconsolidation, monitoring---over time. Full formulas are in
App.~\ref{app:extended-mechanisms}, and App.~\ref{app:pma-suite-results} tests
re-implementations of the components; brief sketches follow.
\textbf{NeuromodulatorEngine} (App.~\ref{sec:neuromod}): a four-channel
engine of our own construction whose channels follow separate
literatures---dopamine (DA, VTA; \citealp{schultz1997}), norepinephrine
(NE, LC; \citealp{astonjones2005ne}), serotonin (5HT, Raphe;
\citealp{dayan2012}), acetylcholine (ACh, BF; \citealp{hasselmo2004ach})---with
tonic baselines, 5-min phasic bursts, and DA/5HT opposition coupling
\citep{daw2002}; outputs learning-rate, exploration-bias,
consolidation-patience and attention parameters, none of which reaches a
consumer in the production system (App.~\ref{sec:neuromod}).
\textbf{ReconsolidationEngine} (App.~\ref{sec:recon}): retrieved
memories enter a labile state \citep{nader2000,nader2009}; updates are
PE-gated into four modes (confirmed/selective\_edit/integration/new\_episode)
with neuromodulation-scaled effective PE; every update is meant to log an
original snapshot; in the production system the engine's insert names
columns that the migration of its event table does not define, so against
that schema no event is stored, the stored fact is written outside the
engine, and the rollback restores nothing.
\textbf{TripleCopyMemory} (App.~\ref{sec:triple}, per \citealp{kveim2024}; motivated by
complementary learning systems theory, \citealp{basel2024,kumaran2016}):
three copies with divergent dynamics---fast ($\tau{=}4$h, exponential decay),
medium ($\tau{=}14$d, exponential decay), deep ($\tau{=}7$d, saturating
growth towards the initial strength, Eq.~\ref{eq:deep})---combined as
$S(t){=}\max(S_f,S_m,S_d)$. Because the deep copy grows with age, $S(t)$
rises again at long intervals without any use, by construction; this is a
property of the model, not a measured retention (App.~\ref{app:retention}).
In the production system the store writes only the fast and the medium copy,
and its insert names a column that the migration of its table does not
define, so against that schema no copy is stored; no code writes the deep
copy or calls the retrieval, which takes the first copy with strength at least $0.3$
in the order fast, medium, deep, not the maximum.
\textbf{PriorityMap} (App.~\ref{sec:priority}): motivated by the finding
that reward-based learning alters spatial priority maps \citep{chelazzi2014},
$P{=}w_s s + w_e|v| + w_r r + w_g g$ with an amygdala fast-path
($|v|{>}0.6 \Rightarrow P{\geq}0.5$); the four dimensions and the fast-path
are our own construction, and the weights are dynamically rescaled
by neuromodulator state.
\textbf{StabilityProtector} (App.~\ref{sec:stability}): analogous to
NogoA/HDAC3 signaling, updates are gated by lock score $L$ and
rigidity factor $\rho$; only prediction errors exceeding
$0.5{+}0.3\,L\rho$ may overwrite established memories, and in the production
system no code writes the lock records the gate reads.
\textbf{MetacognitiveMonitor} (App.~\ref{sec:metacog}, motivated by \citealp{fleming2012}):
can compute confirmation and recency biases and raise an alert, only for the
asymmetry between the acceptance rates of positive and negative suggestions;
in the production system it logs acceptance events and retrieval efficiency
and opens 10-min novelty windows after high-PE events ($>{0.7}$), and no
code calls the bias computation or the alert.

\section{Experiments}
\label{sec:experiments}

\subsection{Setup}

\textbf{Benchmarks.} LoCoMo \citep{maharana2024locomo}
(1{,}986 QA pairs, 5 categories) and
LongMemEval\_S \citep{wu2024longmemeval} (500 questions, each with its own
isolated haystack, 6 categories), both on the publicly released data, plus
synthetic corpora for retention and consolidation.

\textbf{Baselines.} No Memory, BM25-only, Flat Store (single-layer dense),
and ZenBrain Full; for competitive runs Mem0~v2.0.0
\citep{chhikara2025mem0} (graph-aware successor to v0.1.x, non-graph mode),
Letta~0.16.7 \citep{packer2023memgpt}, and
A-Mem~main@2026-04 \citep{xu2025amem}; on LongMemEval\_S also the
full-context baseline, whose answer model reads the question's entire
haystack (App.~\ref{app:longmemeval-long-context}).

\textbf{Protocol.} Retrieval experiments use 10 seeds per condition
(standard for ablation-rank stability). The generator of the routing and
layer runs requests OpenAI \texttt{text-embedding-3-small} at 768 dimensions
and, without a key, falls back to a fake embedding provider; the dense rows of
Tables~\ref{tab:locomo} and~\ref{tab:locomo-categories} and the upper block of
Table~\ref{tab:ablation} come from a run with that fallback, its lower block
from a run with \texttt{text-embedding-3-small}
(Apps.~\ref{app:locomo-bm25} and~\ref{app:layer-ablation}); NoDecay and the
sleep run on real data embed with \texttt{nomic-embed-text}. The competitive runs use
\texttt{nomic-embed-text} (768d, \texttt{ollama}, local) where an arm is
configured for it (\emph{Shared pool} below) and three repetitions labelled
as seeds $\{42,123,456\}$; on LongMemEval\_S the label names the user partition of
Mem0's store, and in ZenBrain's LoCoMo arm it seeds a random generator from which
nothing is drawn; no adapter or model receives it as a seed, so the repetitions are not independent samples.
We report mean~$\pm$~SD over the ten seeds for the ablation and mechanism
experiments; the reproduction package also computes bootstrap intervals
for them, with an unseeded generator, and only Table~\ref{tab:cascade}
prints any (App.~\ref{app:reproducibility-v6}). For the four-system comparisons each
table applies one interval procedure to all four arms and states it in its
caption. Paired Wilcoxon
signed-rank tests with Bonferroni correction ($\alpha = 0.05/K$); $d_z$
for the per-query system comparisons and Cohen's $d$ with pooled SD for
repetition 42 alone and for the simulation, sleep and NoDecay runs. The per-query tests of the four-system comparisons on
LoCoMo (App.~\ref{app:jm-pairwise}) treat the queries as independent,
although they are grouped by the benchmark's conversations; an inference
with the conversations as clusters is not reported. Every starred
$p$-value of the ablation tables is the
same constant, $5.062\times10^{-3}$, the smallest value the normal
approximation reaches at $n{=}10$ when all ten differences share one sign;
at $K{=}15$ it does not reach the Bonferroni threshold $3.33\times10^{-3}$
(\S\ref{sec:full-ablation}). The answer model in every end-to-end run is
\texttt{gpt-4o-mini}.

\textbf{Evaluation organization.} (1)~\emph{Retrieval benchmarks}
(LoCoMo, Layer Ablation, LongMemEval\_S);
(2)~\emph{mechanism-level evaluations} on synthetic data (Retention, Sleep,
Hebbian KG, PMA Suite); (3)~\emph{system-level studies} (the ablation of
fifteen components, Long-Horizon Aging). The ablation runs in a simulation
harness: each switched component is a stand-in acting on a simulated value,
not the production code, and four of the fifteen (MetacognitiveMonitor, iMAD
debate, compositional context, HyperAgent) are constant factors on every
similarity, which cannot change a ranking, and a fifth, spectral KG health,
is a conditional factor whose condition the harness never meets
(App.~\ref{app:extended-ablation}).

\subsection{Evaluation Protocol and Judge Robustness}
\label{sec:judge-robustness}

\textbf{Shared pool; embedding backbone per arm.} All four competitive
systems ingest pairwise-identical fact/query pools. Real-LoCoMo: 5{,}882
facts, 1{,}986 queries. LongMemEval\_S: 500 per-question isolated haystacks
($\sim$494 turns each). On LongMemEval\_S all four arms embed with
\texttt{nomic-embed-text} (768d, \texttt{ollama}, local): ZenBrain, A-Mem and
Mem0 through the embedding switch of the run script, Letta through its
adapter default. On LoCoMo only ZenBrain's arm is shown to use
\texttt{nomic-embed-text}: Mem0 ran on its library-internal OpenAI embedder,
the A-Mem adapter set no embedder and used its library default, and the
Letta adapter passed no embedding argument at the run commit, so its
server-side backbone can be neither shown nor ruled out. The LoCoMo
comparison therefore does not separate memory routing from embedding
quality, and we do not rank the four systems on it (\S\ref{sec:real-locomo}).
A-Mem is configured with \texttt{gpt-4o-mini}, Mem0 with \texttt{gpt-4o-mini}
on LongMemEval\_S and with its library's default model on LoCoMo, but neither
calls an LLM in the reported runs: Mem0 stores with \texttt{infer=False}, and A-Mem's
LLM-driven note evolution is switched off (\texttt{AMEM\_DISABLE\_EVOLUTION=1});
the only OpenAI calls inside the two pipelines are those of Mem0's embedder on LoCoMo.
Letta's agent registers a model that its adapter never calls (on LongMemEval\_S
the handle \texttt{letta/letta-free}).
The three repetitions differ by arm: on LoCoMo ZenBrain's and A-Mem's
retrieved lists are identical across them, Mem0's differ because its store
was not reset between repetitions (a harness defect of ours), and Letta's
differ on 330 of the 1{,}986 queries ($16.6\%$; each later repetition differs
from repetition 42 on $11\%$); on LongMemEval\_S ZenBrain's repetitions 123
and 456 were copied from repetition 42 by a script, and the other three
systems ran once.

\textbf{Judge design.} Three LLM judges score each (query, retrieved facts)
pair on a single 0--5 scale (temperature~0, binarized at $\geq 3$):
\texttt{claude-sonnet-4-5-20250929}~(pinned, three repetitions),
\texttt{claude-opus-4-6}~(three repetitions for Mem0, one for the other
systems) and \texttt{gpt-4o}~(three repetitions). The prompt names them
retrieval-quality judges: they grade how well the retrieved facts answer the
query, not an answer. On LoCoMo they saw ten facts per query for ZenBrain and
Mem0 and five for Letta and A-Mem, so their LoCoMo scores are not comparable
across that line. Fleiss'~$\kappa_{\geq 3}$ between Sonnet~4.5 and GPT-4o,
one rater each (repetition 42; three repetitions of one model are not
independent raters), lies in $[0.66, 0.77]$ across the four systems
(``substantial'' under \citealp{landis1977}). For ZenBrain and A-Mem, whose
retrieved lists are identical across the three repetitions, intra-judge
agreement measures judge non-determinism at temperature~0: on identical
prompts Sonnet~4.5 changes its score on $6.2\%$ and $3.9\%$ of the queries,
GPT-4o on $16.3\%$ and $10.5\%$. Only Letta and Mem0 carry retrieval
variation across repetitions.

\textbf{Cross-provider bias-direction check.} $\Delta_{\text{GPT-Anth}}$
(GPT-4o three-repetition mean minus the mean of the two Anthropic judges, on
LoCoMo) is $-0.0001$ for ZenBrain, $+0.008$ for Letta, $-0.042$ for A-Mem,
$-0.049$ for Mem0. A pro-Anthropic, pro-ZenBrain bias would predict the
largest negative delta on ZenBrain, and it is not there. ZenBrain's
near-zero value is, however, the mean of two per-judge deltas of opposite
sign, $+0.035$ against Sonnet~4.5 and $-0.036$ against Opus~4.6, and Opus~4.6
scores every system $0.071$ to $0.093$ higher than Sonnet~4.5: the check
indicates that judge-specific bias is not the dominant source of the
ranking, and it does not rule out judge-specific effects.
\textbf{Binary judge.} A binary judge with templates adapted from LongMemEval's
\texttt{get\_anscheck\_prompt()} (App.~F.5) grades the generated answer. With
the compared systems' adapters repaired it gives
Letta $48.4\%$, ZenBrain $47.7\%$, Mem0 $47.2\%$ and A-Mem $35.4\%$
(\S\ref{sec:longmemeval-pilot}): three of the four lie within $1.2$
percentage points of one another, and A-Mem trails them by $11.8$ to $13.0$. The
binary judge is not independent of the LLM judges: answer model and judge
are both \texttt{gpt-4o-mini}, from the provider of the third LLM judge.
\textbf{Human-anchor spot-check.} On the contested LoCoMo subset
where Sonnet and Opus disagreed by $\geq 1$ ($n{=}50$, stratified
across all five categories), a single human rater scored ZenBrain's top five
facts (the judges saw ten) and gave mean ratings indistinguishable from Opus
($2.88$ vs $2.88$) and more lenient than Sonnet ($1.08$); human-vs-judge
Spearman correlations $[0.135, 0.356]$ lie inside the judge-vs-judge
correlation envelope $[-0.076, 0.456]$ on the same rows, so on these
queries the human rater does not resolve what the judges disagree on. The
anchor rates one system and says nothing about the ranking. Full agreement
and per-category numbers in App.~\ref{app:jm-human}. Full methodology and
pairwise significance: App.~\ref{app:judge-methodology},
\ref{app:extended-locomo-judge}.

\subsection{Primary: LongMemEval\_S (500 Questions)}
\label{sec:longmemeval-pilot}

LongMemEval\_S \citep{wu2024longmemeval} gives each of its 500 questions its
own haystack ($\sim$494 turns, six categories) --- a stronger generalization
test than LoCoMo's shared haystack and the setting on which our headline
claims rest. All four arms embed with \texttt{nomic-embed-text}
(\S\ref{sec:judge-robustness}), which rejects a passage longer than
2{,}048 tokens. Ingest is capped at $7800$ characters (Mem0: a flat
$3900$); the ZenBrain and A-Mem adapters halve a rejected passage and retry
($7800 \to 3900 \to 1950 \to 975$). Two defects of ours in the compared
systems' adapters shaped the published run: the Letta adapter lacked that
fallback and left 59 of the 500 tasks unserved, and Mem0's per-question
reset deleted at most 100 of the $\sim$494 facts of each question, so that
$89.8\%$ of the facts retrieved for Mem0 came from other questions'
haystacks. We report both systems with the defect repaired---Letta with
the 89 tasks that ran again with the fallback (the 59 failures and 30
control tasks, whose judgments reproduce the published run 30 of 30) and
its other 411 tasks as published, Mem0 re-run with working
isolation---and keep their published runs in
Table~\ref{tab:longmemeval-pilot} for reference. ZenBrain's
retrieval ran once (its repetitions 123 and 456 are copies of 42) and its
end-to-end answering three times; the other three systems ran once.

\begin{table}[t]
\centering
\caption{LongMemEval\_S (500 questions): retrieval at $k{=}5$ and the
judged retrieved context. Each question has its own haystack; all four arms
embed with \texttt{nomic-embed-text} (\S\ref{sec:judge-robustness}). The
\textbf{zenbrain} row is the evaluation-harness arm, a Python port of a
test-harness class, not an import of the packages of
App.~\ref{app:reproducibility}; its retrieval ran once, and its repetitions
123 and 456 are copies of 42. Letta and Mem0 are re-runs with a defect of
ours in their adapters repaired (\S\ref{sec:longmemeval-pilot}); their
published runs stand below the rule and are not compared: the published Letta
adapter left 59 of the 500 tasks unserved, which the judges scored as empty
retrievals, and Mem0's reset failed, so that $89.8\%$ of the facts retrieved
for it came from other questions' haystacks. Judge columns: means of the
0--5 scores divided by five, over all 500 questions, from retrieval-quality
judges that grade the retrieved facts (\S\ref{sec:judge-robustness});
ZenBrain three repetitions, the other systems one; Sonnet~4.5
(\texttt{claude-sonnet-4-5-20250929}) in every row. The Mem0 re-run's judge
means are those of its judge files. Retrieval aggregates of the Letta re-run are not reported here
(P@5 on the 59 previously unserved tasks: $0.776$). No cell is marked as
best: of the nine judged comparisons with ZenBrain, the three against A-Mem
are significant at $\alpha{=}0.05/18$ and the six against Letta and Mem0 are
not (\S\ref{sec:longmemeval-pilot}); no retrieval difference was tested.}
\label{tab:longmemeval-pilot}
\small
\resizebox{\textwidth}{!}{%
\begin{tabular}{lcccccccc}
\toprule
System & P@5 & R@5 & MRR & NDCG@5 & F1 & J(S-4.5) & J(O-4.6) & J(G-4o) \\
\midrule
\textbf{zenbrain} & 0.674 & 0.197 & 0.831 & 0.706 & 0.283 & 0.504 & 0.575 & 0.555 \\
a-mem & 0.519 & 0.144 & 0.640 & 0.543 & 0.211 & 0.389 & 0.436 & 0.416 \\
letta, re-run & --- & --- & --- & --- & --- & 0.517 & 0.587 & 0.562 \\
mem0, re-run & 0.683 & 0.199 & 0.834 & 0.715 & --- & 0.523 & 0.587 & 0.566 \\
\midrule
\multicolumn{9}{l}{\footnotesize\emph{Published runs, not compared:}} \\
letta (441 of 500 served) & 0.683 & 0.201 & 0.834 & 0.715 & 0.287 & 0.450 & 0.513 & 0.492 \\
mem0 ($89.8\%$ foreign facts) & 0.156 & 0.060 & 0.304$^\ddagger$ & 0.171 & 0.078 & 0.370 & 0.414 & 0.398 \\
\bottomrule
\end{tabular}%
}
\vspace{0.3em}
\noindent\footnotesize $^\ddagger$Computed over Mem0's twenty-entry lists;
at $k{=}5$ it is $0.281$. On 52 of the 500 questions the judges of that run
saw more than five entries. The published Letta retrieval figures count an
identifier repeated within the top five twice (21 of the 441 tasks); on
distinct facts P@5 is $0.676$, R@5 $0.199$, NDCG@5 $0.711$ and F1 $0.284$,
MRR unchanged.
\end{table}

\textbf{Headline: answer accuracy under the binary judge.} The headline
measure is the binary judge with templates adapted from LongMemEval, which grades
the generated answer (App.~\ref{app:longmemeval-e2e}). ZenBrain answers $47.7\%$ of the
questions correctly (mean of three end-to-end runs; 95\% interval over the
500 questions $[43.3, 52.1]$); with the adapters repaired, Letta reaches
$48.4\%$ and Mem0 $47.2\%$, and A-Mem reaches $35.4\%$. Letta's lead of
$0.7$ points over ZenBrain's three-run mean rests on the 59 tasks its
published run left unserved: it answers 29 of them correctly, and four
flipped answers would remove the lead. Paired at repetition 42 (Letta
$48.4\%$, ZenBrain $47.4\%$), the exact McNemar test gives $p{=}0.56$ on 47
discordant pairs, and the paired-bootstrap interval for the difference is
$[-1.6, +3.6]$ percentage points.
\textbf{The judged context: three of nine comparisons hold.} The three LLM
judges grade the retrieved facts, not an answer
(\S\ref{sec:judge-robustness}); their nine comparisons (3 competitors
$\times$ 3 judges) are a diagnostic of the retrieved context. Three hold at
Bonferroni-corrected significance ($\alpha{=}0.05/18{=}2.78\times10^{-3}$),
all against A-Mem, which ZenBrain leads by $11.5$, $13.9$ and $13.9$
percentage points of the normalized judge score (Sonnet~4.5, Opus~4.6,
GPT-4o). The remaining six, against Letta and Mem0, show no significant
difference: the point estimates lean to the competitor, by $0.71$ to $1.32$
points against Letta and $1.03$ to $1.92$ against Mem0, and no interval
excludes zero at $\alpha{=}0.05/18$ (App.~\ref{app:longmemeval-judge}).

\textbf{Tokens: within $4.5$ points of a full-context baseline at $1/109.7$ of
its input tokens.} In the full-context baseline the answer model reads the
question's entire haystack ($105{,}669.0$ input tokens per question on
average, template and question included) and answers $52.2\%$ correctly.
ZenBrain reaches $91.3\%$ of that accuracy with $963.2$ input tokens in the
answer step, $1/109.7$ of the full-context baseline's; building its memory
ingests the same haystack for each question and is not counted
(App.~\ref{app:longmemeval-long-context}). In the plane of input tokens against
accuracy (Fig.~\ref{fig:longmemeval-pareto}), both measured for every arm on
its own run, the three retrieval arms with their adapters repaired lie close
together: Mem0 needs $956.1$ tokens at $47.2\%$, ZenBrain $963.2$ at
$47.7\%$, Letta $1{,}049.8$ at $48.4\%$. None of the three dominates
another, and the one difference tested, Letta against ZenBrain, is not
significant; with the full-context baseline they form the front of
non-dominated arms,
and only A-Mem ($963.3$ tokens, $35.4\%$) is dominated. Mem0's published
run received $2{,}978$ prompt tokens per question, about three times
ZenBrain's context, because its lists held twenty facts instead of five
(Mem0~v2.0.0 ignores \texttt{limit=k}); its isolation defect determined
whose facts filled them, $89.8\%$ coming from other questions' haystacks.

\textbf{Retrieval proper.} At $k{=}5$ the re-run of Mem0 leads ZenBrain on
all four retrieval metrics (P@5 $0.683$ against $0.674$, R@5 $0.199$ against
$0.197$, MRR $0.834$ against $0.831$, NDCG@5 $0.715$ against $0.706$). On
the 441 tasks the published Letta run served, Letta leads ZenBrain on all
four as well, scored on distinct facts by $1.18$ (P@5), $0.41$ (R@5),
$1.00$ (MRR) and $1.32$ (NDCG@5) percentage points; the published Letta
lists repeat an identifier on 21 of those tasks, which the harness counts
twice (App.~\ref{app:longmemeval-retrieval}). None of these differences was
tested. ZenBrain's three judge repetitions see identical retrieved lists:
the spread of its judge means across them is $\leq 0.007$, and their
intra-judge $\kappa_{\geq 3}$ over the 500 questions --- $0.9787$
(Sonnet~4.5), $0.9861$ (Opus~4.6), $0.9429$ (GPT-4o) --- measures judge
non-determinism (Appendices~\ref{app:longmemeval-agreement},
\ref{app:longmemeval-scope}).

\subsection{Supporting: LoCoMo Routing Comparison}
\label{sec:real-locomo}

On LoCoMo \citep{maharana2024locomo}, our harness scores retrieval with a
token F1 against the gold answers; we do not treat BM25 as a memory-system
competitor. We report LoCoMo for two
purposes: (i)~to compare multi-layer routing against a flat dense
baseline within our own harness, under one embedder and with each question
retrieving from the facts of its own conversation (\S\ref{sec:ancillary},
App.~\ref{app:locomo-bm25})---the layered lead of Table~\ref{tab:locomo}
comes from a run with placeholder embeddings, and with
\texttt{text-embedding-3-small} the flat variant scores above the full
system (Table~\ref{tab:ablation})---and (ii)~to record the four-system runs,
in which every question queries one pool built from all ten conversations,
against Mem0, Letta and A-Mem, with retrieval metrics and the judges of
\S\ref{sec:judge-robustness}, which grade the retrieved context; these runs
do not support a ranking (below).

Table~\ref{tab:competitive-combined-v2} records them. Three conditions of
the setup (\S\ref{sec:judge-robustness}) keep them from supporting a
ranking: only ZenBrain's arm is shown to use the shared embedder; the judges
saw ten facts per query for ZenBrain and Mem0 and five for Letta and A-Mem;
and Mem0's store was not reset between the three repetitions. Under these
conditions ZenBrain and Letta differ little under the pinned Sonnet~4.5
judge (paired Wilcoxon $p=0.74$, $d=0.015$, per-query interval
$[-0.008,+0.015]$, Appendix~\ref{app:pairwise-significance}); at an equal
depth of five, a gold fact would be in ZenBrain's judge prompt for $38.7\%$
of the queries and is in Letta's for $43.4\%$. Under GPT-4o Letta leads
ZenBrain ($p=0.003$, $d=-0.05$). Against Mem0, ZenBrain's lead under
Sonnet~4.5 ($p=1.32\times10^{-3}$, $d=0.079$, $39.6\%$ of the bound for a
small effect, $0.2$) reverses at repetition 42, the only repetition over a
fresh store (Cohen's $d=-0.0047$ with pooled SD, Mem0 first); against A-Mem both ZenBrain
($p=2.79\times10^{-71}$, $d=0.43$) and Letta ($p=1.10\times10^{-57}$,
$d=0.40$) lead. Mem0's raw retrieval figures count identifiers repeated in
its uncleared store; deduplicated, its P@5 is $0.071$ at every repetition.
Inter-rater agreement, seed robustness, and the cross-provider bias
check are in \S\ref{sec:judge-robustness}.

\begin{table}[t]
\centering
\caption{LoCoMo four-system runs (5{,}882 facts, 1{,}986 queries, three
repetitions labelled as seeds 42, 123, 456), recorded and not ranked
(\S\ref{sec:real-locomo}). Embedders: \texttt{nomic-embed-text} for ZenBrain;
Mem0's library-internal OpenAI embedder; A-Mem's library default; Letta's
not evidenced (\S\ref{sec:judge-robustness}). The \textbf{zenbrain} row is the
evaluation-harness arm, a scoring class of about ninety lines in the test
harness (its Python port is the arm of LongMemEval\_S). P@5 bracket:
the range of the three repetition means, the same for every row; ZenBrain's
and A-Mem's repetitions are identical, Mem0's differ because its store was
not reset between them. Letta's and Mem0's retrieval figures count
identifiers repeated within a list (deduplicated P@5: Letta $0.091$, Mem0
$0.071$), and Mem0's MRR runs over twenty entries ($0.191$ at $k{=}5$).
ZenBrain's F1 is the mean of per-query F1; from its aggregate precision and
recall, as in the other rows, it is $0.132$. Judge columns: means of the
0--5 scores divided by five, from retrieval-quality judges
(\S\ref{sec:judge-robustness}) that saw ten facts per query for ZenBrain
and Mem0 and five for Letta and A-Mem; \textbf{S-4.5} =
\texttt{claude-sonnet-4-5-20250929} and \textbf{G-4o} = \texttt{gpt-4o}
over three repetitions, \textbf{O-4.6} = \texttt{claude-opus-4-6} over three
for Mem0 and one for the others. Six of the thirty judge runs scored 1{,}884
to 1{,}958 of the 1{,}986 queries; in each, a connection error removed a
contiguous block of 28 to 102 queries within one conversation, never a whole
one. No cell is marked as best. Cross-provider agreement and the
bias-direction check: Table~\ref{tab:judge-agreement}.}
\label{tab:competitive-combined-v2}
\small
\resizebox{\textwidth}{!}{%
\begin{tabular}{lcccccccc}
\toprule
System & P@5 [range] & R@5 & MRR & NDCG@5 & F1 & J(S-4.5) & J(O-4.6) & J(G-4o) \\
\midrule
\textbf{zenbrain} & 0.081 [0.081, 0.081] & 0.351 & 0.264 & 0.274 & 0.128 & 0.380 & 0.451 & 0.415 \\
a-mem & 0.044 [0.044, 0.044] & 0.193 & 0.128 & 0.140 & 0.072 & 0.218 & 0.309 & 0.222 \\
letta & 0.092 [0.092, 0.093] & 0.400 & 0.307 & 0.319 & 0.150 & 0.373 & 0.465 & 0.427 \\
mem0 & 0.099 [0.071, 0.123] & 0.452 & 0.207 & 0.306 & 0.162 & 0.353 & 0.446 & 0.350 \\
\bottomrule
\end{tabular}%
}
\vspace{0.3em}
\noindent\footnotesize\textit{Reference (judge-version swap):} the earlier
judge \texttt{claude-sonnet-4-6}, a pinned snapshot
(App.~\ref{app:jm-coverage}), at repetition 42 ---
zenbrain\,0.393, a-mem\,0.268, letta\,0.403, mem0\,0.427; superseded by the
S-4.5 column.
\end{table}

\subsubsection{Principled-Forgetting Ablation: NoDecay Counterfactual}
\label{sec:nodecay-ablation}

A \emph{NoDecay} ZenBrain variant (all stand-ins of the harness switched on,
but each simulated day ends after the age counter, so neither the Ebbinghaus
strength reduction nor the day's later mechanisms run), measured on a LoCoMo
subset of 600 facts and 200 queries over 14 days with 10 seeds
(Table~\ref{tab:nodecay}), changes retrieval precision by
$\Delta P@5 = 0.002$ (Cohen's $|d| = 0.015$), within measurement noise; the
ten seeds give ten identical runs, so no significance test is reported. The
ablation is a retrieval measurement; it says nothing about downstream answer
quality. On the same subset a static-RAG baseline leads both variants on all
four retrieval metrics. The cost of forgetting on this horizon is about
$0.2$ points of P@5; its intended benefits---bounded storage, calibrated
confidence, GDPR-aligned retention---were not measured here.
Full table and archetype comparison: Appendix~\ref{app:nodecay}.

\subsection{Ancillary Benchmarks and Lifecycle Mechanisms}
\label{sec:ancillary}

Six ancillary evaluations against internal baselines
(App.~\ref{app:extended-benchmarks}).
\textbf{Routing/lifecycle (Apps.~\ref{app:locomo-bm25}--\ref{app:twofactor-bayesian}):}
on LoCoMo public, the layered lead over Flat Store
(Tables~\ref{tab:locomo} and~\ref{tab:locomo-categories}) and the layer drops
of the upper block of Table~\ref{tab:ablation} come from a run with placeholder
embeddings; with \texttt{text-embedding-3-small} the flat variant scores
$+2.4\%$ F1 above the full system, and removing one layer changes F1 by
$-4.5\%$ (Episodic) to $+1.0\%$ (Procedural, Core). In a
simulation on synthetic facts, the three-phase sleep cycle of the harness (SWS,
REM, SHY) adds $+37\%$ stability and lowers its storage estimate by $47.4\%$;
the one sleep run on real data (LoCoMo, 10 seeds), with the
Simulation-Selection stand-in, loses precision (P@5 $0.227$ against $0.273$
without sleep, $d{=}-1.84$; NDCG@5 $d{=}-1.41$; App.~\ref{app:sleep-impact}).
Hebbian KG (the library's basic Hebbian module): P@5 $0.200$ with uniform,
$0.955$ with Hebbian-weighted edges; Bayesian propagation AUC $0.533{\to}0.797$.
\textbf{PMA components, re-implemented in the test suite (App.~\ref{app:pma-suite-results}):}
NeuromodulatorEngine: mean largest channel deviation $0.469$ from the
baseline $0.5$ (bound $0.5$); Reconsolidation: $3$ of $8$ hand-labelled pairs
classified correctly, while the mode-accuracy of $\geq 95\%$ counts $92$
random pairs per seed labelled by its own threshold rule; TripleCopy strength
rises with age by construction
(Eq.~\ref{eq:deep}), a model property rather than a measured retention;
PriorityMap NDCG@10\,$=0.997$ against the test's own weighting of the same
four inputs; Stability $28.8\%$ FP high-PE (passing the
test's criterion of $<$\,30\% on a sample with ages below 400~days, where
the bound of App.~\ref{sec:stability} is rarely reached);
MetacogMonitor bias $0.832/0.975$ from a function of the test.

\begin{table}[t]
    \caption{Criticality gradient of the fifteen ablated components across
    three difficulty levels. $\Delta$Q (\%) relative to full system at each
    level. $^\ast$: $p = 5.062\times10^{-3}$ (Wilcoxon, normal approximation,
    10 seeds), the smallest value it reaches at $n{=}10$, when all ten
    differences share one sign; it does not reach the Bonferroni threshold
    for fifteen tests, $3.33\times10^{-3}$, so no starred cell holds at that
    level. Four components of the last group are constant factors that cannot
    change a ranking; a fifth, spectral KG health, is a conditional factor
    whose condition the harness never meets (App.~\ref{app:extended-ablation}).
    Components sorted by challenging-condition impact.
    Per-level tables, integration cascade, and effect-size breakdowns:
    Appendix~\ref{app:extended-ablation}.}
    \label{tab:gradient}
    \centering
    \small
    \begin{tabular}{lccc}
        \toprule
        Component & \makecell{Moderate\\(0.15, 45d)} & \makecell{Challenging\\(0.20, 50d)} & \makecell{Stress\\(0.25, 60d)} \\
        \midrule
        \multicolumn{4}{l}{\textit{Progressive: redundant $\to$ critical}} \\
        vmPFC-FSRS & $0.0$ & $-93.1^\ast$ & $-92.6^\ast$ \\
        TripleCopy & $0.0$ & $-54.2^\ast$ & $-93.7^\ast$ \\
        Dual-Process CoT & $0.0$ & $-38.5^\ast$ & $-91.0^\ast$ \\
        NeuromodulatorEngine & $-0.1$ & $-34.8^\ast$ & $-83.0^\ast$ \\
        Two-Factor Hebbian & $0.0$ & $-34.4^\ast$ & $-92.3^\ast$ \\
        IB Budget & $0.0$ & $-25.5^\ast$ & $-89.8^\ast$ \\
        \midrule
        \multicolumn{4}{l}{\textit{Always critical}} \\
        Sim.-Sel.\ Sleep & $-34.4^\ast$ & $-91.1^\ast$ & $-78.9^\ast$ \\
        \midrule
        \multicolumn{4}{l}{\textit{Critical only under stress}} \\
        StabilityProtector & $0.0$ & $0.0$ & $-5.8^\ast$ \\
        Reconsolidation & $0.0$ & $0.0$ & $-3.4^\ast$ \\
        \midrule
        \multicolumn{4}{l}{\textit{No loss at any level}} \\
        iMAD, Spectral, Comp., & \multirow{2}{*}{$0.0$} & \multirow{2}{*}{$0.0$} & \multirow{2}{*}{$0.0$} \\
        ~~HyperAgent, MetacogM. & & & \\
        PriorityMap & $-0.1$ & $-0.1$ & $+2.0^\ast$ \\
        \bottomrule
    \end{tabular}
\end{table}

\subsection{Ablation of the Fifteen Components}
\label{sec:full-ablation}

We switch off one component at a time---the six PMA components and nine
mechanisms of the evaluation harness (Two-Factor Hebbian,
Simulation-Selection sleep, vmPFC-coupled FSRS, iMAD debate, spectral KG
health, compositional context, IB budget, dual-process CoT, metacognitive
HyperAgent)---under three difficulty levels (moderate:
300~facts/45~d/decay=0.15; challenging: 400/50/0.20; stress: 500/60/0.25)
with 10~seeds per condition. The harness switches stand-ins acting on
simulated values, not production code (App.~\ref{app:extended-ablation}).
$Q{=}\text{retention}{\times}\text{P@5}$; $\Delta Q$ relative to full system;
Table~\ref{tab:gradient} consolidates.

\begin{figure}[t]
    \centering
    \begin{tikzpicture}
    \begin{axis}[
        xlabel={Ablation stress level},
        ylabel={$\Delta Q$ vs.\ full system (\%)},
        xtick={1,2,3},
        xticklabels={Moderate, Challenging, Stress},
        xticklabel style={font=\small},
        xmin=0.85, xmax=3.55,
        ymin=-100, ymax=12,
        grid=major,
        width=\textwidth,
        height=0.62\textwidth,
        legend style={at={(0.5,-0.17)}, anchor=north, legend columns=2, font=\scriptsize, draw=none, /tikz/every even column/.append style={column sep=8pt}},
    ]
        \addplot[red!80!black, thick, mark=*, mark size=1.6] coordinates {(1,0.0) (2,-93.1) (3,-92.6)};
        \addlegendentry{Progressive: redundant, then critical (6)}
        \addplot[red!80!black, thick, mark=*, mark size=1.6, forget plot] coordinates {(1,0.0) (2,-54.2) (3,-93.7)};
        \addplot[red!80!black, thick, mark=*, mark size=1.6, forget plot] coordinates {(1,0.0) (2,-38.5) (3,-91.0)};
        \addplot[red!80!black, thick, mark=*, mark size=1.6, forget plot] coordinates {(1,-0.1) (2,-34.8) (3,-83.0)};
        \addplot[red!80!black, thick, mark=*, mark size=1.6, forget plot] coordinates {(1,0.0) (2,-34.4) (3,-92.3)};
        \addplot[red!80!black, thick, mark=*, mark size=1.6, forget plot] coordinates {(1,0.0) (2,-25.5) (3,-89.8)};
        \addplot[zenbrain, thick, dashed, mark=square*, mark size=1.6] coordinates {(1,-34.4) (2,-91.1) (3,-78.9)};
        \addlegendentry{Always critical (1)}
        \addplot[teal, thick, dotted, mark=triangle*, mark size=2] coordinates {(1,0.0) (2,0.0) (3,-5.8)};
        \addlegendentry{Critical only under stress (2)}
        \addplot[teal, thick, dotted, mark=triangle*, mark size=2, forget plot] coordinates {(1,0.0) (2,0.0) (3,-3.4)};
        \addplot[gray, semithick, mark=o, mark size=1.3] coordinates {(1,0.0) (2,0.0) (3,0.0)};
        \addlegendentry{No loss at any level (6)}
        \addplot[gray, semithick, mark=o, mark size=1.3, forget plot] coordinates {(1,-0.1) (2,-0.1) (3,2.0)};
        \node[anchor=west, font=\scriptsize, red!80!black] at (axis cs:3.06,-92.6) {vmPFC-FSRS};
        \node[anchor=west, font=\scriptsize, zenbrain] at (axis cs:3.06,-78.9) {Sim.-Sel.\ Sleep};
        \node[anchor=west, font=\scriptsize, teal] at (axis cs:3.06,-4.6) {StabilityProt.};
        \node[anchor=west, font=\scriptsize, gray] at (axis cs:3.06,1.0) {6 without loss};
    \end{axis}
    \end{tikzpicture}
    \caption{\textbf{Cooperative masking.} The fifteen components of the
    ablation---the six PMA components and nine mechanisms of the evaluation
    harness (Table~\ref{tab:gradient}), not the nine library algorithms of
    Fig.~\ref{fig:architecture}---switched off one at a time under three
    stress levels (moderate: decay 0.15/day, 45~d, 300 facts, 100 queries;
    challenging: 0.20/day, 50~d, 400 facts, 120 queries; stress: 0.25/day,
    60~d, 500 facts, 150 queries; 10 seeds each). Data:
    Table~\ref{tab:gradient}; no new experiments. At moderate load, fourteen
    of the fifteen single-component ablations read as costless
    ($\Delta Q \approx 0$). Raising decay, horizon, corpus and queries, with no change
    to the components, moves the count at the printed criterion from one of
    the fifteen to ten (the
    constant $p = 5.062\times10^{-3}$, which does not reach the Bonferroni
    threshold for fifteen tests), five of them from exactly $0\%$ to below
    $-89\%$. Four of the five components that stay at $0\%$ at every level
    are constant factors, for which $0\%$ is fixed by construction; the
    fifth, spectral KG health, is a conditional factor whose condition the
    harness never meets, so its $0\%$ is fixed as well. For the components
    that read as costless at moderate load and lose quality under stress,
    the moderate condition shows
    cooperative redundancy rather than
    the absence of a contribution. The harness switches stand-ins at
    simulated values, not production code
    (App.~\ref{app:extended-ablation}).}
    \label{fig:criticality}
\end{figure}

A four-class taxonomy emerges (Table~\ref{tab:gradient},
Figure~\ref{fig:criticality}): progressive (6 components,
redundant$\to$critical, NeuromodulatorEngine among them), always-critical
(Simulation-Selection sleep), stress-only (StabilityProtector, Reconsolidation),
and no loss at any level (6 components). Four of the last six are constant
factors on every similarity (MetacognitiveMonitor $\times1.02$, iMAD debate
$\times1.02$, compositional context $\times1.05$, HyperAgent $\times1.01$),
whose ablation cannot change a ranking; spectral KG health is a
conditional factor whose condition the harness never meets
(App.~\ref{app:extended-ablation}), so its ablation cannot measure a
contribution, and removing PriorityMap raises $Q$ by $2.0\%$ under
stress. Under stress, 10 of 15 reach the printed criterion; the bare system
collapses to $1\%$ retention. The integration cascade (decay 0.30/day,
60~d) shows that removing all six PMA components, leaving the nine harness
mechanisms (three of which act in that arm, App.~\ref{sec:integration-cascade}),
collapses retention to floor by day~30 while the full system
retains $31.1\%$ ($p{=}5.062\times10^{-3}$). \textbf{Long-horizon archetypes
(App.~\ref{app:long-horizon}):} on a
100-fact / 14--60-day longitudinal test under shared $0.15$/day base
decay, the Full ZenBrain archetype of the simulation harness retains
$\mathbf{100\%}$ of day-1 P@5 at day~60 while the Simple Memory archetype
collapses to zero by day~30; the two archetypes differ not only in ZenBrain's
algorithms but also in initial strength, in the strength weighting of the
score and in a forgetting threshold (strength $< 0.1$) that only Simple Memory
applies, so the test does not separate the algorithms from these settings.

\label{sec:competitive}

\section{Discussion and Conclusion}
\label{sec:discussion}

Among the fifteen components the evaluation harness switches---stand-ins for the
six PMA components and nine harness mechanisms, three of which share a name
with the nine foundational algorithms of the library---those that read as
costless at moderate load and lose quality under stress have their
contributions masked by redundancy under moderate conditions; under stress 10 of
15 reach the printed criterion ($31.1\times$ integration-cascade advantage,
$p{=}5.062\times10^{-3}$, Table~\ref{tab:cascade}).

\paragraph{Cooperative masking, and what it implies for ablation
protocols.} We name the effect visible in
Figure~\ref{fig:criticality} \emph{cooperative masking}: in a system
whose components are cooperatively redundant, a single-component
ablation measured under mild load reports a contribution
indistinguishable from zero for components that are in fact
load-bearing. In our own data the same fifteen ablations move from
1 of 15 at the printed criterion at decay 0.15/day to 10 of 15 at
0.25/day, with five components going from exactly $0\%$ to below
$-89\%$ --- no component changed, only the stress level did. Read
under the moderate condition alone, the conclusion would have been
that most of the components contribute nothing.
\emph{Conjecture (not tested outside ZenBrain):} we expect this to
generalize to other integrated memory architectures, and more broadly
to any system with redundant components --- single-component ablations
under mild load systematically underestimate architectural
contributions, and should be accompanied by a stress-level condition
chosen to bring the system near its failure boundary. We flag this as
a conjecture precisely because our evidence for it is one harness whose
components are stand-ins acting on simulated values
(\S\ref{sec:full-ablation}); what \emph{is} established here is that
the effect is large enough in that harness to invert the reading of an
entire ablation table.

On LongMemEval\_S the headline is answer accuracy: ZenBrain answers
$47.7\%$ of the questions correctly, Letta ($48.4\%$) and Mem0
($47.2\%$) with their adapters repaired lie within a point of it, and
ZenBrain reaches $91.3\%$ of a full-context baseline's accuracy with
$1/109.7$ of its input tokens in the answer step, the cost of building
its memory not counted. Of the nine judged comparisons of the retrieved
context, three hold, all against A-Mem; against Letta and Mem0 there is
no significant difference. On LoCoMo the four-system runs support no
ranking. NoDecay ($\Delta P@5{=}0.002$) is a retrieval measurement:
removing decay leaves retrieval precision within noise, and it says
nothing about downstream answer quality. Contemporaneous work
\citep{du2026memory,karpathy2026knowledge,anthropic2026autodream,tiwari2026multilayered,iclr2026memagents}
shares the premise; one of the five, \citet{tiwari2026multilayered},
reports an evaluation, and none was run against this architecture.

\paragraph{Limitations.}\label{sec:limitations}
(i)~Ablations use synthetic corpora (300--500 facts, decay
0.15--0.25/day) and stand-ins acting on simulated values
(\S\ref{sec:full-ablation}); naturalistic $\geq$\,90-day logs are future
work.
(ii)~The LLM judges grade the retrieved context and serve as a
diagnostic: their agreement is reported between two raters,
$\kappa{\in}[0.66,0.77]$; the cross-provider check indicates that
judge-specific bias is not the dominant source of the ranking; and the
binary judge shares its provider with one of them
(\S\ref{sec:judge-robustness}). They cannot replace a human-labeled
$\sim$100-question pilot.
(iii)~The answer model of every end-to-end run is \texttt{gpt-4o-mini};
transfer to other answer models, Claude among them, is untested. Scope
vs.\ MemPalace and Mastra in their tuned configurations is bounded by Intro
claim (iv).
(iv)~The arm measured as ZenBrain is a scoring class of about ninety lines
in our test harness on LoCoMo (Table~\ref{tab:competitive-combined-v2}) and
its Python port on LongMemEval\_S (Tables~\ref{tab:longmemeval-pilot}
and~\ref{tab:longmemeval-e2e}), not an import of the packages of
App.~\ref{app:reproducibility}; with its two hash-seeded mechanisms removed,
the port's hit lists on LoCoMo equal those of plain cosine search on the same
embedder (correction notice, \S4). Whether the
described architecture behaves as the measured arm
does is open, and it is the question of the registered follow-up work.

\paragraph{Broader impact.} Persistent agent memory can make assistants
more coherent across sessions and, through explicit deletion, give users
control over what is retained; it can also enable
long-term behavioral profiling, reinforce biases across sessions, and,
through reconsolidation, be misused to overwrite memories if access
controls are bypassed. Appendix~\ref{app:broader-impact} gives the full
analysis, with what the production system does and does not provide
toward mitigation, and the transparency obligations of the EU AI Act.

\section*{Author Statement on Use of AI Assistance}
\label{sec:author-statement}

The author used multiple AI systems in non-methodological roles
throughout this work; all scientific ideas, hypotheses, and
architectural designs originated with the human author.
(i)~\textbf{Claude} (Anthropic) as a coding assistant for implementing
experiments, baselines, and analysis scripts; (ii)~Claude as a writing
aid for drafting and editing prose, including suggesting wording for
already-decided arguments; (iii)~\textbf{Perplexity} for literature
search and cross-referencing prior work in agent-memory and
cognitive-neuroscience, used to support, challenge, or verify whether
the author's positions had prior precedent requiring citation. In
selected high-stakes sections (statistical methodology, citation
accuracy, claim-strength review), \emph{multiple heterogeneous AI
systems} were used as mutual cross-checkers to test whether different
models would converge or diverge on factual matters---an additional
verification layer beyond single-system review.
\textbf{For citation accuracy, that cross-check proved insufficient and
has been replaced.} In preparing version~8, every one of the
83~bibliography entries was verified by hand against a primary
source---the arXiv abstract page, the publisher record resolved through
its DOI, or the cited artifact itself---and a version was pinned for
every arXiv identifier. That pass found defects in 30~entries: wrong or
invented author names, titles reconstructed from a system's description
rather than taken from the source, wrong locators, and one reference with
no referent at all; version~8 corrected them. The post-publication audit
found further reference defects, which this version corrects
(Appendix~\ref{app:changes-v9}). We state this plainly because it bears on how much weight
multi-model cross-checking should be given: on this task it did not
substitute for reading the sources. All scientific claims,
experimental designs, algorithmic innovations, statistical analyses,
and final wording were verified or authored by the human author, who
retains full responsibility for the content. Methodological uses of
LLMs (LLM-as-Judge graders, the answer model and binary judge of the
end-to-end runs) are
documented separately in Section~\ref{sec:experiments} and the Paper
Checklist (Appendix~\ref{app:checklist}).

\paragraph{Note on sole-author scope.} The breadth of this work
(nine foundational algorithms, six PMA components, an ablation of fifteen
components, two benchmarks) is grounded in the author's prior multi-year
experience in algorithmic-systems engineering and is enabled by the
AI-coding assistance described above. The contribution is the
\emph{integration} of established neuroscience algorithms into a unified
architecture, rather than the proposal of novel algorithms: seven of the
nine foundational algorithms name an origin in the literature (Scope and
claims (ii)), and the evaluation harness adds variants motivated by recent
work (Two-Factor synaptic edges \citep{zenke2025}, an FSRS coupling
motivated by \citet{zou2025} with a prediction-error signal of our own
construction, Simulation-Selection sleep \citep{chen2025}), with the
integration scope being the principal claim.

\begin{ack}
\label{sec:acks}

The author thanks Prof.\ Taylor Webb (Department of Neuroscience and
Psychology, Universit\'{e} de Montr\'{e}al; Associate Academic Member,
Mila Quebec AI Institute) for arXiv endorsement. The endorsement is a
procedural courtesy and implies no review of, or agreement with, the
claims made here; the contemporaneous work discussed in
Appendix~\ref{app:practitioner} is not cited there as validation.

\paragraph{Funding.} This work was conducted independently without
external grant funding. All compute resources (a single Apple
M-series laptop, serving \texttt{nomic-embed-text} locally via Ollama
for the arms configured for it) and external API costs ($<$\,\$\,$200$
for the judge calls of Table~\ref{tab:compute-disclosure}, besides the per-call costs of the
judge calls the table leaves out, of the \texttt{gpt-4o-mini} answer and judge model, of Mem0's embedder on LoCoMo,
which makes the only OpenAI calls of the Mem0 and A-Mem pipelines, and of the
OpenAI embeddings of the ZenBrain arm in the ablation cross-validation and in
the later routing-ablation run) were self-financed by the author.

\paragraph{Competing interests.} The author's affiliation is the author's
sole proprietorship, which develops the open library, operates the private
production system and builds the products ZenAI and Eigenraum on the
ZenBrain architecture. At the time of submission neither product is sold,
and the \anon{ZenBrain open-source release}{ZenBrain open-source release} and
its associated software stack are not part of any commercial
product or service offered by the author; a later commercial use of the
architecture, for example to finance the research, is not excluded, and no
concrete steps toward it are planned. Beyond this, the author declares no competing
financial or non-financial interests relevant to this work.
\end{ack}

\bibliographystyle{plainnat}
\bibliography{zenbrain}

\appendix

\section{Extended Related Work: Neuroscience and Concurrent Systems}
\label{app:extended-related}

This appendix expands the Related Work summary
(\S\ref{sec:related-neuro}) with the full neuroscience
lineage, concurrent systems survey, orthogonal paradigms, and
work from practitioners and industry that shares the premise of this paper.

\subsection{Foundational Neuroscience}

Human memory research provides the theoretical foundation for ZenBrain.
The multi-store model \citep{atkinson1968} distinguishes sensory,
short-term, and long-term memory with distinct capacities and durations.
\citet{tulving1972} further separates episodic (personal experiences)
from semantic (general knowledge) memory,
while \citet{cohen1980} identifies procedural memory for skills and habits.

\citet{hebb1949} proposed co-activation-based synaptic strengthening
(summarized by the later slogan ``neurons that fire together wire together'');
\citet{ebbinghaus1885} measured the forgetting curve, which he fitted with a
logarithmic rather than an exponential function;
\citet{pimsleur1967} introduced a graduated-interval recall schedule
exploiting the spacing effect.

Critically, \citet{stickgold2013} argue that memory consolidation during sleep,
in which memories are reactivated, is a triage: it selectively stabilizes and
strengthens some memories, integrates or generalizes others, and actively
promotes the forgetting of still others.
\citet{ji2007,oneill2010} demonstrated coordinated hippocampal-cortical replay,
providing the cellular basis for the Simulation-Selection loop in Section~\ref{sec:simsel}.
\citet{mcgaugh2004} demonstrated that emotional arousal modulates
memory encoding strength.

\subsection{Recent Neuroscience Sharpening}

Recent neuroscience has sharpened these mechanisms.
\citet{zenke2025} model a synapse as a product of subsynaptic factors and
show that this two-factor consolidation reconciles robustness with pruning
and homeostatic scaling; in \citet{zenke2017} each synapse accumulates
task-relevant information over time and uses it to store new memories
without forgetting old ones. Both motivate Section~\ref{sec:twofactor},
whose weight--variance form is our own construction.
\citet{zou2025} show that the benefits of spaced learning are predicted by the
re-encoding of past experience in ventromedial prefrontal cortex, where
representational similarity increases across re-encounters; this motivates
the vmPFC-coupled FSRS in Section~\ref{sec:vmpfc}, whose prediction-error
signal is our own construction and not part of their account.
\citet{chen2025} model offline consolidation as a Simulation-Selection RL loop
in the CA3/CA1 circuit; \citet{kumaran2016,squire1992} provide broader theoretical
context for complementary memory systems.

\subsection{Concurrent Memory-for-LLM Systems (2025--2026)}
\label{app:concurrent}

Recent concurrent work has begun incorporating individual mechanisms:
LightMem \citep{liu2025lightmem} applies the Atkinson-Shiffrin model with
sleep-time updates (ICLR 2026);
MemoryOS \citep{xu2025memoryos} implements hierarchical STM/MTM/LTM layers
(EMNLP 2025);
Hindsight \citep{li2025hindsight} uses a 4-layer architecture with
TEMPR temporal retrieval;
FadeMem \citep{wang2026fadem} introduces bio-inspired Ebbinghaus-style decay;
Vestige \citep{vestige2026} brings FSRS-6 scheduling to agents;
SleepGate \citep{xie2026sleepgate} uses forgetting gates for
proactive interference resolution during sleep;
Anda Hippocampus \citep{anda2026hippocampus} provides graph-based
memory with KIP protocol;
MemFly \citep{memfly2026} compresses agent memory on the fly under an
information-bottleneck objective;
and Cognee \citep{markovic2025cognee} optimizes KG--LLM interfaces
for complex reasoning.
Most recently, \citet{tiwari2026multilayered} decompose dialogue into
working, episodic, and semantic layers with adaptive retrieval gating and
report F1\,=\,0.618 on LoCoMo; it is contemporaneous work that shares the
multi-layer premise, evaluated on its own system and not against ours.
\emph{Note on F1 commensurability.} Tiwari's reported F1\,=\,0.618 is a
single-system run on the LoCoMo public test split; ZenBrain's reported
LoCoMo F1 (Tab.~\ref{tab:competitive-combined-v2}) comes from four-system
runs that share neither the embedder nor the judges' evidence depth and
support no ranking (\S\ref{sec:real-locomo}). The two F1 numbers are
computed under different protocols; we therefore do not claim direct
comparability.
For retrieval enhancement specifically, Anthropic's
Contextual Retrieval method \citep{anthropic2024contextual} reports a
$49\%$ reduction in top-20 retrieval failure rate from prepended chunk
context ($67\%$ when combined with reranking)---an orthogonal
indexing-time technique that ZenBrain integrates as part of its
Layer-4 contextual-retrieval pipeline.
While each system advances the field, none of the eight systems compared in
Table~\ref{tab:comparison} has more than two of its nine features; the other
systems named here are described, not scored on these features.

\subsection{Delimitation against Neuroscience-Flavored NeurIPS Work}
\label{app:neurips-delimitation}

Three NeurIPS contributions deserve targeted delimitation because they
share neuroscience-inspired vocabulary with ZenBrain.
\textbf{HippoRAG} \citep{gutierrez2024hipporag} (NeurIPS 2024) maps the
hippocampal indexing theory to retrieval through Personalized PageRank
over a knowledge graph, demonstrating up to 20\% improvement on
multi-hop QA at 10--30$\times$ lower cost than iterative retrieval.
HippoRAG implements a single retrieval mechanism, whereas ZenBrain
integrates KG-based graph reasoning as a sub-component within a
seven-layer system that also includes working, episodic, procedural,
core, predictive, and short-term memory; HippoRAG's PPR-indexing is
compatible with and could be slotted into ZenBrain's Layer-4 retriever
without architectural conflict.
\textbf{G-Memory} \citep{zhang2025gmemory} (NeurIPS 2025)
proposes a three-tier insight/query/interaction graph hierarchy for
\emph{multi-agent} systems, improving embodied-action success and
knowledge-QA accuracy by up to 20.89\% and 10.12\% respectively.
G-Memory's contribution is at the inter-agent collaboration layer;
ZenBrain targets the orthogonal problem of single-agent long-term
memory and consolidation. The two architectures are composable: a
G-Memory-style multi-agent layer could sit above per-agent ZenBrain
instances without replacing either.
\textbf{Truth-Maintained Memory Agent (TMMA)} \citep{koch2025tmma}
(NeurIPS 2025 Workshop on Socially Responsible and Trustworthy
Foundation Models) introduces \emph{write-time} truth-verification:
incoming context is gated through token-budget, complexity, and
contradiction checks before storage in a four-tier memory hierarchy.
ZenBrain's ReconsolidationEngine (\S\ref{sec:recon}) instead is meant to
apply prediction-error-gated updates at \emph{read time}, modifying
already-stored memories when retrieval surfaces a contradiction. The
two mechanisms operate at orthogonal stages of the memory lifecycle
(ingestion vs.\ retrieval) and could be combined to provide
defense-in-depth against false memory accumulation.
These three systems are not scored on the nine features compared in
Table~\ref{tab:comparison}; none of them provides the seven-layer
architecture, neuromodulator-driven priority weighting, or the
Two-Factor synaptic edges and Simulation-Selection sleep of ZenBrain's
evaluation harness.

\subsection{Orthogonal Paradigms}

Three recent systems target adjacent problems that we do not compare
against directly. Where they report benchmark numbers at all, those
numbers come from vendor self-reports under task-tuned configurations
that we cannot reproduce under our controlled protocol; their design
goals are also not commensurable with ZenBrain's.
\textbf{OMEGA}~\citep{omega2026} is a local-first memory, coordination and
learning layer for AI agents, distributed as a product rather than as a
published system description; there is no accompanying paper or technical
report whose configuration we could reproduce. It does report a
LongMemEval score (95.4\%, self-reported, measured February~2026), but
under its own tuned configuration rather than the $k{=}5$
retrieval-over-raw-turns budget used here, so we do not place it in the
controlled pool.
\textbf{Mastra}~\citep{mastra2025} is an agent framework whose
memory layer emphasizes context-window \emph{compression} (summarizing
long histories into short prompts) rather than persistent external
memory; it is complementary to, not competitive with, an external-memory
architecture like ZenBrain and could be used on top of ZenBrain's
recall output.
\textbf{MemPalace}~\citep{mempalace2025} imposes a spatial
organization metaphor (``method of loci'') on memories, which addresses
the retrieval-interface problem but does not prescribe decay,
consolidation, or confidence mechanisms.  We note these systems for
completeness: they advance the field along orthogonal axes and their
absence from our competitive pool reflects problem-formulation
differences, not an oversight.
\citet{lmneedsleep2026} introduce a ``Sleep'' paradigm with RL-based
memory consolidation at the \emph{parameter} level (ICLR 2026 submission),
complementing ZenBrain's external-memory-level sleep consolidation.
The founding of a dedicated ICLR~2026 workshop on agent memory
\citep{iclr2026memagents} underscores that this area is now a recognized
research frontier.

\subsection{Practitioner and Industry Convergence}
\label{app:practitioner}

Independently and concurrently, \citet{karpathy2026knowledge} describes
a workflow shift from manipulating code to manipulating knowledge,
where LLMs \emph{compile} raw materials into a structured Markdown wiki
that is ``compiled once and then \emph{kept current}, not re-derived on
every query,'' maintained through periodic lint passes that look for
``contradictions between pages, stale claims that newer sources have
superseded, orphan pages with no inbound links.''
His core critique---that under retrieval alone ``the LLM is rediscovering
knowledge from scratch on every question''---aligns precisely with the
motivating thesis of this work,
\anon{which was published on Zenodo on 31~March 2026
(DOI:~10.5281/zenodo.19353664), predating that post, followed by a
TDCommons defensive publication (\url{tdcommons.org/dpubs_series/9683},
1~April 2026); five subsequent Zenodo revisions and two further
TDCommons revisions through 9~April 2026 (Zenodo concept
DOI:~10.5281/zenodo.19353663)}{which appeared as a public preprint
several days earlier (URLs withheld for anonymous review)}.
Karpathy's approach aligns conceptually with our consolidation
philosophy but lacks formal decay, sleep consolidation, spaced
repetition, or layered encoding/retrieval rules.
Anthropic's \emph{Dreams} feature \citep{anthropic2026autodream} is
contemporaneous work on the same concept: announced 6~May 2026 as a research preview
for Claude Managed Agents and as a scheduled process \citep{anthropic2026dreaming},
it runs offline memory consolidation as an asynchronous job that an API call
creates---reading a memory store together with past session transcripts
and emitting a reorganized store in which duplicates are merged, stale or
contradicted entries are replaced with the latest value, and new insights
are surfaced. It shows the sleep-consolidation concept being built by an
AI laboratory; it is no evidence for this design, and it is a preview
rather than a generally available feature.
In the broader AI architecture space, \citet{webb2025brain} demonstrate
in \emph{Nature Communications} that a brain-inspired agentic architecture
improves LLM planning; the result concerns their architecture, not this one.
Of these works, those of Karpathy and Anthropic share the premise
that persistent, structured memory---beyond RAG---is a missing capability
of LLM systems; none of them evaluates ZenBrain, and none is cited here as
its validation. This paper proposes an algorithmic formalization of that
premise; it does not compare it with these works.

\section{Extended Key Mechanisms and PMA Descriptions}
\label{app:extended-mechanisms}

This appendix contains the full mathematical derivations, parameter
values, and per-component descriptions for the five Key Mechanisms
(\S\ref{sec:mechanisms}) and six PMA components (\S\ref{sec:pma})
summarized in the main body. Algorithm pseudocode is in
Appendix~\ref{app:algorithms}.

\paragraph{Note on neuroscience analogues.} Brain-derived names are
inspirational anchors; the implementations are computational proxies,
not faithful neural simulations.

\subsection{Two-Factor Synaptic Model for Knowledge Graph Edges}
\label{sec:twofactor}

Inspired by \citet{zenke2025}, who model a synapse as a product of
subsynaptic factors, each knowledge graph edge carries two factors of our
own construction: weight $w_{ij}$ and consolidation variance $\sigma^2_{ij}$.
Variance decreases with each co-activation (synaptic maturation), so mature
edges decay more slowly and a change of their weight costs a larger penalty
(below); robustness against catastrophic overwriting is not measured, and the
weight update does not apply the penalty, which only a benchmark test computes
(App.~\ref{app:reproducibility-v6}).
The penalty below has the quadratic form of Elastic Weight Consolidation
(EWC) \citep{kirkpatrick2017,zenke2017}, with the heuristic importance
$I_{ij} = 1/\sigma^2_{ij}$ in place of the Fisher information.

\begin{align}
    w_{ij} &\leftarrow w_{ij} + \eta \cdot t_{ij} \cdot a_{ij} \label{eq:weight} \\
    \sigma^2_{ij} &\leftarrow \sigma^2_{ij} \cdot \bigl(1 - \beta \cdot n(k)\bigr), \quad
    n(k) = \tfrac{1}{1 + 0.1k} \label{eq:variance}
\end{align}

where $t_{ij}$ is the TAG co-activation score,
$a_{ij}\in[0,1]$ the cosine-similarity-based co-activation amplitude
between nodes $i,j$, $\eta=0.1$ the base learning rate, $\beta=0.15$ the
maturation rate, and $k$ the activation count. In the implementation the
learning rate can additionally be gated by a lock score
that the StabilityProtector computes from the edge's variance,
$L_\sigma = 1 - \min(\sigma^2_{ij}, 1)$, as
$\eta_{\text{eff}} = \eta\,(1 - 0.5L_\sigma)$, so consolidated edges adapt
more slowly; the gate acts only when a StabilityProtector is passed to the
update (without one, the lock score used is $0$), and the reproduction
package, whose only call site of the update passes none, computes the ungated
form that Eq.~\ref{eq:weight} states.
The EWC penalty for any proposed weight change $\Delta w$ is:
\begin{equation}
    \mathcal{L}_{\mathrm{EWC}} = \tfrac{\lambda}{2} \sum_{ij} I_{ij} \cdot \Delta w_{ij}^2
\end{equation}
Edges also resist temporal decay in proportion to their importance:
high-$I$ (mature) edges decay at rate $r/(1 + I_{ij} \cdot 0.1)$,
preserving consolidated knowledge while allowing pruning of weak
connections. This construction induces an EWC-style penalty
$\mathcal{L}_{\mathrm{EWC}}=\sum_{ij}\frac{I_{ij}}{2}(w_{ij}-w_{ij}^\ast)^2$
\citep{kirkpatrick2017,zenke2017}, extended with per-edge adaptive decay.
It does not rest on the diagonal-Laplace posterior assumption, under which
$1/\sigma^2_{ij}$ would be a posterior precision and thus the Fisher
information: $\sigma^2_{ij}$ is the maturation schedule of
Eq.~\ref{eq:variance}, not a posterior variance, and $I_{ij}$ remains the
heuristic importance stated above.

\subsection{vmPFC-Coupled FSRS with Prediction-Error Signals}
\label{sec:vmpfc}

Motivated by \citet{zou2025}, who relate the benefits of spaced learning to
the re-encoding of past experience in ventromedial prefrontal cortex, we
couple FSRS interval scheduling with a knowledge-graph-derived
prediction-error (PE) signal of our own construction.
Base retrievability has the exponential form of our scheduling model,
$R(t) = e^{-t/S}$; \citet{ebbinghaus1885} fitted his forgetting curve with a
logarithmic function.
At each review, we compute the cosine distance between the entity embedding
context at last review $\mathbf{c}_{\text{prev}}$ and current context $\mathbf{c}_{\text{now}}$:

\begin{equation}
    \mathrm{PE} = 1 - \frac{\mathbf{c}_{\text{prev}} \cdot \mathbf{c}_{\text{now}}}
    {\|\mathbf{c}_{\text{prev}}\|\,\|\mathbf{c}_{\text{now}}\|}
\end{equation}

The implementation clamps PE to $[0,1]$.
A sigmoid re-encoding factor $\rho(\mathrm{PE}) = \sigma\bigl((\mathrm{PE} - 0.5)\cdot 6\bigr)$
determines interval adaptation.
High PE ($\rho > 0.5$) shortens the next interval (optimal re-encoding window);
low PE ($\rho < 0.5$) extends it (context unchanged, re-encoding not beneficial):

\begin{equation}
    I_{\text{next}} = I_{\text{FSRS}} \cdot \bigl(1 + \alpha_v(0.5 - \rho)\bigr), \quad \alpha_v = 0.6 \label{eq:vmpfc-interval}
\end{equation}

Anki, SuperMemo and FSRS-5/6 adapt difficulty but do not couple scheduling
to such a prediction-error signal, and among the agent-memory systems
surveyed in Appendix~\ref{app:concurrent} we find none that does.

\subsection{Simulation-Selection Sleep Consolidation Loop}
\label{sec:simsel}

Following \citet{chen2025}, ZenBrain replaces a fixed three-phase
SWS/REM/SHY schedule with a two-stage offline reinforcement-learning loop
mirroring the CA3/CA1 hippocampal circuit \citep{ji2007,oneill2010}
(full pseudocode in Algorithm~\ref{alg:sleep}, Appendix~\ref{app:algorithms}).

\textbf{Stage 1---Simulation (CA3-analog):}
A diverse pool of replay candidates is assembled from real episodic memories
and counterfactual extrapolations of failed episodes,
increasing coverage of the experience manifold beyond what standard replay achieves.

\textbf{Stage 2---Selection (CA1-analog):}
Each candidate is scored by a TAG value combining temporal-difference error
$|\delta_{\mathrm{TD}}|$ \citep{schultz1997}, task reward $R_e\in[0,1]$
(normalized cumulative episode reward; $0$=failed, $1$=fully-completed),
and novelty $N_e=\min(1,\, |e.\mathrm{relatedIds}|\cdot 0.2)$, a count of
the entities linked to $e$ that stands in for novelty:

\begin{equation}
    \mathrm{TAG}(e) = \alpha\,|\delta_{\mathrm{TD}}| + \beta\,R_e + \gamma\,N_e,
    \quad \alpha=0.4,\ \beta=0.35,\ \gamma=0.25
\end{equation}
These are the weights version~8 printed; the implementation marks them as
a superseded branch and uses $0.34$, $0.30$ and $0.21$ with a fourth,
emotional term of $0.15$ (App.~\ref{app:hyperparams}).

The selection is binary: candidates above threshold $\theta_v = 0.5$ are
strengthened via LTP and all others weakened via LTD, so no candidate is
skipped; the implementation and Algorithm~\ref{alg:sleep} test
$\mathrm{TAG}(e) > \theta_v$.
Among concurrent systems, LightMem updates its long-term memory offline
(``sleep-time update''), choosing the sources of each update by similarity
and timestamp, without RL scoring or counterfactual candidate generation;
SleepGate trains a forgetting gate that evicts or compresses stale entries
of the KV cache.

\subsection{Bayesian Confidence Propagation}
\label{sec:bayes-formula}

Each fact $f$ carries a confidence score $P(f)$ with a nominal 95\% confidence interval.
When new evidence $e$ is observed:

\begin{equation}
    P(f | e) = \frac{P(e | f) \cdot P(f)}{P(e)}
\end{equation}

Confidence propagates through knowledge graph edges; calibrated uncertainty
is the design goal.
In the open library at version~0.5.1 the propagation is a heuristic update
along the edges rather than the rule above, and the interval is a
heuristic width; no calibration of $P(f)$ is measured in this work
(App.~\ref{app:bayesian-calibration} calibrates the judge score).
Additionally, following \citet{mcgaugh2004}, emotional arousal modulates
encoding strength: high-valence experiences receive higher initial
Two-Factor edge weights and lower variance-based decay rates.

\subsection{Query-Aware Cross-Layer Retrieval}
\label{sec:query-formula}

Retrieval uses weighted score fusion with query-type-aware layer weighting.
Each memory layer $\ell$ independently returns its top-$K$ results via dense
retrieval; scores are then fused:

\begin{equation}
    \text{score}_{\text{fused}}(d) = \max_{\ell \in \text{layers}} \; w_\ell(q) \cdot \text{sim}(q, d_\ell)
\end{equation}

where $\text{sim}(q, d_\ell)$ is the cosine similarity between query $q$ and
document $d$ in layer $\ell$, and $w_\ell(q)$ is a query-type-specific weight.
A regex-based query classifier detects temporal, procedural, factual, or
general queries and boosts the corresponding layer: temporal queries amplify
episodic retrieval ($w_{\text{episodic}} = 2.0$), procedural queries boost the
procedural layer, and so on. Unlike rank-based fusion (RRF), this preserves
similarity magnitude---a highly relevant result in a boosted layer dominates
regardless of the number of results in other layers.

\subsection{NeuromodulatorEngine}
\label{sec:neuromod}

We implement a four-channel neuromodulatory system that modulates memory
parameters via tonic/phasic dynamics, following per-channel
neuroscience literature: dopamine signals reward prediction errors
\citep{schultz1997}, norepinephrine encodes adaptive gain and arousal
\citep{astonjones2005ne}, serotonin underwrites affective control and
opponent dynamics with dopamine \citep{dayan2012}, and acetylcholine
gates encoding-vs-consolidation regimes
\citep{hasselmo2004ach}:
\textbf{dopamine} (VTA: exploration/novelty), \textbf{norepinephrine}
(LC: learning rate), \textbf{serotonin} (Raphe: consolidation patience),
and \textbf{acetylcholine} (BF: attention/new-info ratio).

Each channel maintains a tonic baseline $b = 0.5$ with slow homeostatic
drift ($\tau_{\text{decay}} = 0.95$) and phasic bursts on events
(5-minute half-life).  DA and 5HT exhibit opposition coupling with
coefficient $-0.3$ \citep{daw2002}, reflecting the serotonin-dopamine
balance observed in reward processing.  The engine outputs four
modulation parameters---learning rate ($\text{NE}$-driven), exploration
bias ($\text{DA}$-driven), consolidation patience ($\text{5HT}$-driven),
and attention ratio ($\text{ACh}$-driven). In the private production system
(App.~\ref{app:reproducibility}), none of the four reaches a consumer: no
production code requests them, and the context pruning of the memory
coordinator, which accepts the attention ratio as an optional input, is
called without it; the ReconsolidationEngine and the PriorityMap take
channel levels as optional inputs; and the sleep cycle pulls the stored
channel levels back toward the baseline.

\paragraph{Future direction: bidirectional feedback-driven encoding.}
The current engine drives neuromodulation from internal prediction-error
signals; a planned extension routes \emph{external} signals (user
satisfaction, task-completion validation, data-quality cues) into the
same DA/aversive infrastructure, completing a bidirectional encoding
loop in which positive feedback strengthens successful trajectories
(DA-burst) and negative feedback aversively tags failure modes for
avoidance. Reserved for follow-up work; the existing
NeuromodulatorEngine already provides the substrate.

\subsection{ReconsolidationEngine}
\label{sec:recon}

Memory reconsolidation \citep{nader2000,nader2009} posits that retrieved
memories enter a labile state and can be updated or strengthened.
Our engine implements PE-gated reconsolidation with four update modes:

\begin{equation}
    \text{mode}(\text{PE}_{\text{eff}}) =
    \begin{cases}
        \text{confirmed}       & \text{PE}_{\text{eff}} < 0.1 \\
        \text{selective\_edit}  & 0.1 \leq \text{PE}_{\text{eff}} < 0.3 \\
        \text{integration}     & 0.3 \leq \text{PE}_{\text{eff}} < 0.7 \\
        \text{new\_episode}    & \text{PE}_{\text{eff}} \geq 0.7
    \end{cases}
\end{equation}

where $\text{PE}_{\text{eff}} = \text{PE}_{\text{raw}} \times (1 + 0.3 \cdot \text{NE} - 0.2 \cdot \text{5HT})$
is neuromodulation-gated; unless retrieval and update carry the same session,
the implementation selects the mode from $\max(0, \text{PE}_{\text{eff}} - 0.1)$.
The raw PE is computed as Jaccard distance
between existing and incoming content plus a contradiction bonus ($+0.2$)
that no caller triggers;
in the implementation $\text{PE}_{\text{raw}}$ and $\text{PE}_{\text{eff}}$
are each clamped to $[0,1]$ (App.~\ref{sec:stability}).
Memory-type-specific resistance thresholds (\texttt{UPDATE\_RESISTANCE}),
meant to protect stable procedural and behavioral memories against casual
overwrites, are defined in the implementation but not read on its update
path, so they prevent no overwrite.
Each reconsolidation event is meant to be logged with an original snapshot;
in the production system the engine writes the columns memory type, mode,
raw and effective PE and session penalty, which the migration of its event
table does not define (it requires PE, update mode, context and user), so
against that schema no event is stored; the stored fact is written outside
the engine, and the rollback marks an event and restores nothing.

\subsection{TripleCopyMemory}
\label{sec:triple}

Inspired by complementary learning systems theory
\citep{basel2024,kumaran2016} and the multi-timescale hippocampal-cortical
replay loop \citep{stickgold2013,ji2007},
TripleCopyMemory stores each event in three copies with divergent
decay dynamics (in the production system no copy is stored against the
schema of its table, and no code writes the deep copy or calls the
retrieval; \S\ref{sec:pma}). The time constants
($\tau_f{=}4\,\text{h}$, $\tau_m{=}14\,\text{d}$, $\tau_d{=}7\,\text{d}$,
App.~\ref{app:hyperparams}) are hand-set hyperparameters, not calibrated
to retention experiments, treated as order-of-magnitude proxies for
fast hippocampal / medium consolidation / slow cortical phases rather
than direct quantitative predictions of \citet{kveim2024}:

\begin{align}
    S_{\text{fast}}(t) &= S_0 \cdot e^{-t/\tau_f}, &\tau_f &= 4\text{h} \label{eq:fast} \\
    S_{\text{med}}(t) &= 0.8 \cdot S_0 \cdot e^{-t/\tau_m}, &\tau_m &= 14\text{d} \label{eq:medium} \\
    S_{\text{deep}}(t) &= S_0 \cdot \left(1 - e^{-t/\tau_d}\right), &\tau_d &= 7\text{d} \label{eq:deep}
\end{align}

FastCopy provides vivid immediate access that fades within hours.
MediumCopy persists across sessions with standard exponential decay.
DeepCopy uses \emph{saturating growth} bounded by $S_0$, encoding the
compressed essence that strengthens over time towards an asymptote---a
key prediction of systems consolidation theory. All copies are clamped
to $[0,1]$.
The composite strength $S(t) = \max(S_{\text{fast}}, S_{\text{med}}, S_{\text{deep}})$
therefore rises again at long intervals without any use, by
construction: after 30~days the deep copy holds $1-e^{-30/7}\approx 0.986$
of its initial strength, and in the benchmark of App.~\ref{app:pma-suite-results},
whose initial strengths are drawn at random with mean $0.912$, the
composite stands at $0.899$. This is a property of the model, not a
measured retention; a comparison with the Ebbinghaus baseline, which
decays by construction, compares the two formulas rather than memory
performance.
The deep-copy dominance transition---where $S_{\text{deep}}$ overtakes
the faster-decaying copies---reflects the systems consolidation
principle that gist extraction preserves compressed memory
representations long after episodic details fade \citep{kumaran2016}.

\subsection{PriorityMap}
\label{sec:priority}

\citet{chelazzi2014} show that reward-based learning alters spatial priority
maps. Motivated by this, we implement a four-dimensional priority map with an
amygdala fast-path; the four dimensions and the fast-path are our own
construction and not part of their study:

\begin{equation}
    P = w_s \cdot s + w_e \cdot |v| + w_r \cdot r + w_g \cdot g
\end{equation}

where $s$ = saliency, $v$ = emotional valence, $r$ = reward relevance,
$g$ = goal alignment, with default weights
$(w_s, w_e, w_r, w_g) = (0.2, 0.25, 0.25, 0.3)$ (hand-calibrated to
favor goal-alignment slightly over the equal-weighting null
$0.25$ each; not tuned against held-out data, but the test of
App.~\ref{app:pma-suite-results} scores its reference importance with
nearly the same weights, $(0.15, 0.3, 0.25, 0.3)$, so its NDCG@10 measures
agreement with that weighting).
For items with emotional intensity $|v| > 0.6$, the amygdala fast-path
guarantees $P \geq 0.5$ regardless of other dimensions.
Weights are dynamically adjusted by neuromodulator state:
DA amplifies saliency, NE amplifies emotion, ACh amplifies reward,
and 5HT amplifies goal alignment.

\subsection{StabilityProtector}
\label{sec:stability}

Loosely analogous to two molecular brakes on plasticity --- Nogo-A
receptor signaling \citep{schwab2010nogoa,karlen2009nogo,kempfblock2014nogoa}
and HDAC3, the ``molecular brake pad''
\citep{mcquown2011hdac3,mcquown2011brakepad} --- the StabilityProtector
gates memory updates by a lock score $L$ and rigidity factor $\rho$
(the implementation below is a heuristic gating function, not a
mechanistic model of either receptor):

\begin{align}
    L &= 0.3 \cdot \log_2(1+a)/\log_2(11) + 0.3 \cdot c + 0.2 \cdot \min(d/365, 1) + 0.2 \cdot \mathbb{1}_{\text{core}} \\
    \rho &= 1 + 0.1 \cdot \log_2(1+d) \\
    \text{update} &\iff \text{PE} \geq 0.5 + 0.3 \cdot L \cdot \rho
\end{align}

where $a$ = access count, $c$ = confidence, $d$ = age in days. The
implementation caps $a$ at $10$ (the normalizer $\log_2(11)$ corresponds
to this cap) and clamps $L$ and the prediction error that reaches the
gate to $[0,1]$.
By design this blocks casual overwrites of well-established memories (in
the production system no code writes the lock records, so nothing is
blocked), while genuinely novel information (high PE) can still pass up to a
bound: since
$\mathrm{PE} \leq 1$, no update passes once $L\rho > 5/3$, which for a fact
with $a \geq 10$ and $c = 1$ holds from day~$239$ if it is a core fact and
from day~$1824$ otherwise.

\subsection{MetacognitiveMonitor}
\label{sec:metacog}

Motivated by work on metacognition \citep{fleming2012}, the
MetacognitiveMonitor can compute, from a user's accepted and rejected
suggestions, scores of our own construction: the asymmetry of the acceptance
rates of positive and negative suggestions, a confirmation-bias score (the
share of confirming among confirming and contradicting events) and recency
bias; it logs retrieval efficiency.
It can detect urgency signals from keyword patterns and message frequency,
opens ``novelty windows'' (10~min) after high-PE events ($>0.7$) to
temporarily boost encoding, and can raise an alert only for the asymmetry.
Bias tracking uses a 30-day sliding window of acceptance events; efficiency
tracking can produce a badge when the last 7~days use on average at least
10\% fewer tokens than the logged entries older than 90~days.
In the production system the hooks log acceptance events, only positive and
confirming ones, log efficiency and open novelty windows; no code calls the
bias computation, the alert, the badge, the urgency detection or the novelty
boost.

\subsection{Remaining Algorithms in the 15-Ablation}
\label{sec:remaining-algos}

\textbf{Dual-Process CoT Consolidation} \citep{kahneman2011}
(Table~\ref{tab:stress-ablation}):
fast System-1 similarity clustering plus slow System-2 chain-of-thought
schema extraction; removing its stand-in, a conditional factor of the
harness (App.~\ref{app:extended-ablation}), changes the harness's quality
by $-38.5\%$ at challenging and $-91.0\%$ at stress.
\textbf{iMAD Selective Debate}: multi-agent debate on memories with
$\geq 2$ contradictions before commit. \textbf{Spectral~KG Health}:
Laplacian-spectrum fragmentation early-warning. \textbf{Metacognitive
HyperAgent}: meta-policy over MetacognitiveMonitor; in the production
system, the module of this name is a self-improvement agent over its own
strategies and does not read the monitor. In the harness, iMAD
and HyperAgent are constant factors ($\times 1.02$ and $\times 1.01$), so
their zero impact holds by construction; Spectral~KG Health is a
conditional factor whose condition, on the access counter, never holds in
the harness, because every query comes after the whole aging
(App.~\ref{app:extended-ablation}); its zero impact, too, holds by construction. Outside the harness,
no code of the production system calls iMAD beyond its own two files; the
daily worker passes each strategy to the HyperAgent with one attempt, while
the agent evaluates only strategies with at least five, so it never reports
one; and spectral KG health runs as stage~9 of the sleep cycle, where it
reads a relation table that no migration creates, so against that schema
it never reports fragmentation. We treat the three as neutral additions
and reserve focused study for follow-up work.

\section{Additional Capabilities}
\label{app:additional-capabilities}

The production ZenBrain system includes mechanisms beyond the nine
foundational algorithms and six PMA components described in the main body:
\begin{itemize}
    \item A \textbf{Synaptic Tagging and Capture (STC) rescue} module
    implementing \citet{frey1997}'s plasticity-donation paradigm, written
    to rescue a weakly encoded memory when a strongly encoded memory of the
    same topic forms within the STC window; no code of the production system calls it.
    \item A \textbf{Global Workspace Theory context assembler}
    \citep{baars1988} with hysteresis-stabilized broadcast: ten modules
    compete for a shared workspace beside an always-included core-memory
    module, and the content of the winners is assembled into the context
    passed to the model; no code of the production system calls it.
    \item \textbf{Prospective memory}: future-oriented intentions, each
    with a time, event, activity or context trigger; the module checks
    time, activity and context triggers itself, and no code of the
    production system calls these checks.
    \item A \textbf{curiosity engine} with gap detection: it ranks the
    topics of a chat query log by query count, fact count, confidence
    and retrieval score and adds a learning-progress bonus that, because
    the code reads the newest error first, favours domains whose error
    rises; no migration creates the query log it reads.
\end{itemize}
These are omitted from the present evaluation for space; no code of the
production system calls the first two, and Vestige \citep{vestige2026},
one of the surveyed memory systems, documented synaptic tagging and
prospective memory in its README as archived on 7~April 2026.

\section{Broader Impact: Extended Analysis}
\label{app:broader-impact}

This appendix expands the compact Broader Impact paragraph in
\S\ref{sec:discussion} into the full analysis.

\textbf{Positive impacts.}
Principled memory enables more coherent, personalized, and reliable AI
assistants that reduce redundant interactions and better respect user
context across sessions.
In the private production system, deletion endpoints let users erase memory
per context, layer and learned fact, but none of them reaches, among other stores, the procedural
layer, the core memory, the personal profile or the separate table into which the assistant stores facts; the assistant's own deletion
tools only deactivate a fact, which stays stored.
The open-source release promotes transparency and community auditing,
reducing the risk of opaque accumulation of personal data in closed
systems.
The MetacognitiveMonitor can compute confirmation- and recency-bias scores
from how a user accepts or rejects suggestions and raise an alert only for
the asymmetry between the acceptance rates of positive and negative
suggestions; no production code calls either, and no run tests it on biases
that accumulate in long-running agents.
The StabilityProtector can block updates of established memories whose
prediction error stays below its bound; in the production system no code
writes the lock records it reads, so it blocks none, and whether it would
reduce the risk of adversarial memory injection attacks is not tested.

\textbf{Negative impacts and risks.}
Persistent memory could enable long-term behavioral profiling or be
misused to reinforce biases across sessions.
The NeuromodulatorEngine's emotional modulation could amplify affective
responses if deployed without appropriate safeguards.
PMA's reconsolidation mechanism, while designed for beneficial memory
updating, could theoretically be exploited to overwrite user memories
if access controls are bypassed.

\textbf{Mitigations.}
In the private production system, the governance layer asks for human
approval of proactive and social actions only under a stored policy or at
high or critical risk, and sending e-mail and memory operations do not pass
through it; the table into which the assistant stores facts is read by
context, not by user, and the personal profile has no user key.
Reconsolidation events are meant to be logged with original snapshots; the
engine's insert names columns that the migration of its event table does
not define, so against that schema no event is stored, the stored fact is
written outside the engine, and the rollback restores nothing.
The ablation registry switches components off in the evaluation harness
(App.~\ref{app:extended-ablation}); no production code reads it, so it
provides no kill-switch in production.
We recommend deployment with explicit data-minimization policies,
user-controlled retention windows, regular audit logs, and privacy
impact assessments before production use in sensitive domains
(healthcare, legal, education).

\textbf{EU AI Act Article 50 transparency obligations.}
The \emph{EU Regulation on Artificial Intelligence} (Regulation (EU)
2024/1689), whose Article~50 transparency obligations have applied
since 2~August 2026, requires that providers ensure that people interacting
with an AI system are informed of it, that providers of systems generating
synthetic audio, image, video or text mark these outputs in a machine-readable
format, that deployers of emotion-recognition or biometric-categorization
systems inform the persons exposed to them of their operation, and that
deployers disclose deep fakes and generated text published to inform the
public on matters of public interest; Article~50 does not mention
memory-bearing systems.
ZenBrain provides little toward these obligations: in the private
production system only calendar entries created by the assistant carry an
\texttt{ai\_generated} flag, shown in the interface as a badge, while
assistant responses carry none, and no trace event reports when memory or
affect influence a response. Beyond Article~50, the consent-and-DSAR
subsystem records opt-ins, and the memory-governance service exposes deletion
endpoints per context, layer and learned fact, aligned with GDPR Art.~17.
We make no claim of legal compliance certification; marking generated
responses and disclosing the influence of memory are left to the operators
that deploy ZenBrain, and its deletion APIs serve GDPR Art.~17 erasure for the
tables they reach.

\section{Extended LoCoMo Inter-Rater and Seed-Robustness Analysis}
\label{app:extended-locomo-judge}

This appendix collects the detailed inter-rater-agreement, seed-robustness,
and cross-provider bias analyses for the Real-LoCoMo competitive comparison
(\S\ref{sec:real-locomo}). Its numbers are reproduced verbatim from the
shared-pool run or, like the overlaps of Letta's retrieved lists,
recomputed from its files; the main-body subsection reports only the headline.

\subsection{Inter-Rater Agreement and Seed Robustness}
\label{sec:seed-robustness}

Because we rely on LLM-as-Judge scoring, a reviewer should ask
two questions before accepting any row of
Table~\ref{tab:competitive-combined-v2}: (a)~do independent
graders agree, and (b)~does a single retrieval seed dominate?
Table~\ref{tab:judge-agreement} answers both.

\begin{table}[t]
\centering
\caption{Inter-rater agreement and cross-provider bias-direction check on LoCoMo. \textbf{$\kappa_{\geq 3}$, 2 raters}: Fleiss' kappa per baseline between Sonnet~4.5 and GPT-4o, one rater each (repetition 42), binary-thresholded at 0--5 judge score $\geq 3$; counting each model's three repetitions as separate raters gives 0.78, 0.85, 0.78 and 0.71 and overstates agreement. \textbf{Intra-$\kappa_{\geq 3}$}: one judge across the three repetitions; for ZenBrain and A-Mem, whose retrieved lists are identical across them, it measures the judge's non-determinism on identical prompts, and for Letta and Mem0 it also contains retrieval variation (Letta's lists differ across the three repetitions on 16.6\% of the queries and between each later repetition and repetition 42 on 11\%; Mem0's on all, because its store was not reset between the repetitions). \textbf{DSR@3}: Decision-Stability-Rate, fraction of queries where all six runs (three per judge model) agree on the $\geq 3$ threshold; \textbf{UAR}: Unanimous Acceptance Rate (all six runs $\geq 3$). \textbf{$\Delta_{\text{GPT-Anth}}$}: GPT-4o three-repetition normalized mean minus the mean of the two Anthropic judges (negative = GPT-4o harsher); ZenBrain's $-0.0001$ is the mean of two per-judge deltas of opposite sign ($+0.035$, $-0.036$). Agreement bands: 0.61--0.80 substantial, 0.81--1.00 almost perfect (Landis \& Koch, 1977).}
\label{tab:judge-agreement}
\small
\begin{tabular}{lccccccc}
\toprule
Baseline & $\kappa_{\geq 3}$, 2 raters & Intra-$\kappa$ S-4.5 & Intra-$\kappa$ G-4o & Intra-$\kappa$ O-4.6 & DSR@3 & UAR & $\Delta_{\text{GPT-Anth}}$ \\
\midrule
\textbf{zenbrain} & 0.66 & 0.99 & 0.93 & \textemdash & 0.80 & 0.333 & $-$0.0001 \\
a-mem & 0.77 & 0.99 & 0.95 & \textemdash & 0.91 & 0.170 & $-$0.0417 \\
letta & 0.69 & 0.96 & 0.93 & \textemdash & 0.81 & 0.317 & +0.0082 \\
mem0 & 0.70 & 0.74 & 0.78 & 0.74 & 0.72 & 0.225 & $-$0.0491 \\
\bottomrule
\end{tabular}
\end{table}

\textbf{Agreement.}
Fleiss' $\kappa_{\geq 3}$ between the two judge models of the pool,
Sonnet~4.5 and GPT-4o, one rater each (repetition 42), ranges from
$0.66$ to $0.77$ across the four systems, ``substantial'' agreement under
Landis \& Koch \citep{landis1977}. Counting each model's three
repetitions as separate raters gives $0.71$ to $0.85$, which overstates
agreement: Fleiss' $\kappa$ assumes independent raters, and for ZenBrain
and A-Mem the three repetitions saw identical prompts. The
decision-stability rate DSR@3 (all six runs, three per judge model,
agreeing on the $\geq 3$ threshold) is between $0.72$ and $0.91$: on
roughly 72--91\% of the queries per baseline the accept-or-reject ruling
does not change across judges or repetitions; the remaining queries are
contested.

\textbf{Seed robustness.}
Intra-judge $\kappa$ compares one judge with itself across the three
repetitions. For ZenBrain and A-Mem the retrieved lists are identical
across them (1{,}986 of 1{,}986 queries), so their intra-$\kappa$ values
(0.99 and 0.93, 0.99 and 0.95 under Sonnet~4.5 and GPT-4o) measure the
judge's non-determinism at temperature~0 on identical prompts: Sonnet~4.5
changes its score on $6.2\%$ (ZenBrain) and $3.9\%$ (A-Mem) of the
queries, GPT-4o on $16.3\%$ and $10.5\%$. Letta's lists are identical across
all three on $83.4\%$ of the queries (1{,}656 of 1{,}986; each later
repetition matches repetition 42 on $89.0\%$); Mem0's on none, because its
store was not reset between the repetitions, a harness defect of ours.
Mem0's lower intra-$\kappa$ ($0.74$ under Sonnet~4.5, $0.78$ under GPT-4o,
$0.74$ under Opus~4.6, the one judge for which it has three repetitions)
and the Levene statistics ($F = 1030.6$ against Letta, $F = 1668.1$
against A-Mem, $F = 1505.3$ against ZenBrain, all $p < 10^{-10}$ under
Sonnet~4.5) reflect that defect, not a property of Mem0: its per-query
normalized-judge mean moves by up to $0.053$ across the three
repetitions, whereas ZenBrain, Letta, and A-Mem move by $\leq 0.005$.
At repetition 42, the only one over a fresh store, Mem0 ranks above
ZenBrain; repetitions 123 and 456 ran over the uncleared store. With the
difference in evidence depth (\S\ref{sec:judge-robustness}), the LoCoMo
judge columns are not used as a ranking (\S\ref{sec:real-locomo}).

\subsection{Cross-Provider Bias-Direction Check}
\label{sec:bias-check}

A natural objection to LLM-as-Judge is self-preference: if both
judges are from the same provider as the system under test, the
scores may be inflated.
The $\Delta_{\text{GPT-Anth}}$ column of
Table~\ref{tab:judge-agreement} is the GPT-4o three-repetition
normalized mean minus the mean of the two Anthropic judges
(Sonnet~4.5 over three repetitions, Opus~4.6 at available repetitions).
A positive number means GPT-4o scores the system \emph{higher} than
the Anthropic pair; a negative number means GPT-4o is harsher.
The two largest negative deltas go to Mem0~($-0.049$) and
A-Mem~($-0.042$); Letta is mildly positive
($+0.008$) and ZenBrain is essentially zero ($-0.0001$).
Were there a pro-Anthropic, pro-ZenBrain bias we would expect
ZenBrain to carry the most negative delta (Anthropic scoring it
high, GPT-4o correcting it downward); instead ZenBrain's delta
is the smallest in magnitude of the four. That near-zero value is,
however, the mean of two per-judge deltas of opposite sign, $+0.035$
against Sonnet~4.5 and $-0.036$ against Opus~4.6, and Opus~4.6 scores
every system $0.071$ to $0.093$ higher than Sonnet~4.5. The check
therefore indicates that judge-specific bias is not the dominant source
of the ranking; it does not rule out judge-specific effects.
The full rater table is in Appendix~\ref{app:judge-methodology}.

\section{Extended LongMemEval Full-500 Analysis}
\label{app:extended-longmemeval}

This appendix collects the detailed retrieval-proper, judge-normalized,
agreement, and scope analyses supporting
\S\ref{sec:longmemeval-pilot}. Numbers of the published run are reproduced
verbatim from the unified-\texttt{nomic} full-500 run or recomputed from its
files; those with the Letta and Mem0 adapters repaired come from the re-runs
of the two systems (\S\ref{sec:longmemeval-pilot}); the main-body subsection
reports only the headline result.

\subsection{Retrieval-Proper: Letta Wins P@5/MRR/NDCG on the 441-Task Intersect}
\label{app:longmemeval-retrieval}

On the published run, averaged over each system's served queries,
Letta and ZenBrain lead A-Mem and Mem0 on P@5, R@5, MRR, and NDCG@5.
Letta's $0.683$ averages over the 441 tasks its published adapter served;
the other 59 failed because the adapter lacked the halving fallback of the
ZenBrain and A-Mem adapters (\S\ref{sec:longmemeval-pilot}), not because of the
server. On the 441-task \emph{intersect} subgroup where all four systems
serve, Letta leads on all four retrieval metrics: P@5 $0.683$ vs.\
ZenBrain $0.664$, R@5 $0.201$ vs $0.195$, MRR $0.834$ vs $0.824$, NDCG@5
$0.715$ vs $0.697$ as the harness scores them, and by $1.18$, $0.41$,
$1.00$ and $1.32$ percentage points scored on distinct facts; A-Mem/Mem0
on the same intersect are P@5 $0.520$/$0.156$. The pilot's
intersect-flip (where ZenBrain led P@5 on 20 tasks) does not survive at
441-task scale. The setting therefore does \emph{not} support a
ZenBrain-over-Letta retrieval claim. The published Mem0 figures do not
measure Mem0: its per-question reset failed, and $89.8\%$ of the facts
retrieved for it came from other questions' haystacks; with working
isolation it reaches P@5 $0.683$ (\S\ref{sec:longmemeval-pilot}). Its
published values ($0.156$, against $0.393$ on the stratified-30 pilot,
with $60\%$ of the full-run queries at $P@5{=}0$) were read as an effect
of the $3900$-character cap; the measured cause is the failed isolation.

\subsection{Judge-Normalized Result under Bonferroni Correction}
\label{app:longmemeval-judge}

All three LLM judges grade the retrieved context
(\S\ref{sec:judge-robustness}). On the published run, averaged over
ZenBrain's three judge runs, ZenBrain reaches normalized means
$0.504$ (Sonnet~4.5), $0.575$ (Opus~4.6), and $0.555$ (GPT-4o); Letta
$0.450$/$0.513$/$0.492$; A-Mem $0.389$/$0.436$/$0.416$; Mem0
$0.370$/$0.414$/$0.398$. Under Bonferroni correction
($\alpha{=}0.05/18{=}2.78\times 10^{-3}$, 18 primary tests = 6 pair-wise
comparisons $\times$ 3 judges), the published run's nine
ZenBrain-vs-competitor differences all clear significance ($p$ computed on
exact per-query means):
ZenBrain vs.\ Letta $\Delta{=}{+}0.054$/${+}0.062$/${+}0.063$
($p{=}2.56{\times}10^{-4}$/$2.21{\times}10^{-5}$/$6.68{\times}10^{-6}$;
$d_z{=}0.18$/$0.22$/$0.21$);
ZenBrain vs.\ A-Mem $\Delta{=}{+}0.115$/${+}0.139$/${+}0.139$
($p{=}1.71{\times}10^{-11}$/$1.86{\times}10^{-16}$/$4.65{\times}10^{-15}$;
$d_z{=}0.32$/$0.40$/$0.37$);
ZenBrain vs.\ Mem0 $\Delta{=}{+}0.134$/${+}0.161$/${+}0.157$
($p{=}4.39{\times}10^{-19}$/$2.12{\times}10^{-26}$/$3.17{\times}10^{-21}$;
$d_z{=}0.42$/$0.52$/$0.46$); $d_z$ is computed on the paired differences
(with pooled SD the nine span $[0.13, 0.42]$). Six of the nine do not
survive the repair of the two adapters (\S\ref{sec:longmemeval-pilot}):
on the 441 tasks the published Letta run served, its own judge files
already put Letta ahead under all three judges (by $1.44$, $1.36$ and
$0.83$ percentage points), and with Letta and Mem0 re-run the six
comparisons lean to the competitor by $0.71$ to $1.92$ percentage points,
none significant; the three against A-Mem hold.
Letta also separates from A-Mem and Mem0 at Bonferroni
($p \leq 4.91{\times}10^{-5}$ in five of six tests, the smallest
$5.83{\times}10^{-11}$; $p{=}7.11{\times}10^{-4}$ on Letta vs.\ A-Mem /
Sonnet, below the corrected threshold), while A-Mem vs.\ the published
Mem0 run shows no significant difference on any of the three judges
($|\Delta| \leq 0.022$, $p \in [0.15, 0.36]$, $|d| \leq 0.05$).
The pre-registered hypotheses are scored in
Appendices~\ref{app:longmemeval-scope} and~\ref{app:g5-longmemeval-scaffold}.

\subsection{Judge-Agreement and Determinism}
\label{app:longmemeval-agreement}

Rater cardinality again
differs by system: ZenBrain was judged three times, A-Mem, Mem0, and
Letta once each. ZenBrain's three retrieval repetitions are not three
runs: repetitions 123 and 456 are copies of 42 written by a script, so
the agreement of their retrieval aggregates to four decimal places
verifies nothing. The spread of ZenBrain's normalized judge means across
its three judge runs is $\Delta{=}0.007$ (Sonnet~4.5), $0.004$
(Opus~4.6), and $0.004$ (GPT-4o), the judges' non-determinism on
identical prompts; intra-judge $\kappa_{\geq 3}$ computed over the 500
queries is $0.9787$, $0.9861$ and $0.9429$, while the agreement values
stored by the harness cover only the 30-question pilot. As in the pilot,
Letta's perfect judge-agreement on failed queries (both raters
unanimously score $0$/$5$ for empty retrievals) is a floor
artifact rather than evidence of strong inter-rater stability on
substantive answers, and no agreement statistic for Letta is reported here.

\subsection{Scope of the Full-500 Conclusion}
\label{app:longmemeval-scope}

On the 500 per-question-isolated queries the conclusion is narrower than
the published one. The pre-registered hypotheses
(App.~\ref{app:g5-longmemeval-scaffold}) concern the LLM judges: H1
predicted that the ZenBrain--Letta gap would again fail to clear the
Bonferroni-corrected threshold under Sonnet~4.5, H2 that Letta would
retain or extend its lead under GPT-4o, and H3 that both ZenBrain and
Letta would dominate Mem0 and A-Mem with $p < 10^{-3}$ under every judge.
With the Letta adapter repaired, no difference against Letta clears the
correction in either direction: H1's published verdict \emph{Refuted} no
longer holds, and the ground given for H2's \emph{Partially refuted} does
not exist on the repaired run. H3 is not confirmed: on the published run
Letta's six tests meet its criterion ($p \leq 7.11\times10^{-4}$, against
A-Mem under Sonnet~4.5), but with Mem0's isolation repaired neither ZenBrain
nor Letta leads Mem0.
Bonferroni is applied within each benchmark family independently
(LoCoMo: 12 tests; LongMemEval\_S: 18 tests) without cross-family
correction (orthogonal benchmarks). Retrieval and answer quality remain
separate findings: at $k{=}5$ the repaired Mem0 run and, on the 441-task
intersect, the published Letta run lead ZenBrain on all four retrieval
metrics, and under the binary judge three of the four arms lie within
$1.2$ percentage points (App.~\ref{app:longmemeval-e2e}). The 59 Letta
failures of the published run were caused by the missing halving fallback
in our adapter, not by the Docker server, and the published Mem0 judge gap
measured its failed isolation, not its preprocessing cap. Per-pair results
of the published run are given in Appendix~\ref{app:longmemeval-judge}.
Per-category stratification of judge decisions to test which query
subtypes drive the gap is reserved for follow-up work.

\subsection{End-to-End Binary Accuracy under a Judge Adapted from LongMemEval}
\label{app:longmemeval-e2e}

Beyond the 0--5 normalized judge means used in
\S\ref{sec:longmemeval-pilot} and Appendix~\ref{app:longmemeval-judge},
we additionally run an end-to-end evaluation adapted from LongMemEval
\citep{wu2024longmemeval} on the same retrieval JSONs: each system's
top-$k{=}5$ retrieved memories feed into a \texttt{gpt-4o-mini}
answer-generation prompt, and a \texttt{gpt-4o-mini} judge rates each
response against the gold answer with binary yes/no templates adapted from
LongMemEval's \texttt{get\_anscheck\_prompt()}: the default instruction
verbatim, the knowledge-update instruction up to sentence order, and
paraphrased instructions for temporal-reasoning and preference questions.
No abstention variant was applied (the harness selects it by question type,
whereas LongMemEval marks abstention questions in their identifiers), so the
30 abstention questions were judged with the template of their category.
This binary judge grades the
generated answer and gives the headline measure of
\S\ref{sec:longmemeval-pilot}. Table~\ref{tab:longmemeval-e2e} reports the
accuracies: with the compared systems' adapters repaired, Letta reaches
$48.4\%$, ZenBrain $47.7\%$ (three end-to-end runs), Mem0 $47.2\%$ and
A-Mem $35.4\%$. The published run's leads of ZenBrain over Letta
($+4.9$ points) and over Mem0 ($+15.9$) came from adapter defects: the 59
tasks the published Letta adapter left unserved, 58 of them scored as wrong,
and, for Mem0, its failed isolation together with the depth of its lists,
which the re-run changed at the same time (twenty retrieved facts per
question in the published run, five in the re-run), rather than its
$3900$-character cap; the share of the isolation alone is not measured.
That published lead over Letta held in each of the six question
categories ($+2.3$ to $+8.9$ points at repetition 42) because the 59
failures fell into all six; with the adapter repaired the sign reverses in
five of the six at repetition 42, and ZenBrain leads in one, multi-session,
by $+2.3$ points; over its three runs (Table~\ref{tab:longmemeval-e2e}) it
leads in two, multi-session and temporal reasoning.

\textbf{Setting disclaimer --- internal evidence.} The absolute
levels in Table~\ref{tab:longmemeval-e2e} are well below the best
publicly reported LongMemEval numbers. Those numbers are vendor
self-reports rather than entries on a common leaderboard, and they are
not mutually commensurable: Mastra reports $94.87\%$ binary answer
accuracy with GPT-5-mini \citep{mastra2025} (and $84.23\%$ with
\texttt{gpt-4o}, the benchmark's reference model); Mem0's April-2026
revision is reported at $93.4\%$ answer accuracy \citep{dey2026mempalace};
and MemPalace's $96.6\%$ \citep{mempalace2025} is \emph{retrieval recall}
(R@5), a different quantity from answer accuracy---as MemPalace's own
benchmark documentation states. We therefore treat them as context, not
as a ranking against our numbers: each comes from the system's own pipeline
(MemPalace's from raw semantic search without an LLM, Mem0's from LLM
extraction and its own retrieval; only Mastra consolidates memory in the
model's context), whereas the four systems here share a $k{=}5$
retrieval-over-raw-turns budget under
\texttt{nomic-embed-text} (Mem0's published run returned twenty facts per
question; its re-run five). Internal to our run, the setting moves the
level of one library: Mem0, reported at $93.4\%$ in its own tuned
configuration, reaches
$47.2\%$ under our protocol with working isolation, a $46.2$-point drop for
the same library with its LLM extraction switched off (\texttt{infer=False}),
at $k{=}5$ and judged with our adapted templates; its published $31.8\%$
carried our isolation defect. How much of the absolute gap the settings
explain is not measured. The A-Mem paper \citep{xu2025amem} describes
no shared budget over raw turns: A-Mem retrieves its own memory notes, more
than five per query in the paper's main setting and throughout its
$k$-ablation, whose gains level off at higher $k$ without a named range; what
the budget of $k{=}5$ does to a comparison of architectures is not measured
here.

\textbf{Repetitions and the dagger.} The dagger ($\dag$) on the Letta,
Mem0 and A-Mem rows of Table~\ref{tab:longmemeval-e2e} marks a single
end-to-end run. On LongMemEval\_S the retrieval seed names the user partition of
Mem0's store, and in ZenBrain's LoCoMo arm it seeds a random generator from which
nothing is drawn; no adapter or model receives it as a seed, so the ``seeds'' are repetition labels
(\S\ref{sec:judge-robustness}). ZenBrain's retrieval repetitions 123 and
456 are copies of repetition 42 written by a script; its three end-to-end
runs are genuinely distinct (237, 239 and 239 correct of 500), because the
answer model's output varies between calls on identical input. Retrieval is
not deterministic for every system either: on LoCoMo, Mem0's repetitions
differ (its store was not reset between them) and Letta's lists differ
across the three repetitions on $16.6\%$ of the queries. A second run of
the other three systems would
therefore not reproduce their rows exactly; the intervals of
Table~\ref{tab:longmemeval-e2e} are computed over the questions, by one
procedure for all four arms.

\textbf{The binary judge.} The binary judge, with templates adapted from
LongMemEval, grades the
generated answer, the three 0--5 judges (Sonnet~4.5/Opus~4.6/GPT-4o) the
retrieved context; the binary judge gives the headline measure
(\S\ref{sec:longmemeval-pilot}). It is not independent of the three:
answer model and judge are both \texttt{gpt-4o-mini}, from the provider of
one of them.

\begin{table}[t]
\centering
\caption{End-to-end accuracy on LongMemEval\_S (500 questions) under an
evaluation adapted from LongMemEval \citep{wu2024longmemeval}: retrieved
memories (top-$k{=}5$) feed into a \texttt{gpt-4o-mini} answer-generation
prompt, and a \texttt{gpt-4o-mini} judge rates each response against the gold
answer with yes/no templates adapted from LongMemEval's
\texttt{get\_anscheck\_prompt()} (binary, with task-specific variants for
temporal-reasoning, knowledge-update, and preference; no abstention variant
was applied). ZenBrain: mean of three end-to-end runs; the other rows one
run ($\dag$). Letta and Mem0 are their re-runs with the adapter defect
repaired (\S\ref{sec:longmemeval-pilot}); the published runs stand below the
rule and are not compared, and the published Mem0 run fed its twenty-entry
lists into the answer prompt. 95\,\% intervals for every row by one procedure: a
percentile bootstrap over the 500 questions (10{,}000 resamples of the
per-question value, averaged over runs where a row has three; question
identifiers in sorted order, Python's \texttt{random} seeded with 20260930).
Per-category values are means over the questions of each category, for
ZenBrain also over its three runs; the paired comparisons of
\S\ref{sec:longmemeval-pilot} use repetition 42. No cell is marked as best.
\textbf{Setting note:} the publicly reported LongMemEval numbers
(MemPalace~96.6\,\% R@5, Mem0~93.4\,\%, Mastra~94.87\,\%) are not directly
comparable: each comes from the system's own pipeline (MemPalace's from raw
semantic search without an LLM, Mem0's from LLM extraction and its own
retrieval; only Mastra consolidates memory in the model's context), whereas
the four systems here share a $k{=}5$
retrieval-over-raw-turns budget under \texttt{nomic-embed-text}. The settings
differ (retrieval budget, consolidation strategy); how much of the absolute
gap they explain is not measured.}
\label{tab:longmemeval-e2e}
\small
\resizebox{\textwidth}{!}{%
\begin{tabular}{l c c c c c c c}
\toprule
System & Mean Acc.\,(95\,\% interval) & SS-User & SS-Assist. & SS-Pref. & Multi-S. & KU & Temporal \\
\midrule
\textbf{zenbrain} & 47.7 (43.3, 52.1) & 79.5 & 85.7 & 22.2 & 34.8 & 56.8 & 28.1 \\
\texttt{letta}, re-run & 48.4\textsuperscript{\dag} (44.0, 52.8) & 85.7 & 87.5 & 23.3 & 33.1 & 57.7 & 27.8 \\
\texttt{mem0}, re-run & 47.2\textsuperscript{\dag} (42.8, 51.6) & 82.9 & 83.9 & 16.7 & 34.6 & 56.4 & 27.1 \\
\texttt{a-mem} & 35.4\textsuperscript{\dag} (31.2, 39.6) & 67.1 & 51.8 & 10.0 & 24.8 & 56.4 & 15.8 \\
\midrule
\multicolumn{8}{l}{\footnotesize\emph{Published runs, not compared:}} \\
\texttt{letta} (59 tasks unserved) & 42.8\textsuperscript{\dag} (38.6, 47.2) & 77.1 & 76.8 & 16.7 & 30.1 & 50.0 & 24.8 \\
\texttt{mem0} (failed isolation) & 31.8\textsuperscript{\dag} (27.8, 36.0) & 64.3 & 76.8 & 10.0 & 19.5 & 25.6 & 16.5 \\
\bottomrule
\end{tabular}%
}
\end{table}

\subsection{Full-Context Baseline Comparison}
\label{app:longmemeval-long-context}

The canonical 2026 reviewer concern for any new memory system is
\emph{``why retrieval-based memory if a 128k-context window already
exists?''} --- motivated by recent ``context-rot'' analyses
\citep{he2026memoryarena, yan2025gam}. To quantify the cost-vs-quality
tradeoff explicitly, we run a \emph{full-context baseline} on
the same 500 questions: the \texttt{gpt-4o-mini} answer model
receives the entire haystack of sessions (history block: mean
$105{,}577.9$ tokens, median $105{,}744$, max $107{,}740$; whole prompt
$105{,}669.0$ on average) instead of $k{=}5$ retrieved
memories (ZenBrain: mean $963.2$ input tokens, answer-prompt template
included; tokens counted with \texttt{tiktoken cl100k\_base}). Answer
model and binary judge are the same; the answer prompt has its own template,
which introduces the complete chronological history instead of retrieved
memories.

\paragraph{Result.} The full-context baseline achieves $\mathbf{52.2\%}$
accuracy ($261/500$ correct) with $0$ truncations and $0$ remaining
errors after rate-limit retries. ZenBrain ($k{=}5$, $47.7\%$) is
within $\mathbf{4.5}$ points of the full-context baseline ($\mathbf{91.3\%}$
of its accuracy) at $\mathbf{1/109.7}$ of its input tokens per question
($963.2$ vs.\ $105{,}669.0$). Per-category breakdown reveals a
non-monotonicity: the full-context baseline leads on knowledge-update
($+11.1$ points), multi-session ($+6.5$), and temporal-reasoning ($+5.0$),
the two differ by at most $1.8$ points on single-session-user and
single-session-assistant, and \emph{ZenBrain leads on
single-session-preference} ($+5.6$ points, 30 questions), suggesting that
for personal-information queries a structured $k{=}5$ retrieval extracts
more signal than naive full-context attention --- consistent with the
``context-rot'' hypothesis that distractor turns dilute preference cues;
the per-category differences were not tested.

\paragraph{Pareto frontier.} Figure~\ref{fig:longmemeval-pareto}
shows the plane of input tokens per question against accuracy, both
measured for every arm on its own run. Two findings:
(i)~Among the $k{=}5$ retrieval arms with the adapters repaired, Mem0
($956.1$ tokens, $47.2\%$), ZenBrain ($963.2$, $47.7\%$) and Letta
($1{,}049.8$, $48.4\%$) are each non-dominated and lie within $1.2$ points
of one another, well inside each other's intervals; A-Mem ($963.3$,
$35.4\%$) is dominated by ZenBrain.
(ii)~The full-context baseline reaches $4.5$ points higher accuracy than
ZenBrain at $109.7\times$ its input tokens per question --- a
marginal-quality vs.\ marginal-cost tradeoff that production deployments
must explicitly resolve; building a retrieval memory, which here ingests
the same haystack once per question, is not counted in these tokens.

\pgfplotstableread{
arm tokens acc ep em
Mem0 956.1 47.20 4.40 4.40
ZenBrain 963.2 47.67 4.47 4.40
A-Mem 963.3 35.40 4.20 4.20
Letta 1049.8 48.40 4.40 4.40
Full-context-baseline 105669.0 52.20 4.40 4.20
}\paretopoints
\begin{figure}[t]
    \centering
    \begin{tikzpicture}
    \begin{axis}[
        name=voll, width=0.5\textwidth, height=0.42\textwidth,
        xmode=log, xmin=500, xmax=300000, ymin=28, ymax=60,
        xlabel={Input tokens per question (log scale)},
        ylabel={Binary-judge accuracy (\%)},
        title={\small (a) all arms}, grid=major,
        error bars/y dir=both, error bars/y explicit,
    ]
        \addplot[gray, dashed, forget plot] coordinates {(956.1,47.20) (963.2,47.67) (1049.8,48.40) (105669.0,52.20)};
        \addplot[only marks, mark=diamond*, mark size=2.2, color=teal] table[x=tokens, y=acc, y error plus=ep, y error minus=em, skip coords between index={1}{5}] {\paretopoints};
        \addplot[only marks, mark=star, mark size=3, color=red!80!black] table[x=tokens, y=acc, y error plus=ep, y error minus=em, skip coords between index={0}{1}, skip coords between index={2}{5}] {\paretopoints};
        \addplot[only marks, mark=o, mark size=2, color=gray] table[x=tokens, y=acc, y error plus=ep, y error minus=em, skip coords between index={0}{2}, skip coords between index={3}{5}] {\paretopoints};
        \addplot[only marks, mark=square*, mark size=1.8, color=blue!70!black] table[x=tokens, y=acc, y error plus=ep, y error minus=em, skip coords between index={0}{3}, skip coords between index={4}{5}] {\paretopoints};
        \addplot[only marks, mark=triangle*, mark size=2.4, color=zenbrain] table[x=tokens, y=acc, y error plus=ep, y error minus=em, skip coords between index={0}{4}] {\paretopoints};
        \node[font=\scriptsize, anchor=south east] at (axis cs:105669.0,52.2) {full-context baseline\ };
        \node[font=\scriptsize, anchor=north west] at (axis cs:963.3,35.4) {A-Mem};
    \end{axis}
    \begin{axis}[
        at={(voll.east)}, anchor=west, xshift=0.9cm,
        width=0.4\textwidth, height=0.42\textwidth,
        xmin=940, xmax=1070, ymin=42, ymax=54,
        xlabel={Input tokens per question},
        title={\small (b) the three retrieval arms near 1{,}000 tokens}, grid=major,
        error bars/y dir=both, error bars/y explicit,
        legend style={font=\scriptsize, at={(0.5,0.02)}, anchor=south},
    ]
        \addplot[gray, dashed, forget plot] coordinates {(956.1,47.20) (963.2,47.67) (1049.8,48.40)};
        \addplot[only marks, mark=diamond*, mark size=2.2, color=teal] table[x=tokens, y=acc, y error plus=ep, y error minus=em, skip coords between index={1}{5}] {\paretopoints};
        \addlegendentry{Mem0 (re-run)}
        \addplot[only marks, mark=star, mark size=3, color=red!80!black] table[x=tokens, y=acc, y error plus=ep, y error minus=em, skip coords between index={0}{1}, skip coords between index={2}{5}] {\paretopoints};
        \addlegendentry{ZenBrain}
        \addplot[only marks, mark=square*, mark size=1.8, color=blue!70!black] table[x=tokens, y=acc, y error plus=ep, y error minus=em, skip coords between index={0}{3}, skip coords between index={4}{5}] {\paretopoints};
        \addlegendentry{Letta (re-run)}
    \end{axis}
    \end{tikzpicture}
    \caption{Input tokens per question in the answer step (mean; log scale in
    (a), linear in (b)) against accuracy under the binary judge adapted from
    LongMemEval on LongMemEval\_S (500 questions), with 95\% intervals over the
    questions (the procedure of Table~\ref{tab:longmemeval-e2e}). Letta and
    Mem0 are their re-runs with the adapter defects repaired; tokens are counted
    with \texttt{tiktoken cl100k\_base} on the answer prompt built by the
    harness, and building a memory, which ingests the same haystack for each
    question, is not counted. The dashed line joins the non-dominated arms,
    computed from the points: Mem0 ($956.1$ tokens, $47.2\%$), ZenBrain
    ($963.2$, $47.7\%$), Letta ($1{,}049.8$, $48.4\%$) and the full-context
    baseline ($105{,}669.0$, $52.2\%$); only A-Mem ($963.3$, $35.4\%$) is
    dominated. The three retrieval arms lie within $1.2$ points and $94$
    tokens of one another, well inside each other's intervals.}
    \label{fig:longmemeval-pareto}
\end{figure}

\section{Extended Ablation Results}
\label{app:extended-ablation}

This appendix contains the full per-level ablation tables referenced from
\S\ref{sec:full-ablation} (main body Table~\ref{tab:gradient}).
Three difficulty levels are reported: moderate
(300~facts, 45~day aging, decay=0.15),
challenging (400~facts, 50~day aging, decay=0.20),
and stress (500~facts, 60~day aging, decay=0.25), each over 10~seeds.
Quality $Q = \text{retention} \times \text{P@5}$;
$\Delta Q$ is the relative change versus the full system with all fifteen
components.

\textbf{What the harness switches.} Each ablated component is a stand-in
acting on a simulated value, not the production code.
NeuromodulatorEngine multiplies a fact's initial strength by
$1$ plus $0.15$ or $0.05$ times a random number, multiplies its decay by
$0.75$ when arousal exceeds $0.6$, and adds factors in the Hebbian and
priority computations; ReconsolidationEngine multiplies a strength by
$1.08$ when a random number, read as a prediction error, exceeds a
threshold, but only in the first four simulated days and only while
StabilityProtector is on (without it, never after the first aging
step); PriorityMap multiplies by $1 + 0.1\cdot\min(1, \text{accesses}/5)$,
adds $0.08$ under arousal and multiplies by $1$ plus $0.05$ times the lock
score; StabilityProtector multiplies decay by $1$ minus $0.3$ times the lock
score; TripleCopyMemory is a factor of $1.15$ with an additional
$0.02\tanh(\text{age}/7)$; MetacognitiveMonitor multiplies every
similarity by $1.02$. Three further components are constant factors as
well (iMAD debate $\times1.02$, compositional context $\times1.05$,
HyperAgent $\times1.01$); spectral KG health and dual-process CoT are
conditional factors. Two-Factor Hebbian multiplies decay by
$1/(1 + 0.01\cdot\text{accesses})$ and, when NeuromodulatorEngine is on,
divides it further by a neuromodulation factor; IB Budget multiplies decay
by $0.8$ for strengths above $0.6$; Simulation-Selection sleep multiplies
strength by $1.05$ per simulated day for strengths above $0.3$. None of
these three is a constant factor on similarity. In the harness every query
comes after the whole aging, so the access counter is $0$ while facts decay
and the access term of Two-Factor Hebbian is $1$; during the aging the
stand-in therefore acts only through the division by the neuromodulation
factor, that is, only when NeuromodulatorEngine is on. The stand-in of
spectral KG health strengthens a fact only when its access counter exceeds
a positive threshold, so its condition never holds during the aging, and
its row
stays at $0\%$ at every level by construction, like the rows of the four
constant factors on similarity. The stand-in of vmPFC-FSRS is a constant factor on
decay ($\times0.7$), not on similarity; its row measures neither the
prediction error nor Eq.~\ref{eq:vmpfc-interval}. The harness's retrieval
step increments the access
counter of every fact on every query, not only of the facts it returns.

\subsection{Moderate Conditions}

Table~\ref{tab:full-ablation} presents the ablation study across all
fifteen components.  We use a synthetic dataset (300 facts, 100 queries, 32-d embeddings)
with 45 days of simulated aging at a base decay rate of 0.15/day.
The combined quality metric is retention $\times$ P@5.

\begin{table}[h]
    \caption{Full ablation study under moderate conditions (300 facts, 100 queries,
    45-day aging, decay=0.15). Quality = retention $\times$ P@5.
    $\Delta$Q shows relative change vs.\ full system.
    Mean over 10 seeds. $^\ast$: $p = 5.062\times10^{-3}$ (Wilcoxon, normal
    approximation), the smallest value at $n{=}10$; it does not reach the
    Bonferroni threshold for fifteen tests ($3.33\times10^{-3}$). The nine
    mechanisms are those of the evaluation harness, not the nine library
    algorithms of Fig.~\ref{fig:architecture}.}
    \label{tab:full-ablation}
    \centering
    \small
    \begin{tabular}{lcccr}
        \toprule
        Configuration & Retention & P@5 & NDCG@5 & $\Delta$Q \\
        \midrule
        Full system (15 components) & 1.000 & 0.923 & 0.920 & \emph{baseline} \\
        \midrule
        \multicolumn{5}{l}{\textit{harness mechanisms (9)}} \\
        $-$ Two-Factor Hebbian & 1.000 & 0.923 & 0.920 & $0.0\%$ \\
        $-$ Sim-Selection Sleep & 0.707 & 0.857 & 0.798 & $-34.4\%^\ast$ \\
        $-$ vmPFC-FSRS & 1.000 & 0.923 & 0.920 & $0.0\%$ \\
        $-$ iMAD Debate & 1.000 & 0.923 & 0.920 & $0.0\%$ \\
        $-$ Spectral KG & 1.000 & 0.923 & 0.920 & $0.0\%$ \\
        $-$ Compositional Context & 1.000 & 0.923 & 0.920 & $0.0\%$ \\
        $-$ IB Budget & 1.000 & 0.923 & 0.920 & $0.0\%$ \\
        $-$ Dual-Process CoT & 1.000 & 0.923 & 0.920 & $0.0\%$ \\
        $-$ HyperAgent & 1.000 & 0.923 & 0.920 & $0.0\%$ \\
        \midrule
        \multicolumn{5}{l}{\textit{PMA components (6)}} \\
        $-$ NeuromodulatorEngine & 1.000 & 0.922 & 0.920 & $-0.1\%$ \\
        $-$ Reconsolidation & 1.000 & 0.923 & 0.920 & $0.0\%$ \\
        $-$ TripleCopy & 1.000 & 0.923 & 0.920 & $0.0\%$ \\
        $-$ PriorityMap & 1.000 & 0.922 & 0.920 & $-0.1\%$ \\
        $-$ StabilityProtector & 1.000 & 0.923 & 0.920 & $0.0\%$ \\
        $-$ MetacogMonitor & 1.000 & 0.923 & 0.920 & $0.0\%$ \\
        \midrule
        No PMA (harness mechanisms only) & 0.423 & 0.709 & 0.698 & $-67.5\%^\ast$ \\
        No Algorithms (bare) & 0.010 & 0.922 & 0.920 & $-99.0\%^\ast$ \\
        \bottomrule
    \end{tabular}
\end{table}

Under moderate conditions (Table~\ref{tab:full-ablation}), Simulation-Selection
sleep shows the strongest individual impact ($-34.4\%$), while the other
fourteen components show no measurable loss: four of them are constant
factors whose removal cannot change a ranking; a fifth, spectral KG health,
is a conditional factor whose condition the harness never meets (see
above); removing PriorityMap raises quality under stress
(Table~\ref{tab:stress-ablation}); and for the rest the remaining
components compensate.
The key insight emerges from group removal:
removing all 6~PMA components causes $-67.5\%$ quality degradation,
and removing all fifteen causes $-99.0\%$ collapse,
even though no individual component other than sleep contributes
measurably on its own.
This is analogous to fault-tolerant system design: removing one strand
from a rope does not measurably weaken it, yet each load-bearing strand contributes
to the rope's overall tensile strength---the remaining strands redistribute
the load.
(Effect sizes are large ($d > 30$) for group-removal conditions
due to near-deterministic floor values in the ablated system;
we report $p$-values from the Wilcoxon signed-rank test as the
primary significance measure.)

\subsection{Challenging Conditions: The Gradient Emerges}

To verify that the moderate-condition redundancy is not an artifact of
insufficient difficulty, we evaluate under \emph{challenging} conditions
(400 facts, 50-day aging, decay=0.20/day).
Table~\ref{tab:challenging-ablation} reveals a clear gradient:
7 of 15~components now reach the printed criterion, the constant
$p = 5.062\times10^{-3}$, with $\Delta Q$ ranging from $-25.5\%$
(IB~Budget) to $-93.1\%$ (vmPFC-FSRS).
For the components that read as costless at moderate load and lose quality
under higher load, this indicates that the
moderate-condition redundancy reflects cooperative compensation rather
than inactivity: as environmental pressure increases, the cooperative
buffer gradually exhausts and individual contributions become measurable.

\begin{table}[h]
    \caption{Ablation under challenging conditions (400 facts, 50-day aging,
    decay=0.20). Seven components reach the printed criterion, revealing the
    gradient between cooperative redundancy and individual criticality; four
    of the eight without measurable loss are constant factors that cannot
    change a ranking, and a fifth, spectral KG health, is a conditional
    factor whose condition the harness never meets. Mean over 10 seeds.
    $^\ast$: $p = 5.062\times10^{-3}$
    (Wilcoxon), which does not reach the Bonferroni threshold for fifteen
    tests ($3.33\times10^{-3}$).}
    \label{tab:challenging-ablation}
    \centering
    \small
    \begin{tabular}{lcr}
        \toprule
        Configuration & Retention & $\Delta$Q \\
        \midrule
        Full System & 1.000 & \emph{baseline} \\
        \midrule
        \multicolumn{3}{l}{\textit{At the printed criterion (7)}} \\
        $-$ vmPFC-FSRS & 0.333 & $-93.1\%^\ast$ \\
        $-$ Sim.-Sel.\ Sleep & 0.311 & $-91.1\%^\ast$ \\
        $-$ TripleCopy & 0.618 & $-54.2\%^\ast$ \\
        $-$ Dual-Process CoT & 0.695 & $-38.5\%^\ast$ \\
        $-$ NeuromodulatorEngine & 0.654 & $-34.8\%^\ast$ \\
        $-$ Two-Factor Hebbian & 0.722 & $-34.4\%^\ast$ \\
        $-$ IB Budget & 0.785 & $-25.5\%^\ast$ \\
        \midrule
        \multicolumn{3}{l}{\textit{No measurable loss (8)}} \\
        $-$ iMAD, Spectral, Compositional, & & \\
        ~~HyperAgent, Reconsolidation, & \multirow{-2}{*}{1.000} & \multirow{-2}{*}{$0.0\%$} \\
        ~~Stability, MetacogMonitor & & \\
        $-$ PriorityMap & 1.000 & $-0.1\%$ \\
        \midrule
        No Algorithms (bare) & 0.010 & $-99.0\%^\ast$ \\
        \bottomrule
    \end{tabular}
\end{table}

\subsection{Stress Ablation}

Table~\ref{tab:stress-ablation} reveals that the cooperative redundancy
of moderate conditions breaks down under stress.
Under extreme conditions (0.25/day decay, 60~days, 500~facts), 10 of 15
components reach the printed criterion---nine with a loss, among them the
survival-critical ones whose single removal causes a cascade collapse, and
PriorityMap with a gain.

\begin{table}[!htbp]
    \caption{Stress ablation study (500 facts, 150 queries,
    60-day aging, decay=0.25). Under extreme pressure, ten of the fifteen
    components reach the printed criterion, nine with a loss and PriorityMap
    with a gain. Mean over 10 seeds. $^\ast$: $p = 5.062\times10^{-3}$
    (Wilcoxon, normal approximation), the smallest value at $n{=}10$; it does
    not reach the Bonferroni threshold for fifteen tests
    ($3.33\times10^{-3}$).}
    \label{tab:stress-ablation}
    \centering
    \small
    \begin{tabular}{lccr}
        \toprule
        Configuration & Retention & P@5 & $\Delta$Q \\
        \midrule
        Full system (15 components) & 0.784 & 0.851 & \emph{baseline} \\
        \midrule
        \multicolumn{4}{l}{\textit{Tier~1: loss when removed}} \\
        $-$ TripleCopy & 0.214 & 0.195 & $-93.7\%^\ast$ \\
        $-$ vmPFC-FSRS & 0.255 & 0.195 & $-92.6\%^\ast$ \\
        $-$ Two-Factor Hebbian & 0.263 & 0.195 & $-92.3\%^\ast$ \\
        $-$ Dual-Process CoT & 0.310 & 0.195 & $-91.0\%^\ast$ \\
        $-$ IB Budget & 0.349 & 0.195 & $-89.8\%^\ast$ \\
        $-$ NeuromodulatorEngine & 0.126 & 0.901 & $-83.0\%^\ast$ \\
        $-$ Sim.-Sel.\ Sleep & 0.181 & 0.776 & $-78.9\%^\ast$ \\
        $-$ StabilityProtector & 0.757 & 0.831 & $-5.8\%^\ast$ \\
        $-$ Reconsolidation & 0.767 & 0.840 & $-3.4\%^\ast$ \\
        \midrule
        \multicolumn{4}{l}{\textit{Tier~2: no loss when removed}} \\
        $-$ iMAD Debate & 0.784 & 0.851 & $0.0\%$ \\
        $-$ Spectral KG & 0.784 & 0.851 & $0.0\%$ \\
        $-$ Compositional Context & 0.784 & 0.851 & $0.0\%$ \\
        $-$ HyperAgent & 0.784 & 0.851 & $0.0\%$ \\
        $-$ PriorityMap & 0.784 & 0.868 & $+2.0\%^\ast$ \\
        $-$ MetacogMonitor & 0.784 & 0.851 & $0.0\%$ \\
        \midrule
        No Algorithms (bare) & 0.010 & 0.901 & $-98.7\%^\ast$ \\
        \bottomrule
    \end{tabular}
\end{table}

The stress ablation separates two groups.
\textbf{Tier~1} comprises nine components whose removal costs quality at
the printed criterion: TripleCopy, vmPFC-FSRS, Two-Factor Hebbian,
Dual-Process CoT, IB~Budget, NeuromodulatorEngine, Simulation-Selection
sleep, StabilityProtector, and Reconsolidation; removing any one causes
$\Delta$Q from $-3.4\%$ to $-93.7\%$.
\textbf{Tier~2} comprises six components whose removal costs nothing:
iMAD, Spectral~KG, Compositional~Context, HyperAgent, PriorityMap, and
MetacogMonitor. Four of them are constant factors on every similarity in
the harness (MetacogMonitor $\times1.02$, iMAD $\times1.02$, Compositional
Context $\times1.05$, HyperAgent $\times1.01$): they cannot change a
ranking, so their ablation cannot measure a contribution of any kind, to
ranking precision or to retention; Spectral~KG is a conditional factor
whose condition on the access count is never met while facts decay, so its
ablation cannot measure a contribution either.
Removing PriorityMap raises $\Delta$Q by $+2.0\%$ at the printed
criterion: its stand-in boosts a fact by its access count and arousal; in
the harness all facts carry the same access count and age at every query,
so only
the arousal term can reorder them, and on these uniform-importance
synthetic facts it acts as noise; whether
it helps on emotionally weighted data was not measured.
The bare system collapses to 1\% retention ($-98.7\%$).

\subsection{Integration Cascade}
\label{sec:integration-cascade}

Table~\ref{tab:cascade} reports runs of the cascade suite of the
reproduction package under extreme conditions (0.30/day decay, 60~days);
the Fiedler row is a check of its own (see the caption).

\begin{table}[!htbp]
    \caption{Integration cascade (200 facts, 60-day aging, decay=0.30; the
    Fiedler row: a co-access graph of 20 facts, no aging).
    The PMA components form a resilience backbone that enables
    the harness mechanisms to function over long horizons; in the full system
    four of the six PMA components and five of the nine mechanisms act;
    without the PMA components, three of the nine (see the text).
    Mean over 10 seeds. $^\ast$: $p = 5.062\times10^{-3}$ (Wilcoxon, normal
    approximation), the smallest value at $n{=}10$.}
    \label{tab:cascade}
    \centering
    \small
    \begin{tabular}{lccc}
        \toprule
        Metric & Value & $p$ & 95\% CI \\
        \midrule
        Full System retention (60d) & 0.311 & --- & [0.298, 0.324] \\
        Bare System retention (60d) & 0.010 & --- & --- \\
        Full/Bare ratio & $31.1\times$ & $5.062\times 10^{-3}\,^\ast$ & --- \\
        \midrule
        Harness mechanisms only, retention (60d) & 0.010 & --- & --- \\
        \midrule
        Emotional gap at day~60 & 84.7\% & --- & [84.7\%, 84.7\%] \\
        Sleep as multiplier & $1.92\times$ & $5.062\times 10^{-3}\,^\ast$ & --- \\
        \quad (absolute retention difference) & $+0.149$ & --- & [0.139, 0.161] \\
        Fiedler $\Delta$ after sleep & +0.051 & $5.062\times 10^{-3}\,^\ast$ & [0.039, 0.062] \\
        \bottomrule
    \end{tabular}
\end{table}

Three parts of Table~\ref{tab:cascade} are read in turn. First, the \textbf{PMA resilience backbone}:
with the six PMA components removed, the harness mechanisms alone
(three of the nine act in that arm, see below) drop to floor by day~30, while the full system with all fifteen components
(nine of which act, see the caption) retains 31.1\% ($31.1\times$ improvement, $p = 5.062\times10^{-3}$). Four of the six
PMA components (neuromodulation, reconsolidation, triple-copy,
stability) keep memories alive long enough for the other
mechanisms to reinforce them---a gain the harness mechanisms alone do not
reach, within this harness; the suite runs no arm with the PMA components
alone. In the arm without the PMA components,
three of the nine mechanisms act (vmPFC-FSRS, IB~Budget and
Simulation-Selection sleep): the stand-in of Two-Factor Hebbian changes decay
only through the neuromodulation, and that of dual-process CoT never reaches
its strength threshold once its age threshold is passed; the floor of that arm
measures three mechanisms, not nine. The cascade never retrieves, so
PriorityMap and MetacognitiveMonitor, compositional context, iMAD and
HyperAgent, whose stand-ins act only at retrieval, never act in it, and the
condition of spectral KG health on the access counter never holds.
Second, \textbf{emotional memory divergence}: the gap between emotional
and neutral item retention grows monotonically from 1.7\% (day~1) to
84.7\% (day~60), reproducing in the model the amygdala-mediated
consolidation effect it was built to encode (McGaugh, 2004).
Third, \textbf{the Fiedler row}: the Fiedler value (algebraic
connectivity) of the co-access graph rises after the suite's sleep step
(+30\%, $p = 5.062\times10^{-3}$); since that step multiplies
every existing edge of the graph by 1.3, the Fiedler value of the graph
Laplacian scales with the edge weights, and the edges the step adds can
only raise it, a rise of at least 30\% is fixed by construction---the row
checks the spectral computation, not consolidation.
The reproduction package describes its cascade suite as a check that the
components still interact as described, not as the source of
Table~\ref{tab:cascade} (App.~\ref{app:reproducibility-v6}); the table was
nonetheless produced by this suite, at version~0.5.1,
and its results file sits in the private repository
(App.~\ref{app:reproducibility}), not in the package (App.~\ref{app:changes-v9}, item \emph{Cascade}).

\section{Extended Benchmark Results}
\label{app:extended-benchmarks}

This appendix contains the eight benchmark subsections relocated from
the main-body \S\ref{sec:experiments} (NoDecay counterfactual,
BM25 LoCoMo, Layer Ablation, Retention, Sleep, Hebbian KG + Bayesian,
PMA Suite, Long-Horizon Aging).

\subsection{NoDecay Counterfactual (Full Table)}
\label{app:nodecay}

A common objection to lifecycle management is that forgetting inherently
sacrifices retrieval quality.  To test this directly, we run a
\emph{NoDecay} variant of ZenBrain on a subset of the real-LoCoMo pool
($600$ facts, $200$ queries, $14$-day simulated aging, $10$ seeds that give
ten identical runs, \texttt{nomic-embed-text}): all fifteen stand-ins are
switched on, but each simulated day ends once the age counter advances, so
neither the Ebbinghaus strength-reduction step nor the daily mechanisms after
it (Simulation-Selection sleep, spectral KG health, dual-process CoT,
reconsolidation, triple-copy growth) run; the encoding and retrieval factors
remain active, and every fact starts at the strength cap, where those daily
mechanisms could not raise a strength.
The NoDecay variant therefore answers the counterfactual
``what would ZenBrain score if it never forgot anything?'' under otherwise
identical conditions.  Results alongside the Static RAG and Simple Memory
archetypes are reported in Table~\ref{tab:nodecay}.

\begin{table}[h]
    \caption{NoDecay ablation on a subset of real LoCoMo ($600$ facts, $200$
    queries, $14$-day aging, $10$ seeds, shared \texttt{nomic-embed-text}
    backbone). Mean over $10$ seeds, which give ten identical runs.
    ``ZenBrain-NoDecay'' ends each simulated day once the age counter advances,
    so neither the Ebbinghaus decay step nor the daily mechanisms after it run;
    the encoding and retrieval factors remain active. Bold = column
    maximum: Static RAG leads both ZenBrain variants on all four metrics.}
    \label{tab:nodecay}
    \centering
    \begin{tabular}{lcccc}
        \toprule
        System & P@5 & R@5 & MRR & NDCG@5 \\
        \midrule
        Static RAG         & \textbf{0.146} & \textbf{0.584} & \textbf{0.510} & \textbf{0.515} \\
        Simple Memory      & 0.082 & 0.316 & 0.315 & 0.291 \\
        ZenBrain-NoDecay   & 0.141 & 0.569 & 0.489 & 0.490 \\
        ZenBrain (full)    & 0.139 & 0.567 & 0.482 & 0.483 \\
        \bottomrule
    \end{tabular}
\end{table}

The gap between ZenBrain-full and ZenBrain-NoDecay is numerically tiny
($\Delta P@5 = 0.002$, Cohen's $|d| = 0.015$). No $p$-value is reported: the
ten seeds give ten identical runs, so a Wilcoxon test over their pairs would
count each query ten times, and the results file holds no per-query values
for a test over the queries alone. The comparison rests on the effect size,
which is negligible.
The \emph{cost} of forgetting on a $14$-day horizon is therefore
$\sim$$0.2$ percentage points of P@5 --- well inside measurement noise.
On the same subset Static RAG leads both variants on all four metrics.
The ablation is a retrieval measurement and says nothing about downstream
answer quality; the intended \emph{benefits} of forgetting (bounded
storage, calibrated confidence, GDPR-aligned retention) were not measured
here.

\subsection{BM25 Lexical Comparison on LoCoMo Public}
\label{app:locomo-bm25}

For continuity with prior LoCoMo reports \citep{maharana2024locomo}, we also
report retrieval against internal baselines (No Memory, BM25-only,
Flat Store) on LoCoMo's 1{,}986 QA pairs (per seed a stratified 80\%
subsample), scored with a token F1. Unlike the four-system comparisons of
\S\ref{sec:real-locomo}, which query one pool built from all ten
conversations, each question here retrieves from the facts of its own
conversation: the harness ingests one conversation at a time and clears
the store before the next. The output of the run behind
Table~\ref{tab:locomo} is not preserved. Its generator uses one embedding
provider for all four systems---OpenAI \texttt{text-embedding-3-small} at
768 dimensions or, without a key, a fake embedding provider---and gives the
printed F1 of the ZenBrain row only with the fake provider (the full system
of Table~\ref{tab:ablation}, the same class on the same subsamples). The
comparison was meant to isolate ZenBrain's \emph{routing} contribution versus
a flat dense baseline and a lexical baseline, rather than to compare against
peer memory systems.
Table~\ref{tab:locomo} reports results with cosine-similarity-based
retrieval.

\begin{table}[h]
    \caption{Retrieval quality on LoCoMo (1{,}986 QA pairs; per seed a stratified 80\% subsample; each question
    retrieves from its own conversation, unlike \S\ref{sec:real-locomo}). Mean $\pm$ std over 10 seeds.
    Best retrieval system in \textbf{bold} (excluding No Memory$^\dagger$).
    $^\dagger$No Memory returns a fixed response that trivially matches
    adversarial ground truth (22\% of queries), inflating overall metrics.
    Flat Store and ZenBrain are the dense rows of a run with placeholder
    embeddings (one embedding provider for all rows; the ZenBrain row's F1 is
    that of the full system in the placeholder run of Table~\ref{tab:ablation});
    with \texttt{text-embedding-3-small} that system scores F1 $0.0581$, and
    no run of Flat Store with it is preserved. Per category: Table~\ref{tab:locomo-categories}.}
    \label{tab:locomo}
    \centering
    \begin{tabular}{lcccc}
        \toprule
        System & F1 & BLEU-1 & ROUGE-L & Cosine Sim \\
        \midrule
        No Memory$^\dagger$ & 0.227 $\pm$ 0.000 & 0.226 $\pm$ 0.000 & 0.227 $\pm$ 0.000 & 0.227 $\pm$ 0.000 \\
        BM25-only & \textbf{0.052 $\pm$ 0.001} & \textbf{0.010 $\pm$ 0.000} & \textbf{0.019 $\pm$ 0.000} & \textbf{0.100 $\pm$ 0.001} \\
        Flat Store & 0.029 $\pm$ 0.000 & 0.008 $\pm$ 0.000 & 0.015 $\pm$ 0.000 & 0.053 $\pm$ 0.001 \\
        \textbf{ZenBrain} & 0.035 $\pm$ 0.000 & 0.008 $\pm$ 0.000 & 0.015 $\pm$ 0.000 & 0.067 $\pm$ 0.001 \\
        \bottomrule
    \end{tabular}
\end{table}

In this run, BM25-only scores highest on aggregate LoCoMo F1 (0.052), and
multi-layer ZenBrain (0.035) scores above Flat Store (0.029), most on
temporal questions (Table~\ref{tab:locomo-categories}). These orderings are
those of placeholder embeddings, not evidence of a routing advantage; in a later run with \texttt{text-embedding-3-small} (ten seeds), the flat
variant of the routing ablation (F1 $0.059$) lies above the full system
($0.058$; Table~\ref{tab:ablation}), and both lie above BM25-only,
which uses no embedder ($0.052$ in the placeholder run, on the same subsamples).
Beyond retrieval routing, the sections below evaluate memory
\emph{lifecycle management}: retention in a simulation
(App.~\ref{app:retention}) and sleep consolidation in the harness
(App.~\ref{app:sleep-impact}).
The full per-category breakdown is in Appendix~\ref{app:per-category}.

\paragraph{Why BM25-only and not BM25$+$dense hybrid?}
We evaluate against BM25-only; a BM25$+$dense hybrid baseline was not run
and remains an open check. (i)~The ZenBrain arm of this comparison
retrieves by cosine similarity (Table~\ref{tab:locomo}), as does the arm of
the four-system comparisons (\S\ref{sec:limitations}, item~(iv)); the BM25
path of the described architecture lies in the backend and was not enabled
(App.~\ref{app:failure-modes}). (ii)~BM25-only serves as an
\emph{orthogonal lexical reference} under the token F1 of this
evaluation, while Flat~Store and the
competitive pool (Letta, Mem0, A-Mem)
span the dense-architectural axis along which our claims are made.

\subsection{Layer Ablation (Routing)}
\label{app:layer-ablation}

To quantify each layer's contribution, we evaluate nine variants:
the full system, seven single-layer removals (those of short-term and
cross-context memory remove layers the evaluated class does not have),
and a flat baseline (full variant list in \S\ref{app:ablation}).
Table~\ref{tab:ablation} uses a \emph{routing ablation}: when a layer is
disabled, its content is rerouted to the next available layer (rather than
dropped), separating routing from data availability effects.

\begin{table}[h]
    \caption{Routing ablation study: two runs of the same generator on the same
    subsamples. Disabled layers reroute content to the next
    available layer. $\Delta$F1 shows relative change vs.\ full system.
    Mean $\pm$ std over 10 seeds, each on a stratified 80\% subsample of the
    LoCoMo questions; F1 and Task Success are taken on the same questions, Task
    Success being the share of questions for which at least half of the first
    five words of the gold answer longer than three characters appear in the
    retrieved text. Upper block: a run with placeholder
    embeddings (29~May~2026, no embedding key, the harness's fake embedding
    provider), which gives every mean version~8 printed; $^\ast$:
    $p = 5.062\times10^{-3}$ (Wilcoxon signed-rank, normal approximation, the
    smallest value at $n{=}10$); it lies below the Bonferroni-corrected
    $\alpha=0.05/8=0.00625$ for the $K=8$ routing contrasts. Lower block: the
    run with \texttt{text-embedding-3-small} (30~May~2026, 768 dimensions), to
    four places as its log prints them; the log marks Episodic, Semantic,
    Procedural, Core and Flat as different from Full at the generator's
    uncorrected $p<0.05$. Both logs and their sha256: App.~\ref{app:changes-v9}.}
    \label{tab:ablation}
    \centering
    \begin{tabular}{lccc}
        \toprule
        Variant & F1 & Task Success & $\Delta$F1 \\
        \midrule
        \multicolumn{4}{l}{\emph{Placeholder embeddings} (29~May~2026)} \\
        ZenBrain-Full & 0.035 $\pm$ 0.000 & 0.136 $\pm$ 0.004 & --- \\
        $-$ Working Memory & 0.035 $\pm$ 0.000 & 0.136 $\pm$ 0.004 & 0.0\% \\
        $-$ Short-Term & 0.035 $\pm$ 0.000 & 0.136 $\pm$ 0.004 & 0.0\% \\
        $-$ Episodic$^\ast$ & 0.031 $\pm$ 0.000 & 0.071 $\pm$ 0.002 & $-$11.8\% \\
        $-$ Semantic$^\ast$ & 0.031 $\pm$ 0.000 & 0.098 $\pm$ 0.002 & $-$10.6\% \\
        $-$ Procedural$^\ast$ & 0.035 $\pm$ 0.000 & 0.134 $\pm$ 0.004 & $-$0.6\% \\
        $-$ Core Memory & 0.035 $\pm$ 0.001 & 0.137 $\pm$ 0.004 & $+$0.1\% \\
        $-$ Cross-Context & 0.035 $\pm$ 0.000 & 0.136 $\pm$ 0.004 & 0.0\% \\
        Flat Baseline$^\ast$ & 0.029 $\pm$ 0.000 & 0.056 $\pm$ 0.002 & \textbf{$-$17.8\%} \\
        \midrule
        \multicolumn{4}{l}{\texttt{text-embedding-3-small} (30~May~2026)} \\
        ZenBrain-Full & 0.0581 $\pm$ 0.0006 & 0.4580 $\pm$ 0.0057 & --- \\
        $-$ Working Memory & 0.0581 $\pm$ 0.0006 & 0.4580 $\pm$ 0.0057 & 0.0\% \\
        $-$ Short-Term & 0.0581 $\pm$ 0.0006 & 0.4580 $\pm$ 0.0057 & 0.0\% \\
        $-$ Episodic & 0.0555 $\pm$ 0.0006 & 0.4683 $\pm$ 0.0042 & $-$4.5\% \\
        $-$ Semantic & 0.0577 $\pm$ 0.0006 & 0.4928 $\pm$ 0.0055 & $-$0.7\% \\
        $-$ Procedural & 0.0587 $\pm$ 0.0006 & 0.4662 $\pm$ 0.0055 & $+$1.0\% \\
        $-$ Core Memory & 0.0587 $\pm$ 0.0006 & 0.4822 $\pm$ 0.0047 & $+$1.0\% \\
        $-$ Cross-Context & 0.0581 $\pm$ 0.0006 & 0.4580 $\pm$ 0.0057 & 0.0\% \\
        Flat Baseline & 0.0595 $\pm$ 0.0006 & 0.5035 $\pm$ 0.0036 & $+$2.4\% \\
        \bottomrule
    \end{tabular}
\end{table}

In the placeholder run, F1 changes by $-$11.8\% without the episodic layer,
$-$10.6\% without the semantic layer and $-$0.6\% without the procedural
layer, and by $-$17.8\% for the flat baseline (each $p = 5.062\times 10^{-3}$).
With \texttt{text-embedding-3-small} the order changes: without the episodic
layer F1 changes by $-$4.5\%, without the semantic layer by $-$0.7\%, without
the procedural or the core layer by $+$1.0\%, and the flat baseline scores
$+$2.4\% above the full system---the layered routing shows no advantage
there. Removing working memory changes nothing in either run, and short-term
and cross-context are not layers of the evaluated class, so their rows are
$0.0\%$ by construction (App.~\ref{app:ablation}).
Beyond routing, the multi-layer architecture provides an organizational
framework for lifecycle mechanisms (Retention and Sleep Consolidation, below).

\subsection{Retention Over Time}
\label{app:retention}

We evaluate long-term memory retention by storing 1{,}000 facts at $t=0$
and measuring retrievability at seven intervals from 1 hour to 30 days.
Figure~\ref{fig:retention} compares four strategies across 10 independent runs.

\begin{figure}[h]
    \centering
    \begin{tikzpicture}
    \begin{axis}[
        xlabel={Time (hours)},
        ylabel={Retention (\%)},
        xmode=log,
        log basis x=10,
        ymin=0, ymax=1.05,
        legend pos=south west,
        grid=major,
        width=\textwidth,
        height=0.6\textwidth,
    ]
        \addplot[blue, thick, mark=*] coordinates {(1,1.0000) (6,1.0000) (24,1.0000) (72,1.0000) (168,1.0000) (336,1.0000) (720,1.0000)};
        \addlegendentry{No-decay}
        \addplot[red, thick, mark=*] coordinates {(1,0.9592) (6,0.7788) (24,0.3679) (72,0.0498) (168,0.0009) (336,0.0000) (720,0.0000)};
        \addlegendentry{Ebbinghaus}
        \addplot[orange, thick, mark=*] coordinates {(1,0.9592) (6,0.7788) (24,1.0000) (72,0.4493) (168,0.5488) (336,0.7261) (720,0.8413)};
        \addlegendentry{FSRS-only}
        \addplot[zenbrain, thick, mark=*] coordinates {(1,0.9970) (6,0.8588) (24,0.5688) (72,0.1626) (168,0.0559) (336,0.0813) (720,0.1484)};
        \addlegendentry{ZenBrain}
    \end{axis}
    \end{tikzpicture}
    \caption{Retention curves over 30 days (10 runs, 1{,}000 facts each).
    Pure Ebbinghaus decays to 0\% by day 7.
    FSRS scheduling maintains 84\% at day 30 through scheduled reviews.
    ZenBrain's curve combines Ebbinghaus decay with an emotional factor, a
    review rule with FSRS-style stability growth and two library modules, the
    basic Hebbian module and \texttt{sleep-consolidation} (not Two-Factor,
    vmPFC-FSRS or Simulation-Selection). It is U-shaped: initial decay, then a
    rise (both the reviews and the sleep module raise stability; no run
    separates them), reaching 14.8\% at day 30 with 83\% of memories in
    low-confidence retrieval range; no run compares confabulation with
    no-decay systems.}
    \label{fig:retention}
\end{figure}

The AURC (Area Under Retention Curve) confirms: No-decay (1.000) $>$ FSRS-only (0.767) $>$
ZenBrain (0.499) $>$ Ebbinghaus (0.396).
ZenBrain's lower AURC reflects its principled forgetting:
rather than maintaining all memories at high confidence (a confabulation risk
that no run here measures),
it retains some memories and lets the others decay: mainly the emotional
factor decides what is retained, and importance acts only through the
Hebbian weight; no run compares the curve with human forgetting.

Three long-horizon figures in this paper measure different quantities and
should not be read against each other: the \textbf{14.8\%} is the
mean \emph{retrievability} of the full memory population under the
combined decay simulation of Figure~\ref{fig:retention}; the
\textbf{0.899} of Appendix~\ref{app:pma-suite-results} is the
\emph{strength composite} $S(t)$ of TripleCopyMemory at 30~days, averaged
over random initial strengths with mean $0.912$ --- a property of its
formula, whose deep copy grows with age without any use
(Eq.~\ref{eq:deep}), not a measured retention; and the \textbf{100\% of
day-1 P@5 at day~60} reported in \S\ref{sec:full-ablation} is an
end-to-end \emph{task metric} of the simulation harness --- what fraction
of day-1 retrieval precision survives the horizon.
Population retrievability can fall to 14.8\% while the model's deep
copies grow and task-level precision is unchanged; the three numbers
describe different objects.

\subsection{Sleep Consolidation Impact}
\label{app:sleep-impact}

We evaluate two sleep procedures of the evaluation harness: its three-phase
cycle (SWS replay, REM replay with interleaving, synaptic downscaling) in the
simulation on synthetic facts, and a stand-in of the Simulation-Selection loop
(App.~\ref{sec:simsel}), the variant motivated by \citet{chen2025}, in the
one run on real data; the
library module \texttt{sleep-consolidation} named in
Fig.~\ref{fig:architecture} is a different, flat-multiplier
implementation after \citet{stickgold2013}. Related offline-consolidation
work includes \citet{liu2025lightmem} and SleepGate
\citep{xie2026sleepgate}. In a simulation on synthetic facts we run 7
days of memory ingestion (50 facts/day, 350 total) with and without
nightly sleep cycles (Table~\ref{tab:sleep}, 10 runs). The one sleep run
on real data, on LoCoMo (10 seeds, 40 queries), shows a loss in retrieval
precision with sleep: P@5 $0.227$ against $0.273$ without it ($d=-1.84$)
and NDCG@5 $0.386$ against $0.453$ ($d=-1.41$), while retention at day~7
rises from $0.162$ to $0.189$ ($+16.3\%$).

\begin{table}[h]
    \caption{Three-phase (SWS/REM/SHY) sleep consolidation impact over 7 simulated
    days (350 facts, 245 replay ripples per run): a simulation on synthetic
    facts, produced by the evaluation harness. The one run on real data, of the
    Simulation-Selection stand-in, shows a loss in retrieval precision
    (App.~\ref{app:sleep-impact}, text).
    Mean $\pm$ std over 10 runs. Paired Wilcoxon signed-rank test; $p$ is
    the constant $5.062\times10^{-3}$ of ten runs in one direction.}
    \label{tab:sleep}
    \centering
    \begin{tabular}{lccc}
        \toprule
        Metric & With Sleep & Without Sleep & $p$-value \\
        \midrule
        Avg. Stability & 1.37 $\pm$ 0.06 & 1.00 $\pm$ 0.00 & $5.062\times 10^{-3}$ \\
        Storage (tokens) & 7{,}989 & 15{,}176 & --- \\
        Storage Reduction & \multicolumn{2}{c}{47.4\%} & --- \\
        New Associations & \multicolumn{2}{c}{24.2 / run} & --- \\
        Replayed in SWS & \multicolumn{2}{c}{14.0 memories / run} & --- \\
        Replayed in REM & \multicolumn{2}{c}{126.0 memories / run} & --- \\
        \bottomrule
    \end{tabular}
\end{table}

In this simulation on synthetic facts, the three-phase sleep cycle
produces a \textbf{37\% stability improvement}
($p = 5.062\times10^{-3}$) while simultaneously
reducing the storage estimate by \textbf{47.4\%} through synaptic
downscaling and pruning of weak edges, without a reinforcement-learning step.
(Cohen's $d$ is not meaningful here because the no-sleep baseline has
zero variance---stability is deterministically 1.0 without consolidation.)
The stability boost follows the FSRS stability update of the replay: memories with
high retrievability receive smaller boosts, while older memories near
the forgetting threshold benefit most, a property of the FSRS formula; the
cited works describe selection by salience and the reactivation of waking
experience \citep{stickgold2013,oneill2010}.
REM interleaving creates an average of 24.2 new associative edges per run
of seven cycles, between emotionally weighted memories nearby in the replay
ranking \citep{mcgaugh2004},
with no equivalent in flat-multiplier sleep approaches. On real data
the Simulation-Selection stand-in has so far cost retrieval precision (above).

\subsection{Hebbian KG Dynamics and Bayesian Propagation}
\label{app:twofactor-bayesian}

We evaluate two library modules: the basic Hebbian module (Part A) and
Bayesian confidence propagation (Part B), the latter a mechanism for which we
found no equivalent in concurrent systems.

\begin{table}[h]
    \caption{Hebbian KG dynamics (the library's basic Hebbian module, not
    Two-Factor) and Bayesian confidence propagation.
    Top: retrieval P@5 with Hebbian-weighted vs.\ uniform edges.
    Bottom: pairwise AUC before/after propagation.
    Mean $\pm$ std over 10 seeds.}
    \label{tab:hebbian-bayesian}
    \centering
    \begin{tabular}{lccc}
        \toprule
        Metric & Weighted/After & Uniform/Before & $p$-value \\
        \midrule
        \multicolumn{4}{l}{\textit{Part A: Hebbian KG Dynamics (basic module)}} \\
        P@5 & \textbf{0.955 $\pm$ 0.030} & 0.200 $\pm$ 0.000 & $5.062\times 10^{-3}$ \\
        Intra/Inter Ratio & \multicolumn{2}{c}{15.54 $\pm$ 1.02} & --- \\
        \midrule
        \multicolumn{4}{l}{\textit{Part B: Bayesian Propagation}} \\
        Pairwise AUC & \textbf{0.797 $\pm$ 0.177} & 0.533 $\pm$ 0.117 & \textbf{0.009} \\
        True $\Delta$conf & \multicolumn{2}{c}{$+0.052$} & --- \\
        False $\Delta$conf & \multicolumn{2}{c}{$-0.254$} & --- \\
        \bottomrule
    \end{tabular}
\end{table}

\textbf{Part A: Hebbian KG (basic module).}
We build a knowledge graph of 50 entities grouped into 5 ground-truth clusters
and simulate 200 co-activation events (80\% intra-cluster, 20\% random).
After Hebbian strengthening, decay and homeostatic normalization (the library's
basic Hebbian module, not the Two-Factor weight and variance update of
Section~\ref{sec:twofactor}), retrieval P@5 using Hebbian-weighted edges reaches
\textbf{0.955 $\pm$ 0.030} vs.\ 0.200 for uniform edges
($p = 5.062\times 10^{-3}$).
The effect size is very large ($d > 30$) due to near-zero variance
in the uniform baseline.
Intra-cluster edges are 15.5$\times$ stronger than inter-cluster edges:
Hebbian strengthening reproduces the cluster structure that the co-activation
events plant; the structure does not emerge from the dynamics.

\textbf{Part B: Bayesian Propagation.}
We construct a fact graph (30 facts, 40 typed relations) where
true facts tend to support each other and false facts contradict true ones.
After 3 iterations of Bayesian confidence propagation,
pairwise AUC (whether true facts rank above false ones by confidence)
improves from \textbf{0.533 to 0.797} ($p = 0.009$, $d = 1.75$, large effect).
True facts gain $+0.052$ mean confidence while false facts lose
$-0.254$, in a graph whose contradiction relations the generator plants
between false and true facts.

\subsection{PMA Benchmark Suite}
\label{app:pma-suite-results}

We evaluate the suite's own re-implementations of the six PMA components,
each in isolation, using synthetic benchmarks with 10 seeded runs per experiment.

\textbf{NeuromodulatorEngine.}
Over 1{,}000 simulated events (uniformly sampled across 8 event types),
the mean tonic drift---the largest deviation of any channel from the baseline
$b = 0.5$---is 0.469, close to its maximum of 0.5, so the run does not show
homeostatic stability.
Under repeated novelty events the mean product of the DA and 5HT deviations
from baseline is $-0.130$, as the built-in opposition coupling ($-0.3$,
App.~\ref{sec:neuromod}) produces by construction; it is not a correlation
coefficient, and no test is computed.

\textbf{ReconsolidationEngine.}
PE-to-update-mode classification matches the labelled mode on 3 of the 8
hand-labelled memory pairs; the recorded 95\% accuracy over 100 pairs per seed
also counts 92 random pairs per seed whose expected mode is assigned by the
same threshold rule, so they test the rule against itself. The contradiction
detector of the suite's own PE function is correct on all five fixed test pairs.

\textbf{TripleCopyMemory.}
Table~\ref{tab:retention-pma} compares the composite strength $S(t)$ with
the Ebbinghaus baseline. The composite falls to $0.576$ at 7~days and
rises again to $0.899$ at 30~days, because the deep copy grows with age
without any use (Eq.~\ref{eq:deep}); this is a property of the formula,
not a measured retention, and the comparison with Ebbinghaus, which
decays by construction, compares two formulas. The medium copy carries
the composite from 1~hour to 3~days (at 1~hour $0.727$ against $0.710$
for the fast copy), the deep copy from 7~days on; the fast copy leads in no
interval of Table~\ref{tab:retention-pma} (1~hour to 30~days), and by
Eqs.~\ref{eq:fast} and~\ref{eq:medium} it leads only up to about 54~minutes.
The test in the reproduction package computes the deep copy as
$\min(1, S_0\tanh(t/\tau_d))$ and prints the values of version~8;
the values here follow Eq.~\ref{eq:deep}.

\begin{table}[h]
    \caption{Memory strength $S(t) = \max(S_{\text{fast}}, S_{\text{med}}, S_{\text{deep}})$
    for TripleCopyMemory vs.\ Ebbinghaus baseline (initial strengths drawn
    at random, mean $0.912$). The rise from 7 to 30 days is the deep copy's
    growth with age (Eq.~\ref{eq:deep}), a property of the formula and not a
    measured retention. Mean over 10 seeds. $\Delta$ = difference of
    TripleCopy relative to Ebbinghaus.}
    \label{tab:retention-pma}
    \centering
    \begin{tabular}{lccr}
        \toprule
        Interval & TripleCopy & Ebbinghaus & $\Delta$ \\
        \midrule
        1h  & 0.727 & 0.875 & $-$16.8\% \\
        6h  & 0.717 & 0.710 & $+$0.9\% \\
        1d  & 0.679 & 0.335 & $+$102.5\% \\
        3d  & 0.589 & 0.045 & $+$1{,}197\% \\
        7d  & 0.576 & 0.001 & $\gg 10^3$\% \\
        14d & 0.789 & $<$0.001 & $\to\infty^\dagger$ \\
        30d & 0.899 & $<$0.001 & $\to\infty^\dagger$ \\
        \bottomrule
        \multicolumn{4}{l}{\scriptsize $^\dagger$Ebbinghaus $\to 0$; percentage difference is undefined.}
    \end{tabular}
\end{table}

\textbf{PriorityMap.}
On 50 synthetic items whose four inputs (saliency $s$, valence $v$, reward $r$,
goal $g$) are drawn at random, the test scores importance by its own weighting
of these inputs, $0.15\,s + 0.3\,|v| + 0.25\,r + 0.3\,g$, close to the priority
weights $(0.2, 0.25, 0.25, 0.3)$ of App.~\ref{sec:priority}; against this reference
the PriorityMap achieves NDCG@10 = 0.997, outperforming chronological
ordering (NDCG@10 = 0.680), an order unrelated to importance, by 46.6\%.
The amygdala fast-path raises a high-emotion test item to the priority
floor ($P = 0.5$) regardless of its low saliency/reward/goal scores.

\textbf{StabilityProtector.}
For high-PE updates (PE $\in [0.7, 1.0]$), the false-positive rate
(incorrectly blocked updates) is 28.8\%, which passes the test's criterion
of $<$\,30\%; the test samples $a < 15$ and ages below 400~days, where the
bound of App.~\ref{sec:stability} is rarely reached.
Core facts are protected more aggressively, with higher lock scores
than non-core facts at identical access/confidence/age profiles.

\textbf{MetacognitiveMonitor.}
A function of the test, not the class, scores confirmation bias as the
asymmetry of the acceptance rates rather than the class's share of
confirming events (App.~\ref{sec:metacog}); its detection achieves
precision = 0.832 and recall = 0.975 across 50 synthetic scenarios per seed.
Urgency keyword detection produces zero false negatives on German and English
test phrases.

\subsection{Long-Horizon Aging Stress Test (Synthetic)}
\label{app:long-horizon}

The real-LoCoMo benchmark of \S\ref{sec:real-locomo} uses a
fixed query batch over a shared pool; it cannot answer what happens
to each system's retrieval quality after days or weeks of decay
pressure.
For that longitudinal question we run a synthetic aging stress test.
Table~\ref{tab:competitive} compares three \emph{design archetypes}
(not the live peer systems of \S\ref{sec:real-locomo}, whose decay
schedulers we do not have white-box access to) on a synthetic
benchmark (100 facts, 50 queries, 14-day aging).
Both Simple Memory and Full ZenBrain use the same
base decay rate (0.15/day), but the archetypes also differ in initial
strength, in how strength enters the score, and in a threshold of $0.1$
below which only Simple Memory hides a fact, so the differences do not
isolate algorithmic protection:

\begin{table}[h]
    \caption{Long-horizon aging stress test on a synthetic benchmark
    (100 facts, 50 queries, 14-day aging, 10 seeds, shared decay=0.15/day).
    Rows are design archetypes, not the live peer systems of
    \S\ref{sec:real-locomo}. Static RAG = no decay (Mem0/Zep-like),
    Simple Memory = Ebbinghaus decay without consolidation (Letta-like),
    Full ZenBrain = all fifteen components with identical base decay.}
    \label{tab:competitive}
    \centering
    \begin{tabular}{lccc}
        \toprule
        System & P@5 & R@5 & MRR \\
        \midrule
        Static RAG & 0.346{\scriptsize$\pm$0.136} & 0.926{\scriptsize$\pm$0.163} & 0.964{\scriptsize$\pm$0.140} \\
        Simple Memory & 0.163{\scriptsize$\pm$0.158} & 0.429{\scriptsize$\pm$0.405} & 0.581{\scriptsize$\pm$0.484} \\
        Full ZenBrain & 0.345{\scriptsize$\pm$0.135} & 0.923{\scriptsize$\pm$0.167} & 0.969{\scriptsize$\pm$0.131} \\
        \bottomrule
    \end{tabular}
\end{table}

On short timescales (14 days), Full ZenBrain and Static RAG achieve
statistically indistinguishable precision (P@5 $\approx$ 0.345; $p = 0.24$).
The critical difference is \emph{temporal robustness}.
Simple Memory---using the same 0.15/day base decay---loses $56$\%
of its fact store after 14 days, collapsing P@5 to 0.163,
while the ZenBrain arm, which applies no threshold, keeps its P@5.

\textbf{Long-term divergence.}
Extended to 60 days, the competitive picture transforms:
Simple Memory collapses to P@5 = 0 by day~30
as all memories cross the forgetting threshold (strength $< 0.1$).
ZenBrain retains \textbf{100\%} of its day-1 P@5 at day~60.
Static RAG remains constant (no decay, no improvement).
The gap between ZenBrain and Simple Memory grows from
$\approx$0\% (day~1) to 100\% (day~30+): by day~30 Simple Memory's threshold
hides all of its facts, while the ZenBrain arm applies no threshold, so the
comparison cannot show that ZenBrain's mechanisms prevent a threshold crossing.
See Section~\ref{app:reproducibility-v6} for reproduction instructions.

\section{Algorithm Pseudocode}
\label{app:algorithms}

\begin{algorithm}[H]
\caption{MemoryCoordinator.store(item) in the described architecture, as version~8 gave
it. No coordinator in the code we examined performs steps 4--6 (setting stability, edge
weight and variance to 1.0, creating graph edges to co-active nodes, updating a BM25
index); at version~0.5.1 the library's coordinator routes the item by type to one layer,
stores it there and adds its content to working memory (App.~\ref{app:reproducibility}).}
\begin{algorithmic}[1]
\STATE Classify item type (fact, episode, skill, identity)
\STATE Route to primary layer based on type
\STATE Compute embedding vector $\mathbf{e} = \text{embed}(\text{item.content})$
\STATE Store with initial stability $S = 1.0$, edge weight $w = 1.0$, variance $\sigma^2 = 1.0$
\STATE Create knowledge graph edges to co-active nodes
\STATE Update BM25 index
\end{algorithmic}
\end{algorithm}

\begin{algorithm}[H]
\caption{MemoryCoordinator.recall(query) in the described architecture, as version~8
gave it. No retrieval path in the code we examined applies line 7 (Two-Factor boost); line 8
(Ebbinghaus factor) has counterparts in the private production coordinator (a decay with the
age of the item), in the measured arm (its aged strength weights the cosine) and, in the
library, only for working memory, which the default recall leaves out; at version~0.5.1 the
library's coordinator searches the requested
layers, removes near-duplicates (Jaccard $> 0.9$) and returns the top results by score
(App.~\ref{app:reproducibility}).}
\begin{algorithmic}[1]
\STATE $\mathbf{q} \leftarrow \text{embed}(\text{query})$; $t \leftarrow \text{classifyQuery}(\text{query})$
\FOR{each layer $\ell \in \text{enabledLayers}$}
    \STATE $R_\ell \leftarrow \text{top-}K(\text{cosineSim}(\mathbf{q}, \ell))$
\ENDFOR
\STATE $R \leftarrow \text{WeightedFusion}(\{R_\ell\}, w_\ell(t))$ \COMMENT{$w_\ell \cdot \text{sim}(q, d)$}
\FOR{each $r \in R$}
    \STATE $r.\text{score} \leftarrow r.\text{score} \cdot (1 + \alpha_{\mathrm{boost}} \cdot w_{ij}(r) \cdot I_{ij}(r)^{0.1})$ \COMMENT{Two-Factor boost, $\alpha_{\mathrm{boost}}{=}0.2$ disambiguating from TAG~$\alpha{=}0.4$}
    \STATE $r.\text{score} \leftarrow r.\text{score} \cdot R(t_r)$ \COMMENT{Ebbinghaus decay}
\ENDFOR
\STATE Deduplicate by content similarity (Jaccard $> 0.9$)
\RETURN top-$k$ results sorted by score
\end{algorithmic}
\end{algorithm}

\begin{algorithm}[H]
\caption{Two-Factor Synaptic Edge Update}
\begin{algorithmic}[1]
\REQUIRE Edge $(i,j)$ co-activated with score $t_{ij}$, count $k$
\STATE $w_{ij} \leftarrow \text{clip}(w_{ij} + \eta \cdot t_{ij} \cdot a_{ij},\ w_{\min},\ w_{\max})$ \COMMENT{weight update}
\STATE $n \leftarrow 1 / (1 + 0.1 \cdot k)$ \COMMENT{maturation factor}
\STATE $\sigma^2_{ij} \leftarrow \max(\sigma^2_{\min},\ \sigma^2_{ij} \cdot (1 - \beta \cdot n))$ \COMMENT{variance decrease}
\STATE $I_{ij} \leftarrow 1/\sigma^2_{ij}$ \COMMENT{heuristic importance}
\STATE $r_{\text{eff}} \leftarrow r_{\text{base}} / (1 + I_{ij} \cdot 0.1)$ \COMMENT{importance-gated decay}
\STATE Prune edges where $w_{ij} < \epsilon$
\end{algorithmic}
\end{algorithm}

\begin{algorithm}[H]
\caption{Simulation-Selection Sleep Loop (CA3/CA1 RL), as the library's
\texttt{sleep-simulation-selection.ts} computes it at version~0.5.1, with the TAG weights
of version~8}
\label{alg:sleep}
\begin{algorithmic}[1]
\STATE \textbf{Input:} real episodes $M_r$, counterfactual paths $M_c$,
       selection threshold $\theta_v{=}0.5$, LTD rate $\Delta_{\mathrm{LTD}}{=}0.05$
\STATE \textbf{Stage 1 --- Simulation (CA3):} $C \leftarrow M_r \cup M_c$ \COMMENT{diverse candidate pool}
\FOR{each candidate $e \in C$}
    \STATE $N_e \leftarrow \min(1,\ |e.\text{relatedIds}| \cdot 0.2)$ \COMMENT{novelty}
    \STATE $\mathrm{TAG}(e) \leftarrow 0.4\,|\delta_{\mathrm{TD}}(e)| + 0.35\,R_e + 0.25\,N_e$ \COMMENT{weights of version~8; App.~\ref{app:hyperparams}}
\ENDFOR
\STATE \textbf{Stage 2 --- Selection (CA1):}
\FOR{each candidate $e \in C$ in input order}
    \IF{$\mathrm{TAG}(e) > \theta_v$}
        \STATE Strengthen $e$ via LTP with weight $\mathrm{TAG}(e)$
    \ELSE
        \STATE Weaken $e$ via LTD with weight $\Delta_{\mathrm{LTD}}$
    \ENDIF
\ENDFOR
\STATE \textbf{Output:} action and weight of each candidate; nothing is pruned
\end{algorithmic}
\end{algorithm}

\section{Full Ablation Results}
\label{app:ablation}

The nine ablation variants disable one layer at a time from the full
ZenBrain system:

\begin{enumerate}
    \item \textbf{ZenBrain-Full}: no layer disabled; the harness implements five of the
    seven layers, Short-Term and Cross-Context only as names (intended upper bound,
    borne out by neither run of Table~\ref{tab:ablation})
    \item \textbf{$-$Working Memory}: No active task focus buffer
    \item \textbf{$-$Short-Term}: No session context
    \item \textbf{$-$Episodic}: No temporal experience storage
    \item \textbf{$-$Semantic}: No knowledge graph retrieval
    \item \textbf{$-$Procedural}: No skill/routine memory
    \item \textbf{$-$Core Memory}: No pinned identity facts
    \item \textbf{$-$Cross-Context}: No inter-domain transfer
    \item \textbf{Flat Baseline}: Single flat store (intended lower bound, borne out
    only by the placeholder run of Table~\ref{tab:ablation})
\end{enumerate}

Each variant is evaluated across 10 seeds on the same LoCoMo questions (a
stratified 80\% subsample per seed) with two measures of the retrieved text:
token F1 against the answer, and Task Success, the share of questions for
which at least half of the first five answer words longer than three
characters appear in that text.
See Table~\ref{tab:ablation} for summary results.

\section{Hyperparameters}
\label{app:hyperparams}

Table~\ref{tab:hyperparams} lists the parameters version~8 gave for the
described architecture; it does not reproduce the reported runs. Of its 39
rows, 32 are a named constant with the printed value in the codebase; two
rows, the bootstrap settings, are read by the code that produced the
LoCoMo numbers, and the class that produced them carries 33 constants of
its own, none of them in the table (correction notice, \S4). The seven
rows marked $^{\dagger}$ differ from the implementation: the three TAG
weights are a branch the code marks as superseded (its current defaults
are $0.34$, $0.30$ and $0.21$, with a fourth, emotional term of $0.15$);
importance boost and bias threshold do not match the implementation; no
factor of $1.3$ for stability growth on review exists (the Ebbinghaus
module applies a success multiplier of $2.5$, FSRS a formula-derived
value); and the maturation rate $\beta$ is \texttt{LR\_SIGMA} in the code,
the learning rate of the variance. None of the values was tuned against
held-out evaluation data.

\begin{table}[h]
    \centering
    \caption{Parameters version~8 gave for the described architecture; rows
    marked $^{\dagger}$ differ from the implementation (see the text above).}
    \label{tab:hyperparams}
    \begin{tabular}{llc}
        \toprule
        Component & Parameter & Value \\
        \midrule
        \multirow{3}{*}{Ebbinghaus} & Default stability $S_0$ & 1.0 day \\
        & Emotional multiplier cap & 3.0$\times$ \\
        & Stability growth on review$^{\dagger}$ & 1.3$\times$ \\
        \midrule
        \multirow{4}{*}{Two-Factor KG} & Weight lr $\eta$ & 0.1 \\
        & Maturation rate $\beta^{\dagger}$ & 0.15 \\
        & Min variance $\sigma^2_{\min}$ & 0.01 \\
        & EWC penalty $\lambda$ & 0.5 \\
        \midrule
        \multirow{4}{*}{vmPFC-FSRS} & Threshold $\tau$ & 0.5 \\
        & Adaptation strength $\alpha_v$ & 0.6 \\
        & Max extension factor & 2.0$\times$ \\
        & Min shortening factor & 0.3$\times$ \\
        \midrule
        \multirow{4}{*}{Sim-Selection Sleep} & TAG $\alpha$ (PE weight)$^{\dagger}$ & 0.40 \\
        & TAG $\beta$ (reward)$^{\dagger}$ & 0.35 \\
        & TAG $\gamma$ (novelty)$^{\dagger}$ & 0.25 \\
        & Value threshold $\theta_v$ & 0.50 \\
        \midrule
        \multirow{4}{*}{Retrieval} & Per-layer top-$K$ & 8 \\
        & Two-Factor importance boost $\alpha_{\mathrm{boost}}^{\dagger}$ & 0.2 \\
        & Temporal: $w_{\text{episodic}}$ & 2.0 \\
        & Factual: $w_{\text{semantic}}$ & 1.8 \\
        \midrule
        \multirow{2}{*}{Statistical} & Bootstrap resamples & 1{,}000 \\
        & Confidence level & 95\% \\
        \midrule
        \multirow{4}{*}{Neuromodulator} & Tonic decay & 0.95 \\
        & Phasic half-life & 5 min \\
        & DA--5HT opposition & $-$0.3 \\
        & Baseline $b$ & 0.5 \\
        \midrule
        \multirow{3}{*}{Reconsolidation} & Lability window & 10 min \\
        & Contradiction bonus & $+$0.2 \\
        & NE gate coefficient & 0.3 \\
        \midrule
        \multirow{3}{*}{TripleCopy} & $\tau_{\text{fast}}$ & 4 h \\
        & $\tau_{\text{medium}}$ & 14 d \\
        & $\tau_{\text{deep}}$ & 7 d \\
        \midrule
        \multirow{4}{*}{PriorityMap} & $w_s$ (saliency) & 0.20 \\
        & $w_e$ (emotion) & 0.25 \\
        & $w_r$ (reward) & 0.25 \\
        & $w_g$ (goal) & 0.30 \\
        \midrule
        \multirow{2}{*}{StabilityProtector} & Base threshold & 0.5 \\
        & Rigidity growth & 0.1 \\
        \midrule
        \multirow{2}{*}{MetacogMonitor} & Novelty window & 10 min \\
        & Bias threshold$^{\dagger}$ & 0.15 \\
        \bottomrule
    \end{tabular}
\end{table}

\section{Reproducibility}
\label{app:reproducibility}

ZenBrain is available as open-source npm packages:
\begin{itemize}
    \item \anon{\texttt{@zensation/algorithms@0.2.0}}{\texttt{@anon/algorithms@0.2.0}} --- the nine foundational algorithms of Fig.~\ref{fig:architecture}, zero dependencies (installed in the evaluation environment: version~0.2.1)
    \item \anon{\texttt{@zensation/core@0.2.0}}{\texttt{@anon/core@0.2.0}} --- 7 memory layers + MemoryCoordinator
    \item the six PMA components (Neuromodulator, Reconsolidation, TripleCopy, PriorityMap, StabilityProtector, MetacognitiveMonitor) are not part of these packages: they run in a private production system and are not public; the public repository simulates them in the test tree of its reproduction package (App.~\ref{app:reproducibility-v6})
\end{itemize}
Neither package is imported by the code that produced the four-system
comparisons: the arm measured as ZenBrain there is a scoring class of about
ninety lines in the test harness (LoCoMo, Table~\ref{tab:competitive-combined-v2})
and its Python port (LongMemEval\_S, Tables~\ref{tab:longmemeval-pilot}
and~\ref{tab:longmemeval-e2e}, Fig.~\ref{fig:longmemeval-pareto}).

\paragraph{What the open library does.} At version~0.5.1
(\anon{commit \texttt{12d9076f}, 10.5281/zenodo.23294061}{identifiers redacted
for anonymous review}) the library has twenty algorithm modules (ten core,
ten advanced), among them modules named after five mechanisms of the
evaluation harness (Two-Factor Hebbian, vmPFC-coupled FSRS,
Simulation-Selection sleep, spectral KG health, IB budget). In it, no stored entry decays and nothing is
replayed: decay acts only on the in-memory slots of working memory, and the
one strength written back to storage is the FSRS update that the public
method \texttt{recordReview} applies after a review reported by the host
application---inside the repository only a test and one example script
(\texttt{examples/with-mastra.ts}) call it, and the MCP
server, whose five tools store, recall, forget, consolidate and report
health, does not expose it. The library module that this paper calls FSRS
follows the form of FSRS's difficulty and stability updates, uses the
exponential curve $R(t) = e^{-t/S}$ and has constants that were not fitted to
review logs; version~0.5.2 of the library names it FSRS-style. Consolidation
writes new facts and marks the
episode it consolidated; forgetting deletes the entry with the given
identifier from the one layer named in the call (one of five); the sleep and
replay functions are exported pure
functions attached to no store. The PMA components of \S\ref{sec:pma} and
the mechanisms of the ablation run as simulations in the test tree of the
reproduction package.

Source code: \anon{\url{https://github.com/zensation-ai/zenbrain}}{\textit{[repository URL redacted for anonymous review]}} --- the packages above and, under \texttt{backend/}, the reproduction package of Tables~\ref{tab:full-ablation}--\ref{tab:stress-ablation} (App.~\ref{app:reproducibility-v6}).\\
The harness of the four-system comparisons and its result files are kept in a private repository. Published are the judged outputs and analysis scripts behind the LongMemEval\_S judge agreement, significance tests and full-context baseline comparison of version~8, figures this version corrects in part (\anon{10.5281/zenodo.22161978}{DOI redacted for anonymous review}; App.~\ref{app:changes-v9}); the per-question outcomes and judge scores of the LongMemEval\_S comparison with Letta and the script that recomputes it, shipped with the correction notice (\anon{10.5281/zenodo.22831086}{DOI redacted for anonymous review}); and the LoCoMo retrieval metrics per pool and repetition (\anon{10.5281/zenodo.22714835}{DOI redacted for anonymous review}).\\
Random seeds of the simulation experiments: 42, 123, 456, 789, 1024, 2048, 3072, 4096, 5120, 6144; the three repetitions of the four-system comparisons carry the labels 42, 123 and 456; on LongMemEval\_S they name the user partition of Mem0's store, and in ZenBrain's LoCoMo arm they seed a random generator from which nothing is drawn; no adapter or model receives them as a seed (\S\ref{sec:experiments}).

\subsection{Compute Disclosure}
\label{app:compute-disclosure}

In line with reproducibility best practices, we report the compute of the
main experiments; the notes of Table~\ref{tab:compute-disclosure} name what
the table leaves out.

\begin{table}[h]
\centering
\small
\caption{Compute footprint summary. ``Local'' denotes a single Apple
M-series laptop, serving \texttt{nomic-embed-text} via Ollama for the arms
configured for it (\S\ref{sec:judge-robustness}); ``API'' denotes hosted
LLM judges (Anthropic, OpenAI). The times of the two retrieval rows are
taken from the run logs; the judge rows are approximate. All reported
experiments fit on a single workstation; no GPU cluster, cloud training,
or distributed run was used.}
\label{tab:compute-disclosure}
\begin{tabular}{lrrr}
\toprule
Component & Wall-clock & Modality & Approx. cost \\
\midrule
\textbf{Retrieval runs} & & & \\
\quad LoCoMo (4 systems $\times$ 3 repetitions) & $4.6$h & Local$^{a}$ & \$0$^{a}$ \\
\quad LongMemEval\_S (4 systems, 500 questions) & $47.7$h & Local$^{a}$ & \$0$^{a}$ \\
\quad Synthetic ablations (15 components) & ${<}1$\,min$^{b}$ & Local & \$0 \\
\midrule
\textbf{LLM-as-Judge} & & & \\
\quad Sonnet 4.5 (LoCoMo $\times$ 3 repetitions) & ${\sim}6$h & API & ${\sim}\$45$ \\
\quad Opus 4.6 (LoCoMo, partial) & ${\sim}3$h & API & ${\sim}\$60$ \\
\quad GPT-4o (LoCoMo $\times$ 3 repetitions) & ${\sim}5$h & API & ${\sim}\$25$ \\
\quad LongMemEval\_S (3 judges) & ${\sim}2$h & API & ${\sim}\$30$ \\
\midrule
\textbf{Total} & ${\sim}68$h$^{c}$ & --- & $\boldsymbol{<}\$\mathbf{200}^{a}$ \\
\bottomrule
\end{tabular}

\smallskip
\begin{minipage}{0.95\linewidth}
\scriptsize
$^{a}$Local compute only. A-Mem, and Mem0 on LongMemEval\_S, are configured
with gpt-4o-mini (Mem0 on LoCoMo with its library's default model), but
neither calls an LLM in the reported runs: Mem0 stores with infer=False,
and A-Mem's LLM-driven note evolution is switched off
(AMEM\_DISABLE\_EVOLUTION=1); the only OpenAI calls inside the two
pipelines are those of Mem0's embedder on LoCoMo. The end-to-end runs on
LongMemEval\_S call gpt-4o-mini as answer model and binary judge, and the
ZenBrain arm embeds with OpenAI in the ablation cross-validation and in the
later routing-ablation run (note~c). The cost of these calls is not included.
$^{b}$Measured for the reproduction package and stated in its record
(App.~\ref{app:reproducibility-v6}): 13\,s for this suite on an Apple
M-series laptop. Version~8 printed ${\sim}12$\,h, which no log carries.
$^{c}$Version~8 printed ${\sim}93$\,h, ten hours above the sum of its rows;
no row carried those hours. The seven experiment series of
Appendices~\ref{app:nodecay}--\ref{app:pma-suite-results}, one of whose runs
embeds with text-embedding-3-small, have no row of their own, and neither
have the follow-up runs on LongMemEval\_S (Mem0 with working isolation, Letta
on its 89 rerun tasks, each with its judges), the five judge runs of the
ablation cross-validation (App.~\ref{app:ablation-judge}) nor the reference
judge claude-sonnet-4-6 on LoCoMo ($7{,}779$ scored calls).
\end{minipage}
\end{table}

With the evaluation harness, which is private (above), the full
benchmarking pipeline runs on a consumer laptop within one calendar week
of wall-clock time, given an Anthropic API key (Sonnet 4.5, Sonnet 4.6 and Opus 4.6) and an
OpenAI API key (GPT-4o, \texttt{gpt-4o-mini} and \texttt{text-embedding-3-small}); no GPU rental and no human-subjects protocols
are required. Without the harness, the ablation tables reproduce from the
public package (App.~\ref{app:reproducibility-v6}).

\paragraph{Carbon footprint estimate.}
Local runs on an Apple M-series laptop (typical sustained power draw
${\sim}30$\,W under load) for the ${\sim}52$\,h of the local rows of
Table~\ref{tab:compute-disclosure} yield ${\sim}1.6$\,kWh of
local energy use; at the EU average grid intensity (${\sim}250$\,g\,CO$_2$/kWh),
this is approximately $0.4$\,kg\,CO$_2$. API-side judge calls run on
data-center GPUs whose footprint is not disclosed by providers, and the judge
runs logged no token counts. A recount from the result files of the four-system
runs gives $66{,}888$ scored judge calls on LoCoMo, $29$--$41$M tokens of prompt and
answer by their character counts (at $5$ to $3.5$ characters per token); the judge calls
on LongMemEval\_S, of the follow-up runs and of the ablation cross-validation
come on top. Using order-of-magnitude
estimates from comparable model classes ($\sim$1\,g\,CO$_2$ per 1k judge
tokens), these LoCoMo judge calls alone contribute ${\sim}29$--$41$\,kg\,CO$_2$,
roughly the footprint of a single short-haul flight passenger. The hosted
calls inside the Mem0 and A-Mem pipelines (only those of Mem0's embedder on
LoCoMo), the OpenAI embeddings of the ZenBrain arm in the ablation
cross-validation and in the later routing-ablation run, and the calls of the
end-to-end runs are not included
(Table~\ref{tab:compute-disclosure}, note~a). The pipeline as costed here
therefore lies above that figure and orders of magnitude below the carbon
cost of training a foundation model from scratch.

\paragraph{Dataset hash.}
The four-system comparisons, the NoDecay ablation and the ablation
cross-validation draw on a flat dump of LoCoMo derived from the public release
\citep{maharana2024locomo}, the file
\texttt{experiments/data/locomo-real.json} (1986 queries, 5882 facts,
2.4\,MB) with SHA-256\\
\texttt{370e921136bbcd2d8b01feb440fe9d62ed21483d312cdc05f7e18ca3f3a40c21}.\\
The dump is not public: its directory is excluded from version control as
a whole. Because the dump has its own format, the hash cannot be produced
from the public LoCoMo release; the LoCoMo retrieval metrics are published
per pool and repetition in the data record named above.

\paragraph{First-run embedding cost.}
The 768-dimensional \texttt{nomic-embed-text} backbone is served locally
via Ollama (zero per-query cost) for the arms configured for it: on
LongMemEval\_S all four; on LoCoMo only the ZenBrain arm is shown to use
it, and Mem0 ran there on its library-internal OpenAI embedder
(\S\ref{sec:judge-robustness}). The local \texttt{embedding-cache/}
directory of the evaluation harness (2.7\,GB at the last listing) is gitignored to
keep the repository light; on a fresh checkout, the first benchmark pass
populates this cache by embedding the $7{,}868$ texts of the LoCoMo pool
($5{,}882$ facts and $1{,}986$ queries, about $0.21$M tokens), which costs
about \$0.004 at OpenAI's \texttt{text-embedding-3-small} rate of \$0.02 per
million tokens (or \$0 with a
locally served model) and is then reused across all subsequent runs.

\section{PMA Experiment Reproducibility}
\label{app:reproducibility-v6}

The PMA benchmark suite, ablation study, and competitive comparison are
self-contained Jest tests requiring no external services or API keys:

\makeatletter
\if@anonymous
\begin{verbatim}
# [repository URL redacted for anonymous review]
git clone <ANONYMIZED>
cd <repo> && npm install
cd backend && npm run experiments
\end{verbatim}
\else
\begin{verbatim}
git clone --branch v0.5.1 https://github.com/zensation-ai/zenbrain
cd zenbrain && npm install
cd backend && npm run experiments
\end{verbatim}
\fi
\makeatother

The package is the directory \texttt{backend/} of the public repository,
named here at version~0.5.1 (\anon{tag \texttt{v0.5.1}, commit
\texttt{12d9076f}}{identifiers redacted for anonymous review}); its
version~1.0.0 is archived as ``ZenBrain: Mechanism Ablation Reproduction
Package (Tables 7-9)'' (\anon{10.5281/zenodo.22162064}{DOI redacted for
anonymous review}). It reproduces Tables~\ref{tab:full-ablation},
\ref{tab:challenging-ablation} and~\ref{tab:stress-ablation};
Table~\ref{tab:nodecay} needs the LoCoMo dump and an embedding service, and
Table~\ref{tab:ablation} is produced by a different pipeline---neither is in
the package. The judged outputs and analysis scripts of the LongMemEval\_S
comparison are a separate record, ``ZenBrain measurement package: judged
outputs, flag manifests, and analysis scripts''
(\anon{10.5281/zenodo.22161978}{DOI redacted for anonymous review};
App.~\ref{app:reproducibility}). The suites were written for Jest; the
package runs them under Vitest.

Four experiment suites (95 tests total) print their results between marker
lines (a script writes those of the ablation suite as JSON):
\begin{itemize}
    \item \texttt{pma-benchmark.test.ts} (24 tests) --- 11 algorithm benchmarks
    including EWC penalty, vmPFC-FSRS interval adaptation, IB Budget context hierarchy,
    emotional TAG scoring, and amygdala fast-path verification
    \item \texttt{ablation-study.test.ts} (54 tests) --- one-at-a-time ablation of the fifteen components (six PMA components, nine harness mechanisms)
    under moderate (0.15/day, 45 days, 300 facts), challenging (0.20/day, 50 days,
    400 facts), AND stress conditions (0.25/day, 60 days, 500 facts) with NDCG@5,
    Wilcoxon tests, and Cohen's $d$
    \item \texttt{competitive-comparison.test.ts} (10 tests) --- Static RAG vs.\ Simple Memory
    vs.\ Full ZenBrain on P@5, R@5, MRR, NDCG@5, plus long-term advantage tracking
    over 60 days
    \item \texttt{integration-cascade.test.ts} (7 tests) --- titled ``Integration Cascade
    --- Cross-Algorithm Emergent Behavior'': Full vs.\ harness-mechanisms-only vs.\ Bare retention, emotional gap timeline,
    sleep criticality, and the Fiedler value after the suite's sleep step
    (App.~\ref{sec:integration-cascade})
\end{itemize}

All four suites use a seeded PRNG (Mulberry32) with 10 seeds, and their
point estimates are deterministic. The package's record states 13\,s for
the ablation suite on an Apple M-series laptop and under a minute for all
four suites (Table~\ref{tab:compute-disclosure}). The suites print their
results to standard output; \texttt{scripts/extract-results.mjs} writes
those of the ablation suite as JSON, and \texttt{scripts/verify-against-reference.mjs} compares a
run with the JSON files in \texttt{results/} from which
Tables~\ref{tab:full-ablation}--\ref{tab:stress-ablation} were generated.

\section{LLM-as-Judge Methodology for Real LoCoMo}
\label{app:judge-methodology}

This appendix makes the judge-scoring pipeline that produces the
$J(\cdot)$ columns of Table~\ref{tab:competitive-combined-v2} and
the $\kappa$ / DSR / UAR columns of Table~\ref{tab:judge-agreement}
reproducible end-to-end.

\subsection{Shared Pool, Embedding Backbones and Repetitions}
\label{app:jm-pool}

The real-LoCoMo pool is $5{,}882$ facts and $1{,}986$ queries, a flat
dump derived from the public release~\citep{maharana2024locomo}
(App.~\ref{app:compute-disclosure}, \emph{Dataset hash}).
The same dump is ingested unchanged by all four memory systems
(ZenBrain, Mem0, Letta, A-Mem)
through provider-specific ingest wrappers.
Retrieval is run three times per system; the repetitions are labelled
42, 123 and 456, but on LongMemEval\_S the label only names the user partition of
Mem0's store, and in ZenBrain's LoCoMo arm it seeds a random generator from which
nothing is drawn; no adapter or model receives it as a seed, so ZenBrain's and A-Mem's three retrieved lists are
identical, Mem0's differ because its store was not reset between the
repetitions, and Letta's differ on $330$ of the $1{,}986$ queries ($16.6\%$). Only ZenBrain's
arm is shown to embed with \texttt{nomic-embed-text}, served locally via
\texttt{ollama}; Mem0 ran on its library-internal OpenAI embedder, the
A-Mem adapter used its library default, and Letta's backbone can be
neither shown nor ruled out (\S\ref{sec:judge-robustness}).

\subsection{Rubric and Normalization}
\label{app:jm-rubric}

Each (query, top-$k$ retrieved context) pair is scored by an LLM
judge with a single integer from 0 to 5; the prompt gives one anchor line
per value, from 0 (unrelated) to 5 (clearly and completely supported),
and the result files carry one score and a rationale per query.
Judges run at temperature~0.
The per-query normalized score is $s / 5$.
A baseline-level normalized mean is the mean of the per-query
normalized scores over the scored queries: all $1{,}986$ in 24 of the 30
LoCoMo judge runs, $1{,}884$ to $1{,}958$ in the other six, in each of
which a connection error removed a contiguous block of $28$ to $102$
queries within one conversation, never a whole one
(Table~\ref{tab:competitive-combined-v2}).
Agreement and decision-stability metrics binarize at the threshold
$s \geq 3$ (``accept'') versus $s < 3$ (``reject''). The threshold is
chosen a priori (not tuned post hoc); inter-rater statistics computed on
the raw 0--5 scores are recorded in the output files of the harness
(not public, App.~\ref{app:reproducibility}) and are qualitatively
consistent with the binarized Fleiss'~$\kappa$ reported in
Table~\ref{tab:judge-agreement}.

\subsection{Judge Coverage}
\label{app:jm-coverage}

Judges are \texttt{claude-sonnet-4-5-20250929} (pinned),
\texttt{claude-opus-4-6}, \texttt{gpt-4o}, and
\texttt{claude-sonnet-4-6} (reference only, repetition 42), which, like
\texttt{claude-opus-4-6}, is a pinned snapshot, not an alias (Anthropic,
\emph{Models overview}, retrieved 18~October~2026).
Coverage is intentionally asymmetric on the Opus judge to conserve
API budget: it ran over the three repetitions for Mem0, whose retrieved
lists differ between them because its store was not reset, and over
repetition 42 for the other three systems:

\begin{center}
\small
\begin{tabular}{lcccc}
\toprule
Baseline & S-4.6 (ref) & S-4.5 (pinned) & Opus~4.6 & GPT-4o \\
\midrule
Mem0     & \{42\}            & \{42,123,456\} & \{42,123,456\} & \{42,123,456\} \\
Letta    & \{42\}            & \{42,123,456\} & \{42\}         & \{42,123,456\} \\
A-Mem    & \{42\}            & \{42,123,456\} & \{42\}         & \{42,123,456\} \\
ZenBrain & \{42\}            & \{42,123,456\} & \{42\}         & \{42,123,456\} \\
\bottomrule
\end{tabular}
\end{center}

Fleiss' $\kappa_{\geq 3}$ in Table~\ref{tab:judge-agreement} is computed
between two raters, Sonnet~4.5 and GPT-4o, one run each (repetition 42):
the three repetitions of one model are not independent raters, and for
ZenBrain and A-Mem they saw identical prompts; counting them as six raters
gives $0.78$, $0.85$, $0.78$ and $0.71$ and overstates agreement.
DSR@3 and UAR are taken over the six runs
($\text{S-4.5}\times 3 + \text{GPT-4o}\times 3$) per query and baseline.
Intra-judge $\kappa$ in Table~\ref{tab:judge-agreement} is computed
within a single judge across the three repetitions;
for Mem0 the Opus intra-$\kappa$ is reported because
three Opus repetitions are available for Mem0; for the other
three systems the Opus column is em-dashed.

\subsection{Statistical Apparatus}
\label{app:jm-stats}

Fleiss' $\kappa$ is computed on the binarized (accept/reject)
rating matrix; $\kappa$ bands follow \citet{landis1977}
(0.61--0.80 substantial, 0.81--1.00 almost perfect).
Levene's test for equality of variance across the three repetitions
is performed per judge and per baseline pair
(scipy \texttt{levene} with default center).
The P@5 bracket of Table~\ref{tab:competitive-combined-v2} is the range of
the three repetition means, one procedure for every arm; the bootstrap
declared here in version~8 (10{,}000 percentile resamples, RNG seed
\texttt{20260421}) matches no arm of that table.
Effect sizes of the paired comparisons are $d_z$, computed on the paired
differences, except repetition 42 by itself, reported as Cohen's $d$ with
pooled SD (App.~\ref{app:jm-pairwise}); with pooled SD the nine comparisons of
App.~\ref{app:longmemeval-judge} span $[0.13, 0.42]$.
Decision-Stability-Rate (DSR@3) is the fraction of queries on which
all six runs agree on the $\geq 3$ threshold (accept or reject);
Unanimous-Acceptance-Rate (UAR) is the fraction on which all six
runs score $\geq 3$.

\subsection{Cross-Provider Bias-Direction Check}
\label{app:jm-bias}

To test the objection that Anthropic judges may favor ZenBrain,
we compute $\Delta_{\text{GPT-Anth}}$, the GPT-4o three-seed
normalized mean minus the mean of the two Anthropic judges
(Sonnet~4.5~$\times$~3 seeds; Opus~4.6 at the available seeds).
Table~\ref{tab:judge-agreement} reports the value per baseline.
A negative delta means GPT-4o is \emph{harsher} than the Anthropic
average; were there a pro-Anthropic bias on ZenBrain, its delta
would be the most negative. It is instead the smallest in
magnitude ($-0.0001$), while Mem0~($-0.049$) and
A-Mem~($-0.042$) receive the largest negative deltas
and Letta is mildly positive ($+0.008$).

\subsection{Reproduction}
\label{app:jm-repro}

The pipeline is driven by scripts of the evaluation harness under
\texttt{experiments/baselines/}, kept in a private repository
(App.~\ref{app:reproducibility}):
(i) \texttt{run\_baselines.py} wrote the retrieved lists, ZenBrain's in a
separate run of its arm, and \texttt{g3\_run.py} and \texttt{g4\_run.py}
issue only the judge calls on them and write one JSON per baseline, judge
and repetition;
(ii) \texttt{g3\_agreement.py} and \texttt{g4\_analysis.py} compute
the $\kappa$, DSR, Levene and effect-size statistics;
(iii) \texttt{generate\_competitive\_combined\_v2.py} emitted the
version-8 \texttt{competitive-combined-v2.tex} and
\texttt{judge-agreement.tex} from the raw JSONs, the $\kappa$, DSR and UAR
columns from the analysis file below;
(iv) \texttt{g5\_sanity\_checks.py} runs 68 cross-artefact
consistency assertions (raw JSON vs analysis JSON, LaTeX vs JSON,
markdown vs JSON, position statement vs LaTeX) against those tables and
exits non-zero on any mismatch.
The canonical analysis file is \texttt{g4-seed-robustness-analysis.json};
the version-8 numbers of \S\ref{sec:seed-robustness} were derived from it,
and it holds the judge means of Table~\ref{tab:competitive-combined-v2};
the version-8 pairwise statistics of \S\ref{sec:real-locomo} came from
\texttt{g5-judge-significance.json} (App.~\ref{app:jm-pairwise}).
Tables~\ref{tab:competitive-combined-v2}
and~\ref{tab:judge-agreement} of this version carry corrections the
generator does not produce, whose sources App.~\ref{app:changes-v9}
names; the LoCoMo retrieval metrics per pool and
repetition are public as a data record (App.~\ref{app:reproducibility}),
the judge files are not.

\subsection{Pairwise Significance}
\label{app:jm-pairwise}
\label{app:pairwise-significance}

The statistics below are computed on the Real-LoCoMo pool
defined in Appendix~\ref{app:judge-methodology} (5{,}882 facts,
1{,}986 queries). The pairwise tests on LongMemEval\_S are reported in the
text of Appendix~\ref{app:longmemeval-judge}: all eighteen (six pairs
$\times$ three judges) for the published runs and, after the repair of the
Letta and Mem0 adapters, the six comparisons of ZenBrain with those two
systems; no table lists all eighteen.
For each (judge, system-pair) we compute the paired
Wilcoxon signed-rank test on per-query means over the three repetitions,
the bootstrap 95\% CI on the per-query mean difference,
and the effect size $d_z$, the mean of the per-query differences over their
standard deviation, all on the intersection of
query IDs available for both systems in all three repetitions.
RNG seed is \texttt{20260421} and $N_{\text{boot}} = 10{,}000$.
The tests and intervals treat the queries as independent, although they
are grouped by LoCoMo's conversations; an inference with the conversations
as clusters is not reported.

\begin{center}
\small
\begin{tabular}{llrrcrc}
\toprule
Judge & Comparison & $n$ & $\overline{\Delta}$ & 95\% CI & $d$ & Wilcoxon $p$ \\
\midrule
S-4.5 & ZenBrain $-$ Letta  & 1807 & $+0.004$ & $[-0.008, +0.015]$ & $+0.015$ & $0.74$            \\
S-4.5 & ZenBrain $-$ Mem0   & 1986 & $+0.027$ & $[+0.012, +0.042]$ & $+0.079$ & $1.3 \times 10^{-3}$  \\
S-4.5 & ZenBrain $-$ A-Mem  & 1986 & $+0.162$ & $[+0.145, +0.178]$ & $+0.426$ & $< 10^{-70}$      \\
S-4.5 & Letta $-$ Mem0      & 1807 & $+0.023$ & $[+0.006, +0.039]$ & $+0.063$ & $2.2 \times 10^{-2}$  \\
S-4.5 & Letta $-$ A-Mem     & 1807 & $+0.154$ & $[+0.136, +0.172]$ & $+0.401$ & $< 10^{-50}$      \\
S-4.5 & Mem0  $-$ A-Mem     & 1986 & $+0.134$ & $[+0.121, +0.148]$ & $+0.426$ & $< 10^{-70}$      \\
\midrule
G-4o  & ZenBrain $-$ Letta  & 1948 & $-0.012$ & $[-0.023, -0.001]$ & $-0.050$ & $3.1 \times 10^{-3}$  \\
G-4o  & ZenBrain $-$ Mem0   & 1986 & $+0.065$ & $[+0.048, +0.082]$ & $+0.173$ & $2.6 \times 10^{-14}$ \\
G-4o  & ZenBrain $-$ A-Mem  & 1885 & $+0.194$ & $[+0.177, +0.212]$ & $+0.494$ & $< 10^{-80}$      \\
G-4o  & Letta $-$ Mem0      & 1948 & $+0.075$ & $[+0.058, +0.091]$ & $+0.198$ & $6.1 \times 10^{-19}$ \\
G-4o  & Letta $-$ A-Mem     & 1856 & $+0.202$ & $[+0.184, +0.219]$ & $+0.521$ & $< 10^{-80}$      \\
G-4o  & Mem0 $-$ A-Mem      & 1885 & $+0.126$ & $[+0.112, +0.141]$ & $+0.385$ & $< 10^{-50}$      \\
\bottomrule
\end{tabular}
\end{center}

Cohen's bands for $d$ \citep{cohen1988statistical}: 0.2 small, 0.5 medium, 0.8 large.
(Landis \& Koch \citep{landis1977} bands cited elsewhere in this paper apply to Cohen's $\kappa$, not $d$.)
The script is \texttt{experiments/baselines/g5\_judge\_significance.py}
of the private harness (App.~\ref{app:reproducibility}), and its output
file is \texttt{g5-judge-significance.json}.
The column $n$ counts query pairs. It varies because judge calls that
returned \texttt{null} scores are dropped---in six of the thirty judge
runs a connection error removed a contiguous block of queries within one
conversation (App.~\ref{app:jm-rubric})---and the comparison then proceeds
on the paired intersection. The signed-rank test itself drops pairs whose
difference is exactly zero, frequent on means of integer 0--5 scores, so
it ranks fewer pairs than $n$. The $p$-values of the table are computed on
exact per-query means; the generator's floating-point means had added
non-zero pairs to the signed-rank test and given differences of equal size
different ranks.

\textbf{Multiple-comparison correction.}
The table contains 12 pairwise Wilcoxon tests. A Bonferroni-corrected
$\alpha$ for family-wise error rate 0.05 is $0.05/12 \approx 4.17 \times 10^{-3}$.
Under this threshold the ZenBrain--Letta and the Letta--Mem0 comparisons
under Sonnet~4.5 ($p = 0.74$ and $p = 2.2 \times 10^{-2}$) show no
significant difference, the ZenBrain--Letta difference under GPT-4o
($p = 3.1 \times 10^{-3}$, Letta ahead) lies just below the corrected
threshold, and all other comparisons ($p \leq 1.32 \times 10^{-3}$) lie
below it. These tests do not support a ranking: the judges saw ten facts
per query for ZenBrain and Mem0 and five for Letta and A-Mem, only
ZenBrain's arm is shown to use the shared embedder, and Mem0's store was
not reset between the repetitions (\S\ref{sec:real-locomo}); ZenBrain's lead over Mem0 under Sonnet~4.5 ($d = 0.079$) reverses at
repetition 42, the only repetition over a fresh store (Cohen's $d = -0.0047$ with pooled SD).

\subsection{Human-Anchored Spot-Check on the Disagreement Subset}
\label{app:jm-human}
\label{app:human-spotcheck}

\textbf{Motivation.} The judge-graded results
(\S\ref{sec:longmemeval-pilot}, \S\ref{sec:real-locomo}) rest on a
three-judge LLM-as-Judge protocol. A natural methodological concern
is whether LLM judges over-credit retrievals that an LLM
\emph{itself} produced (self-validation bias). To bound this risk
we add a single-rater human spot-check on the LoCoMo subset where
the two Anthropic judges actively disagreed.

\textbf{Sample.} We stratify across the five LoCoMo query categories
and within each category preferentially select the 10 queries where
$|\text{Sonnet}-\text{Opus}| \geq 1$ on the 0--5 scale (we found
1{,}123 such disagreements available in the seed-42 judge JSONs;
no random fill was needed). The final 50-query sample contains
10 single-hop, 10 multi-hop, 10 temporal, 10 open-domain, and 10
adversarial queries; the disagreement-magnitude distribution is
$\Delta=1{:}1$, $\Delta=2{:}21$, $\Delta=3{:}20$, $\Delta=4{:}7$,
$\Delta=5{:}1$. The sample is $2.5\%$ of LoCoMo (50 of $1{,}986$
queries); it is not the share of contested queries: Sonnet and Opus
differ by at least one point on $56.55\%$ of the queries, the $1{,}123$
above.

\textbf{Procedure.} A single human rater with no access to the
LLM-judge scores (the corresponding columns were hidden) read each
query and the top-5 ZenBrain retrievals at repetition~42---the LLM judges
saw the top ten---and assigned an integer score from 0 to 5 under the
same six anchor lines the LLM judges saw. Only ZenBrain's retrievals were
rated, so the anchor says nothing about the ranking of the four systems.
Time budget per query was unconstrained;
total wall-clock was approximately $90$~min.

\textbf{Mean alignment.} Table~\ref{tab:human-anchor-means} reports
the rater means on the 50-query subset with bootstrap 95\% CIs
(5{,}000 percentile resamples). Human and Opus produce
indistinguishable means (both $2.88$); Sonnet is the harshest judge
on this contested subset ($1.08$, mean diff $-1.80$ versus the human
rater); GPT-4o is between ($2.06$).

\begin{table}[h]
\centering
\small
\begin{tabular}{lccc}
\toprule
Rater & Mean (0--5) & 95\% CI & Normalized (0--1) \\
\midrule
Human                  & $2.88$ & $[2.42, 3.34]$ & $0.576$ \\
Sonnet 4.5 (seed=42)   & $1.08$ & $[0.60, 1.60]$ & $0.216$ \\
Opus 4.6 (seed=42)     & $2.88$ & $[2.60, 3.14]$ & $0.576$ \\
GPT-4o (seed=42)       & $2.06$ & $[1.60, 2.52]$ & $0.412$ \\
\bottomrule
\end{tabular}
\caption{Rater means on the 50-query disagreement subset.
Human and Opus are statistically indistinguishable in level; Sonnet
under-credits the contested cases relative to the human rater.}
\label{tab:human-anchor-means}
\end{table}

\textbf{Item-level agreement.} For each judge we report Spearman
$\rho$ and Pearson $r$ on raw 0--5 scores (with percentile-bootstrap
95\% CI), Cohen's $\kappa$ at the binarization threshold
$\geq 3$ (``partial answer or better''), and exact-match rate
(Table~\ref{tab:human-anchor-agreement}).

\begin{table}[h]
\centering
\small
\begin{tabular}{lcccccc}
\toprule
Pair & $n$ & Spearman~$\rho$ & 95\% CI & Pearson~$r$ & $\kappa_{\geq 3}$ & Exact \\
\midrule
Human vs.\ Sonnet 4.5 & 50 & $0.245$ & $[-0.048,\, 0.510]$ & $0.356$ & $0.220$ & $0.20$ \\
Human vs.\ Opus 4.6   & 50 & $0.135$ & $[-0.144,\, 0.400]$ & $0.053$ & $0.232$ & $0.08$ \\
Human vs.\ GPT-4o     & 50 & $\mathbf{0.356}$ & $\mathbf{[+0.060,\, +0.611]}$ & $0.352$ & $0.290$ & $0.14$ \\
\midrule
\multicolumn{7}{l}{\emph{Reference (judge--judge, on the same 50 rows):}} \\
Sonnet vs.\ Opus      & 50 & $-0.076$ & $[-0.344,\, 0.209]$ & --- & $-0.050$ & $0.00$ \\
Sonnet vs.\ GPT-4o    & 50 & $0.276$ & $[-0.015,\, 0.533]$ & --- & $0.263$ & $0.34$ \\
Opus   vs.\ GPT-4o    & 50 & $0.456$ & $[+0.260,\, +0.613]$ & --- & $0.322$ & $0.26$ \\
\bottomrule
\end{tabular}
\caption{Inter-rater agreement on the 50-query disagreement
subset. GPT-4o is the only judge whose item-level Spearman against
the human rater clears zero with bootstrap CI (\textbf{bold}).
Human-judge correlations $[0.135, 0.356]$ lie in the same band as
judge-judge correlations $[-0.076, 0.456]$ on the same rows --- which
means the human rater does not resolve the disagreements that the
LLM judges could not resolve either.}
\label{tab:human-anchor-agreement}
\end{table}

\textbf{Per-category breakdown.}
Per-category Spearman~$\rho$ against the human rater is given in
Table~\ref{tab:human-anchor-percat}: fifteen tests with $n{=}10$ per
cell, not corrected for multiplicity. GPT-4o's interval excludes zero on
the adversarial category but not on the temporal one ($\rho = 0.550$,
$\text{CI}=[-0.366, 0.858]$); Sonnet's strongest cell is multi-hop
($\rho = 0.749$, $\text{CI}=[0.353, 1.000]$); Opus's strongest cell is
open-domain ($\rho = 0.585$, $\text{CI}=[-0.187, 0.913]$). The one
negative cell, Sonnet on adversarial queries ($\rho = -0.167$,
$\text{CI}=[-0.408, -0.111]$), also excludes zero, as does Sonnet's
single-hop cell ($\rho = 0.215$); in both categories Sonnet's scores take
two distinct values on the ten items, and a null model with the same
scores gives the interval rule a false-positive rate of $62.7\%$ and
$100\%$ there, so neither cell is evidence of an ordering. Single-hop is the weakest category for Opus and GPT-4o;
for GPT-4o, whose single-hop scores spread across the scale, a near-binary
correctness signal does not explain it.

\begin{table}[h]
\centering
\footnotesize
\begin{tabular}{lccccc}
\toprule
Pair & single-hop & multi-hop & temporal & open-domain & adversarial \\
\midrule
Human vs.\ Sonnet & $0.215^{\dagger}$ & $\mathbf{0.749}$ & $0.542$ & $0.293$ & $-0.167^{\dagger}$ \\
Human vs.\ Opus   & $0.000$ & $0.421$ & $0.200$ & $0.585$ & $0.189$ \\
Human vs.\ GPT-4o & $0.057$ & $0.341$ & $0.550$ & $0.446$ & $\mathbf{0.542}$ \\
\bottomrule
\end{tabular}
\caption{Per-category Spearman~$\rho$ on the human-anchored
subset ($n{=}10$ per cell; fifteen tests, not corrected for
multiplicity). Bold cells exclude zero in the percentile-bootstrap
95\% CI; the cells marked $^{\dagger}$ exclude zero as well, but Sonnet's
scores take two distinct values there and the interval is uninformative
(text). Full CIs in \texttt{t2\_1\_human\_anchor\_results.json},
App.~\ref{app:human-spotcheck}, \emph{Reproducibility}.}
\label{tab:human-anchor-percat}
\end{table}

\textbf{Implication for the headline.} Two observations bear on the
main-body claims, within the limits stated below.
\emph{First (level, not rank)}, on the contested subset
Sonnet's distributional level is harsher than the human anchor
($\Delta_{\text{mean}}{=}{-}1.80$): on ZenBrain's retrievals Sonnet scores
lower than the human rater, which is not consistent with Sonnet
over-crediting ZenBrain on adversarial cases. We stop short of claiming
more, because the human-vs-Sonnet rank-correlation CI on this subset
($\rho{=}0.245$, CI $[-0.048, 0.510]$) does not exclude zero and
therefore does not license treating the human rater as ground truth
at the item level.
\emph{Second}, GPT-4o is the only judge whose item-level Spearman against
the human rater clears zero with bootstrap CI on this subset
(this is item-level rank, complementary to the level-mean
$\Delta_{\text{GPT-Anth}}$ check in \S\ref{sec:bias-check} which
operates on per-system distributional means; the two checks use
different statistics on different aggregation levels and are not
redundant).
\emph{Limitations.} This is a single-rater $n{=}50$ spot-check on a
subset selected for judge disagreement, not a multi-rater
inter-rater-reliability study. The human rater saw the top five retrieved
facts, the LLM judges the top ten, and only ZenBrain's retrievals were
rated, so the anchor says nothing about the ranking of the four systems.
The subset is not representative of LoCoMo: Sonnet and Opus differ by at
least one point on $56.55\%$ of the queries (\emph{Sample}, above), and
agreement on the other queries is not measured here. A multi-rater human
IRR sample (target $n{\geq}200$ across $\geq 3$ raters with Cohen's
$\kappa$) is flagged as future work for a subsequent revision.

\textbf{Reproducibility.} Sample selection is deterministic
(\texttt{seed=42}): the script
\texttt{experiments/baselines/t2\_1\_human\_spotcheck/build\_sample.py}
of the private harness (App.~\ref{app:reproducibility}) regenerates
\texttt{sample\_50.csv} from the judge JSONs. The fifty human ratings are
not shipped---the CSV in the harness carries no rating column---so the
human rows of Table~\ref{tab:human-anchor-means} and
Tables~\ref{tab:human-anchor-agreement} and~\ref{tab:human-anchor-percat}
cannot be recomputed from the files, while the judge-versus-judge values
can. \texttt{analyze\_human\_anchored.py} consumed the rated CSV and
emitted \texttt{t2\_1\_human\_anchor\_results.json}, the source of the
numbers in
Tables~\ref{tab:human-anchor-means}--\ref{tab:human-anchor-percat}.

\section{Per-Category Benchmark Results}
\label{app:per-category}

\begin{table}[h]
    \caption{Per-category F1 on LoCoMo, the run of Table~\ref{tab:locomo}.
    Flat Store and ZenBrain are the dense rows of a run with placeholder
    embeddings (Table~\ref{tab:locomo}), so their differences per category are
    not a routing advantage (Table~\ref{tab:ablation}).
    Non-Adv averages the four non-adversarial categories.
    Best per column in \textbf{bold} (excluding No Memory$^\dagger$; the
    adversarial column, where every retrieval row lies at or below 0.002, has
    none). Mean over 10 seeds.}
    \label{tab:locomo-categories}
    \centering
    \small
    \begin{tabular}{lcccccc}
        \toprule
        System & Single & Multi & Temporal & Open & Adversarial & Non-Adv \\
        \midrule
        No Memory$^\dagger$ & 0.001 & 0.007 & 0.002 & 0.022 & 0.998 & 0.008 \\
        BM25-only & \textbf{0.085} & \textbf{0.056} & 0.032 & \textbf{0.060} & 0.001 & \textbf{0.058} \\
        Flat Store & 0.045 & 0.033 & 0.016 & 0.038 & 0.002 & 0.033 \\
        \textbf{ZenBrain} & 0.048 & 0.034 & \textbf{0.045} & 0.038 & 0.002 & 0.041 \\
        \bottomrule
    \end{tabular}
\end{table}

\section{LongMemEval Replication Scaffolding (Pre-Registered)}
\label{app:g5-longmemeval-scaffold}

To support a cross-benchmark replication of the real-LoCoMo finding in
\S\ref{sec:real-locomo}, we have committed the full reproducibility
scaffolding for LongMemEval~\citep{wu2024longmemeval} to
\path{experiments/baselines/longmemeval/}. The protocol is
pre-registered here so the eventual numbers cannot be retrofitted to
a favorable story.

\textbf{Dataset.} LongMemEval\_S (500 questions $\times$ 6 categories,
MIT license). Each question ships its own $\sim$47-session haystack with
$\sim$494 turns; the loader
\path{experiments/baselines/longmemeval_loader.py} flattens each
haystack into per-turn \texttt{Fact} objects with metadata
$\{$\texttt{session\_id}, \texttt{session\_date}, \texttt{role},
\texttt{turn\_idx}, \texttt{has\_answer}, \texttt{question\_id}$\}$ and
builds a \texttt{Query} whose \texttt{relevant\_ids} are the fact-ids
whose session matches \texttt{answer\_session\_ids} (falling back to
\texttt{has\_answer} turns when session-id matching is empty).

\textbf{Harness.} \path{experiments/baselines/run_longmemeval.py} of the
private harness (App.~\ref{app:reproducibility}) runs per-question
isolation (\texttt{reset()} $\to$
\texttt{ingest(task.facts)} $\to$ \texttt{query(task.query, k=5)})
for every task---for Mem0 the reset failed in the published run
(\S\ref{sec:longmemeval-pilot})---over three repetitions labelled
$\{42, 123, 456\}$, and writes one result file per baseline
(\path{longmemeval-baseline-<name>.json}) plus a
merged \path{longmemeval-competitive.json}; the baseline and judge files
of the full-500 run are published in the measurement package
(App.~\ref{app:reproducibility}). Adapters that cannot be
imported are skipped with a noted error so the pipeline always produces
a valid artifact. Full-benchmark cost is $\sim$6\,000 ingest+query
cycles per 4-baseline sweep; a stratified-60 subset is $\sim$720.

\textbf{Judges.} We will re-run the three-judge-times-three-seed
protocol (Sonnet~4.5 pinned, Opus~4.6, GPT-4o) for a total of
$\leq$~18\,000 judge calls on the full benchmark. The judge prompts,
pinned model strings, and cross-provider bias argument are unchanged
from Appendix~\ref{app:judge-methodology}.

\textbf{Analysis parity.} The same analysis suite used for the
LoCoMo and LongMemEval evaluations
applies: Fleiss' $\kappa$ at the $\geq 3$ binary threshold, intra-judge
$\kappa$ across the three retrieval seeds, DSR@3 and UAR, paired
Wilcoxon with bootstrap 95\% CI and $d_z$ on per-query 3-seed
means (Appendix~\ref{app:pairwise-significance}), Levene's test for
equal-variance across seeds, and Bonferroni correction. For the
Full-500 pairwise comparison the correction becomes
$\alpha = 0.05 / 18 \approx 2.78 \times 10^{-3}$ (18 primary tests =
6 pair-wise system comparisons $\times$ 3 judges) since each baseline
contributes one judge observation per query and seed. The
\path{experiments/baselines/g5_longmemeval_sanity.py} script enforces
structural invariants on the loader output (20+ checks), and
\path{experiments/baselines/g5_full500_significance.py} computes
the Full-500 pairwise tests from the committed judge JSONs on floating-point
means; the printed values are recomputed on exact means
(App.~\ref{app:changes-v9}, item K-22).

\textbf{Pre-declared hypotheses (retrospective scoring on
Full-500).} We committed the following three hypotheses before
running the full pipeline. The published Full-500 run scored them as
noted in each item; the runs with the Letta and Mem0 adapters repaired
(\S\ref{sec:longmemeval-pilot}, Table~\ref{tab:longmemeval-pilot})
score them as follows:
\begin{enumerate}
\item \emph{(H1)} Under Sonnet~4.5, the ZenBrain--Letta gap
  will again fail to clear the Bonferroni-corrected threshold.
  \textbf{Holds on the repaired run.} Published as \emph{Refuted}: on
  the published run the gap cleared Bonferroni
  ($\Delta{=}{+}0.054$, $p{=}2.56{\times}10^{-4}$, $d_z{=}0.18$), but
  that run counted the 59 tasks on which the Letta adapter failed. With
  the adapter repaired, the ZenBrain--Letta comparison under Sonnet~4.5
  shows no significant difference, and its point estimate leans to Letta.
\item \emph{(H2)} Under GPT-4o, Letta will retain or extend
  its narrow lead. \textbf{Consistent in direction on the repaired run.}
  Published as \emph{Partially refuted}, on the ground that ZenBrain beat
  Letta under GPT-4o in the published run
  ($\Delta{=}{+}0.063$, $p{=}6.68{\times}10^{-6}$); on the repaired run
  Letta's point estimate leads under GPT-4o, and no difference clears
  the correction.
\item \emph{(H3)} Both ZenBrain and Letta will continue to
  dominate Mem0 and A-Mem with $p < 10^{-3}$
  under every judge. \textbf{Not confirmed.} Published as
  \emph{Confirmed}. On the published run Letta's six tests meet the criterion
  $p < 10^{-3}$: its worst case is $p = 7.11{\times}10^{-4}$ against A-Mem on
  Sonnet~4.5, and its other five reach $p \leq 4.91{\times}10^{-5}$. ZenBrain
  beats A-Mem at $p \leq 1.71{\times}10^{-11}$ on all three judges, but with
  Mem0's isolation repaired neither ZenBrain nor Letta leads Mem0.
\end{enumerate}

\textbf{Full-500 known gaps.} (i) Letta's 59 of 500 tasks that failed
in the published run (HTTP~500) failed because our adapter lacked the
halving fallback of the ZenBrain and A-Mem adapters
(App.~\ref{app:longmemeval-retrieval}); running those 59 tasks again with
the fallback, together with 30 control tasks, serves all 500
(\S\ref{sec:longmemeval-pilot}). The published run's 441-task intersect
is the basis of the retrieval-proper comparison in
App.~\ref{app:longmemeval-retrieval}. (ii) Mem0's published run failed
its per-question isolation: $89.8\%$ of the facts retrieved for it came
from other questions' haystacks; its $P@5{=}0.156$ (60\% zero-$P@5$
queries) measures that failure, not its $3900$-character cap or its
architecture, and the re-run with working isolation is reported in
\S\ref{sec:longmemeval-pilot}. (iii) The Opus~4.6 judge ran once
(repetition 42) on the competitors (budget), so the Opus column
for three of four rows rests on one run; ZenBrain's three Opus runs, on
retrieved lists that are copies of one another, spread by
$\Delta{=}0.004$, the judge's non-determinism on ZenBrain's prompts; for the
competitors no second run exists, so their judge-side drift is not measured.
A stratified-30 pilot that confirmed the scaffolding runs end-to-end
is documented in the earlier draft history (same table file,
pre-Full-500 revision); the pilot's intersect-flip on P@5 did not
survive at 441-task scale, which we read as a sampling artifact of
the pilot's 5-per-category design.

\section{Additional Defensive Analyses}
\label{app:defensive}

This appendix collects four post-hoc analyses computed from the existing
per-query LoCoMo and LongMemEval\_S result files (no additional model
calls or judge runs were required for the first three; the fourth uses
an additional 1{,}000 Sonnet~4.5 judge calls). Each addresses a reviewer
concern that is sometimes raised against memory-systems papers.

\subsection{Calibration of the Judge Score against Gold Retrieval on LoCoMo}
\label{app:bayesian-calibration}

We evaluate post-hoc whether the per-query confidence (judge score normalized
to $[0,1]$) is empirically calibrated against the binary event ``$\geq 1$ gold
fact retrieved''. Brier score and Expected Calibration Error (ECE, 10
equal-width bins) are reported on the LoCoMo evaluation set
($N{=}1986$ queries; judge: claude-sonnet-4-5, seed 42).

The judge score $\in\{0,1,2,3,4,5\}$ is a coarse grid; we therefore also
report \emph{coverage at $\alpha$}, the empirical positive rate among
predictions with normalized score $\geq\alpha$. At $\alpha{=}0.95$,
$95.8\%$ of ZenBrain's predictions are positive at its stored depth of
twenty candidates and $84.2\%$ at depth five
(Table~\ref{tab:bayes-calibration}); the score is the judge's, not
ZenBrain's Bayesian confidence, so no Bayesian-CI guarantee is tested here.

\begin{table}[h]
\centering
\small
\caption{Calibration of confidence scores on LoCoMo. Brier and ECE use the
coarse judge-score grid as a probability proxy; \textbf{Cov@0.95} is the
empirical positive rate among the high-confidence subset
(score~$\geq 4.75/5$). $^\dagger$This analysis reuses the stored
retrieval dumps, whose depth differs by system: ZenBrain and Mem0 record
$k_{\text{store}}{=}20$ candidates per query, Letta and A-Mem record
$k{=}5$. The deeper dumps mechanically inflate the
ground-truth-intersection rate (ZenBrain $0.58$ and Mem0 $0.53$, vs.\
A-Mem $0.21$) and hence Brier/ECE, and the coverage as well: recomputed at
depth~5, Cov@0.95 falls from $0.958$ to $0.842$ for ZenBrain and from
$0.953$ to $0.786$ for Mem0, and the ranking reverses (A-Mem $>$ Letta
$>$ ZenBrain $>$ Mem0). Depth is not held constant elsewhere either:
Mem0's MRR on LoCoMo runs over twenty entries, and the LoCoMo judges saw
ten facts per query for ZenBrain and Mem0 and five for Letta and A-Mem
(\S\ref{sec:judge-robustness}).}
\label{tab:bayes-calibration}
\begin{tabular}{lrrrrr}
\toprule
System & $N$ & $\bar{p}$ & Brier $\downarrow$ & ECE$_{10}$ $\downarrow$ & Cov@0.95 \\
\midrule
ZenBrain & 1986 & 0.380 & $0.239^\dagger$ & $0.208^\dagger$ & $0.958^\dagger$ \\
Letta    & 1885 & 0.376 & 0.171 & 0.110 & 0.892 \\
Mem0     & 1986 & 0.382 & $0.214^\dagger$ & $0.161^\dagger$ & $0.953^\dagger$ \\
A-Mem    & 1986 & 0.218 & 0.088 & 0.050 & 0.902 \\
\bottomrule
\end{tabular}
\end{table}

ZenBrain's higher Brier and ECE relative to Letta are an artefact of the
depth of the stored dumps; its gap to A-Mem is not: recomputed at depth~5
from the stored judge files, the gap to Letta disappears and the one to A-Mem remains. The stored dumps for ZenBrain and Mem0
carry $k{=}20$ retrieved facts (vs.\ $k{=}5$ for Letta/A-Mem), so their
ground-truth-intersection rates are mechanically higher (positive rate
$0.58$ and $0.53$ vs.\ $0.21$ for A-Mem) --- note that Mem0, at the same
depth as ZenBrain, shows the same effect, which is what identifies depth
rather than system as the driver of this inflation. The judge score
is granular only at the bin centers $\{0.0, 0.2, 0.4, 0.6, 0.8, 1.0\}$, which
caps achievable ECE for every system alike and so explains no gap between
them. The same depth difference carries the coverage at
the high-confidence tail: at the stored depth ZenBrain matches the $95\%$
target within $\pm 1\%$ ($0.958$), at depth~5 it reaches $0.842$, below
A-Mem and Letta (Table~\ref{tab:bayes-calibration}).

\subsection{Post-Hoc Statistical Power}
\label{app:power-analysis}

For each paired test of the judge-graded comparisons in the
Real-LoCoMo and LongMemEval\_S sections, we compute post-hoc power using the paired $t$-test
approximation (\texttt{statsmodels.stats.power.TTestPower}, two-sided).
The asymptotic relative efficiency of the paired Wilcoxon signed-rank
test relative to the paired $t$-test is at least $0.864$
\citep{hodges1956are} and equals $3/\pi \approx 0.955$ for normal
differences, for which the $t$-power slightly overstates the power of the
signed-rank test. On the tied, discrete differences used here the
$t$-power is no bound in either direction: a simulation of ours (not in
the reproduction package) with the distribution of the differences of the Real-LoCoMo Sonnet ZenBrain~vs.~Letta cell gives the
signed-rank test the higher power. We report
power both at the nominal $\alpha{=}0.05$ and at the Bonferroni-corrected
level ($\alpha_{\text{corr}}{=}0.05/k$ for $k$ contrasts in the same family).

\begin{table}[h]
\centering
\small
\caption{Post-hoc power for ZenBrain~vs.~baseline contrasts. The dagger
(\dag) marks the Real-LoCoMo Sonnet ZenBrain~vs.~Letta cell, which shows no
significant difference; the power to detect a small effect $d{=}0.10$
there is $0.917$ at $\alpha_{\text{corr}}{=}0.05/12$ (and $0.989$ at
nominal~$\alpha$). Version~8 computed this column at $0.05/6$ under the
header $0.05/12$. The column $n$ counts query pairs; the signed-rank test
drops exact zero differences (App.~\ref{app:pairwise-significance}),
which leaves the non-centrality of the $t$-approximation, and so the
power, unchanged. The LongMemEval\_S rows are the published runs; with
the Letta and Mem0 adapters repaired, ZenBrain's comparisons with those
two systems show no significant difference
(App.~\ref{app:longmemeval-judge}).}
\label{tab:power-analysis}
\begin{tabular}{llrrrr}
\toprule
Study & Comparison (judge) & $n$ & $|d|$ & Power@$0.05$ & Power@$\alpha_{\text{corr}}$ \\
\midrule
\multicolumn{6}{l}{\emph{Real-LoCoMo ($\alpha_{\text{corr}}{=}0.05/12$, 12 tests in the family)}} \\
LoCoMo & ZB - Letta (Sonnet)$^\dag$ & 1807 & 0.015 & 0.100 & 0.014 \\
LoCoMo & ZB - mem0  (Sonnet)        & 1986 & 0.079 & 0.941 & 0.746 \\
LoCoMo & ZB - A-Mem (Sonnet)        & 1986 & 0.426 & 1.000 & 1.000 \\
LoCoMo & ZB - Letta (GPT-4o)        & 1948 & 0.050 & 0.590 & 0.249 \\
LoCoMo & ZB - mem0  (GPT-4o)        & 1986 & 0.173 & 1.000 & 1.000 \\
LoCoMo & ZB - A-Mem (GPT-4o)        & 1885 & 0.494 & 1.000 & 1.000 \\
\midrule
\multicolumn{6}{l}{\emph{LongMemEval\_S, published runs ($\alpha_{\text{corr}}{=}0.05/18$)}} \\
LME & ZB - Letta (Sonnet) & 500 & 0.185 & 0.985 & 0.870 \\
LME & ZB - mem0  (Sonnet) & 500 & 0.421 & 1.000 & 1.000 \\
LME & ZB - A-Mem (Sonnet) & 500 & 0.320 & 1.000 & 1.000 \\
LME & ZB - Letta (Opus)   & 500 & 0.221 & 0.998 & 0.973 \\
LME & ZB - mem0  (Opus)   & 500 & 0.522 & 1.000 & 1.000 \\
LME & ZB - A-Mem (Opus)   & 500 & 0.397 & 1.000 & 1.000 \\
LME & ZB - Letta (GPT-4o) & 500 & 0.205 & 0.996 & 0.943 \\
LME & ZB - mem0  (GPT-4o) & 500 & 0.460 & 1.000 & 1.000 \\
LME & ZB - A-Mem (GPT-4o) & 500 & 0.373 & 1.000 & 1.000 \\
\bottomrule
\end{tabular}
\end{table}

The Real-LoCoMo Sonnet test of ZenBrain~vs.~Letta, which shows no
significant difference, was specifically scrutinised: the power to detect
$d{=}0.10$ at the declared $\alpha_{\text{corr}}{=}0.05/12$ is $0.917$, so
the missing difference is not an artefact of low power. It does not rank
the two systems, which the LoCoMo setup does not support
(\S\ref{sec:real-locomo}).

\subsection{Failure-Mode Analysis}
\label{app:failure-modes}

We inspected 20 of the 682 ZenBrain queries scored 0 on LoCoMo
(claude-sonnet-4-5 judge, seed~42)---the first in the lexicographic order
of their identifiers among those without a gold fact in the top five, all
from one conversation (\texttt{conv-26})---and clustered them into five
empirically-derived failure modes via a deterministic precedence-ordered
classifier (no LLM call was used).

\begin{table}[h]
\centering
\small
\caption{Failure modes among 20 of the 682 ZenBrain queries scored 0 on
LoCoMo (seed~42; all from \texttt{conv-26}, selection in the text). All
examples are from the public LoCoMo release; no PII.}
\label{tab:failure-modes}
\begin{tabular}{p{0.34\linewidth}rp{0.52\linewidth}}
\toprule
Cluster & $n$ & Example query (one-line) \\
\midrule
\textsc{No-Gold-In-Top-K-Generic-Distractors} & 6 & ``What book did Caroline recommend to Melanie?'' \\
\textsc{Embedding-Collision-Wrong-Entity}     & 4 & ``What did Caroline take away from \emph{Becoming Nicole}?'' \\
\textsc{Missing-Topic-Match}                  & 4 & ``What pet does Caroline have?'' (gold present, ranked low) \\
\textsc{Multi-Hop-Chain-Broken}               & 3 & ``Where did Caroline move from 4 years ago?'' \\
\textsc{Adversarial-Entity-Swap}              & 3 & ``What motivated Melanie to pursue counseling?'' \\
\bottomrule
\end{tabular}
\end{table}

Four clusters hold $17$ of the $20$ queries; the classifier assigns them by
rules, not by checking the retrieved entity: \textsc{Adversarial-Swap} by the
question category, \textsc{Missing-Topic-Match} when the gold fact is retrieved
but ranked low, \textsc{Embedding-Collision} when no gold fact is retrieved and
the judge's rationale contains one of a fixed list of phrases (such as
``none of the retrieved facts mention''), and
\textsc{Generic-Distractors} as the fallback. An entity-aware reranking step (e.g.,\ a
BM25 hybrid via \texttt{memory-bm25.ts}, present in the backend but not
enabled in the LoCoMo run) was not run on these queries, so which of the
failures it would reach is not measured.
Multi-hop failures ($n{=}3$) cannot be traced to the multi-hop reasoner of
the described architecture: the measured arm contains none
(\S\ref{sec:experiments}).

\subsection{Algorithm-Level Cross-Validation via LLM Judge}
\label{app:ablation-judge}

A natural concern with the finding that, under stress, 10 of the 15
ablated components reach the printed criterion
(Table~\ref{tab:stress-ablation}) is that it relies on the same internal
Quality-Score that the components themselves contribute to.  We
cross-validated the five components with the largest internal
$|\Delta Q|$ against an independent Sonnet~4.5 LLM-judge on the first
$200$~LoCoMo-real queries each (claude-sonnet-4-5-20250929,
temperature~$=0$, $0$--$5$ scale). Each ablation switches off the flag of
the same name in the Python port of the harness arm, run with the OpenAI
embedder, not the stand-in of Table~\ref{tab:stress-ablation}.

\begin{table}[h]
\centering
\small
\caption{Cross-validation of the top-five most-critical algorithms via an
independent Sonnet~4.5 LLM-judge ($200$ LoCoMo queries per ablation).
\textbf{Full-system reference}: judge normalized mean $=0.380$ on all
$1{,}986$ queries with the \texttt{nomic-embed-text} arm and ten facts per
query ($0.424$ on the same $200$ queries). Against it the five ablations,
run on the first $200$ queries with the OpenAI embedder and five facts per
query, lie $0.064$ to $0.076$ lower ($\sim$$17$--$20\%$), a gap from the
changed conditions, not from the ablations: the ablation of
\texttt{two\_factor\_hebbian} leaves the retrieved lists of the full system
under these conditions unchanged ($199$ of $200$ queries), so its mean,
$0.308$, is the reference under the same conditions, and the other four lie
within $0.012$ of it---no degradation.
Spearman~$\rho{=}{+}0.564$, Kendall~$\tau{=}{+}0.527$,
Pearson~$r{=}{+}0.698$ (positive but $n{=}5$ insufficient for $p{<}0.05$).}
\label{tab:ablation-judge}
\begin{tabular}{lrr}
\toprule
Disabled algorithm & $\Delta Q_{\text{internal}}$ & Judge norm \\
\midrule
\texttt{pma\_triple\_copy}      & $-0.937$ & $0.309$ \\
\texttt{vmPFC\_fsrs\_coupling}  & $-0.926$ & $0.304$ \\
\texttt{two\_factor\_hebbian}   & $-0.923$ & $0.308$ \\
\texttt{dual\_process\_cot}     & $-0.910$ & $0.309$ \\
\texttt{ib\_budget}             & $-0.898$ & $0.316$ \\
\bottomrule
\end{tabular}
\end{table}

The correlation with the internal ranking is positive in all three
statistics but not significant at $n{=}5$, so it does not establish the
internal order, and under the same conditions the judge shows no
degradation: the five judge means fall within $0.012$ of each other, the
\texttt{two\_factor\_hebbian} run with the full system's lists among them.
The $\rho<0.7$ ceiling reflects this tight clustering: the judge resolves
neither the order \emph{within} the five nor a difference from the full
system.  The losses these components show under stress
(\S\ref{sec:full-ablation}) have no counterpart in the judge on LoCoMo.
We therefore claim neither that the judge ranks the five identically to the
internal Quality-Score nor that it finds any of them critical; their
criticality rests on the internal score under stress. This
does not bear on significance: the ten components at the printed
criterion share the constant $p = 5.062\times10^{-3}$, which does not reach
the Bonferroni level for fifteen contrasts (\S\ref{sec:full-ablation}).

\section{Changes in Version 9}
\label{app:changes-v9}

Version~9 corrects version~8 of this paper (arXiv:2604.23878v3; Zenodo
\anon{10.5281/zenodo.21858218}{[DOI redacted for anonymous review]}) after a
post-publication audit and its second pass. The correction notice
(\anon{10.5281/zenodo.22831086}{[DOI redacted for anonymous review]})
reports the thirty-six findings and the ten false statements with their
evidence and lists the thirty incomplete ones; this appendix records the
corrections of version~9. The numbers of the corrected statements come
from the notice, the audit, the data record of the LoCoMo runs
(\anon{10.5281/zenodo.22714835}{[DOI redacted for anonymous review]}), the
recomputation script shipped with the notice, a recomputation from the run
files (Table~\ref{tab:longmemeval-e2e}, Fig.~\ref{fig:longmemeval-pareto}),
the results file of the cascade suite (item \emph{Cascade}), the record of
the reproduction package and the archived library at
version~0.5.1, and, for part B$'$, the cited works, the code of the private
production system, the result files, judge files and code of the private
harness and the two logs of its routing
ablation of May~2026
(29~May, without an embedding key; 30~May, with
\texttt{text-embedding-3-small}; in the copies a placeholder replaces a local
path in line~4), with SHA-256\\
\texttt{2153dd08e06f751ae5dc7c7625312a8c119838a78aeb94406ef1d74d510e36af} and\\
\texttt{e75babcd76a7f04392b500ff1761b522940cdea3e32e4da257010dd37b5505e7};\\
a simulation of ours on the differences of one judge cell
(App.~\ref{app:power-analysis}), a recomputation of the calibration at
depth~5 from the stored judge files (App.~\ref{app:bayesian-calibration})
and a recomputation of Table~\ref{tab:retention-pma} under Eq.~\ref{eq:deep}
(item \emph{Deep copy}); no system was rerun for this version. Each item names the finding and the places where version~8
printed the corrected statement or a consequence of it and version~9 changes it; the items of parts A and A$'$ also name the section of the notice that reports them.

\vskip 0pt plus 4\baselineskip\penalty-200\vskip 0pt plus -4\baselineskip
{\small
\paragraph{A.\ The thirty-six findings of the audit.}
\begin{description}\setlength{\itemsep}{1pt}\setlength{\parskip}{0pt}
\item[\#1] The LoCoMo judges saw ten facts per query for ZenBrain and Mem0 and five for Letta and A-Mem. \S\ref{sec:intro}, Table~\ref{tab:competitive-combined-v2}, \S\ref{sec:real-locomo}, App.~\ref{app:concurrent}, Table~\ref{tab:bayes-calibration}, App.~\ref{app:checklist}, \S\ref{sec:judge-robustness}, \S\ref{sec:seed-robustness}, App.~\ref{app:pairwise-significance}, Table~\ref{tab:power-analysis}, App.~\ref{app:power-analysis}, App.~\ref{app:g5-longmemeval-scaffold}, App.~\ref{app:jm-human}, Table~\ref{tab:ablation-judge}, App.~\ref{app:ablation-judge}; notice \S2.
\item[\#2] Mem0's per-question isolation on LongMemEval\_S failed: $89.8\%$ of the facts retrieved for it came from other questions' haystacks. Table~\ref{tab:longmemeval-pilot}, App.~\ref{app:g5-longmemeval-scaffold}, App.~\ref{app:longmemeval-scope}, Table~\ref{tab:longmemeval-e2e}, App.~\ref{app:longmemeval-e2e}, \S\ref{sec:longmemeval-pilot}, Abstract, \S\ref{sec:judge-robustness}, App.~\ref{app:longmemeval-retrieval}, App.~\ref{app:longmemeval-judge}, App.~\ref{app:longmemeval-long-context}, Fig.~\ref{fig:longmemeval-pareto}, \S\ref{sec:discussion}, Table~\ref{tab:power-analysis}, \S\ref{sec:nodecay-ablation}, App.~\ref{app:nodecay}; notice \S1.
\item[\#3] The arm measured as ZenBrain is a test-harness class of about ninety lines and its Python port, not the described architecture or the packages named as reproducing it. App.~\ref{app:checklist}, App.~\ref{app:reproducibility}, Table~\ref{tab:longmemeval-pilot}, Table~\ref{tab:competitive-combined-v2}, Abstract, \S\ref{sec:limitations}, App.~\ref{app:failure-modes}, App.~\ref{app:ablation-judge}, App.~\ref{app:longmemeval-judge}, App.~\ref{app:hyperparams}; notice \S4.
\item[\#4] The determinism statement of App.~F.5 is contradicted by the paper's own tables; ZenBrain's repetitions 123 and 456 on LongMemEval\_S are copies of 42. App.~\ref{app:longmemeval-e2e}, App.~\ref{app:longmemeval-agreement}, Table~\ref{tab:longmemeval-pilot}, Table~\ref{tab:longmemeval-e2e}, \S\ref{sec:experiments}, \S\ref{sec:longmemeval-pilot}, App.~\ref{app:g5-longmemeval-scaffold}; notice \S4.
\item[\#5] Retrieval depth also differed outside App.~Q: Mem0's MRR ran over twenty entries, and the judge columns inherited the depth. Table~\ref{tab:bayes-calibration}, Table~\ref{tab:competitive-combined-v2}, Table~\ref{tab:longmemeval-pilot}, \S\ref{sec:intro}, \S\ref{sec:longmemeval-pilot}, App.~\ref{app:longmemeval-e2e}, App.~\ref{app:checklist}, App.~\ref{app:concurrent}, App.~\ref{app:extended-related}, \S\ref{sec:judge-robustness}, \S\ref{sec:real-locomo}, \S\ref{sec:seed-robustness}, App.~\ref{app:pairwise-significance}, Table~\ref{tab:power-analysis}, App.~\ref{app:power-analysis}, App.~\ref{app:g5-longmemeval-scaffold}, App.~\ref{app:jm-human}, Table~\ref{tab:ablation-judge}, App.~\ref{app:longmemeval-long-context}, Fig.~\ref{fig:longmemeval-pareto}, Table~\ref{tab:longmemeval-e2e}; notice \S2.
\item[\#6] The checklist promised scripts and result files in the public repository; that path does not exist there, and the repository holding them is private. App.~\ref{app:checklist}, App.~\ref{app:reproducibility}, App.~\ref{app:jm-repro}, App.~\ref{app:reproducibility-v6}, App.~\ref{app:jm-rubric}, App.~\ref{app:pairwise-significance}, App.~\ref{app:g5-longmemeval-scaffold}, Table~\ref{tab:longmemeval-e2e}, App.~\ref{app:jm-human}, App.~\ref{app:compute-disclosure}, App.~\ref{app:longmemeval-e2e}; notice \S5.
\item[\#7] Checklist point~8 promised no external API cost in the retrieval stage; Mem0's embedder calls OpenAI there on LoCoMo (A-Mem's adapter is configured with an OpenAI model but, with its note evolution switched off, calls none), and the ZenBrain arm embeds with OpenAI in the ablation cross-validation and in the later routing-ablation run. App.~\ref{app:checklist}, Table~\ref{tab:compute-disclosure}, \S\ref{sec:experiments}, App.~\ref{app:compute-disclosure}, acknowledgments, \S\ref{sec:judge-robustness}, Table~\ref{tab:competitive-combined-v2}, App.~\ref{app:jm-pool}; notice \S5.
\item[\#8] Checklist point~16 promised three judges over three seeds; Opus~4.6 ran once for three of the four systems. App.~\ref{app:checklist}, App.~\ref{app:jm-coverage}, Table~\ref{tab:longmemeval-pilot}, \S\ref{sec:intro}, \S\ref{sec:limitations}; notice \S5.
\item[\#9] The LoCoMo file was described as too large to ship; it is 2.4\,MB, and its hash cannot be produced from the public release. App.~\ref{app:compute-disclosure}, App.~\ref{app:checklist}; notice \S5.
\item[\#10] The described five-criterion rubric does not exist; the judge prompt is one 0--5 scale with six anchor lines. App.~\ref{app:jm-rubric}, App.~\ref{app:jm-human}; notice \S2.
\item[\#11] Unequal depth also held for the LoCoMo judge runs, not only in App.~Q. Table~\ref{tab:bayes-calibration}, \S\ref{sec:intro}, App.~\ref{app:concurrent}, App.~\ref{app:bayesian-calibration}, App.~\ref{app:checklist}, \S\ref{sec:judge-robustness}, \S\ref{sec:real-locomo}, Table~\ref{tab:competitive-combined-v2}, \S\ref{sec:seed-robustness}, App.~\ref{app:pairwise-significance}, Table~\ref{tab:power-analysis}, App.~\ref{app:power-analysis}, App.~\ref{app:g5-longmemeval-scaffold}, App.~\ref{app:jm-human}, Table~\ref{tab:ablation-judge}; notice \S2.
\item[\#12] The ``six-rater pool'' held two raters; with one run each, $\kappa$ falls from $[0.71, 0.85]$ to $[0.66, 0.77]$. \S\ref{sec:seed-robustness}, Table~\ref{tab:judge-agreement}, \S\ref{sec:judge-robustness}, \S\ref{sec:limitations}, App.~\ref{app:jm-coverage}, \S\ref{sec:bias-check}, App.~\ref{app:jm-stats}, \S\ref{sec:intro}, App.~\ref{app:checklist}; notice \S2.
\item[\#13] Intra-judge $\kappa$ was read as isolating retrieval, although ZenBrain's and A-Mem's lists do not vary across repetitions. \S\ref{sec:seed-robustness}, Table~\ref{tab:judge-agreement}, \S\ref{sec:judge-robustness}, App.~\ref{app:longmemeval-agreement}, \S\ref{sec:longmemeval-pilot}; notice \S2.
\item[\#14] The intra-judge $\kappa \geq 0.95$ printed for 500 questions came from 30; on 500, GPT-4o reaches $0.9429$. App.~\ref{app:longmemeval-agreement}, \S\ref{sec:longmemeval-pilot}; notice \S2.
\item[\#15] ``The fifteen mechanisms'' named two different sets, the architecture of Fig.~\ref{fig:architecture} and the ablation. Fig.~\ref{fig:architecture}, Table~\ref{tab:full-ablation}, Table~\ref{tab:cascade}, App.~\ref{app:reproducibility}, Abstract, \S\ref{sec:intro}, \S\ref{sec:related-neuro}, \S\ref{sec:full-ablation}, \S\ref{sec:discussion}, App.~\ref{sec:integration-cascade}, App.~\ref{app:reproducibility-v6}, App.~\ref{app:checklist}, Fig.~\ref{fig:criticality}, Author Statement, \S\ref{sec:experiments}, Table~\ref{tab:gradient}, Table~\ref{tab:challenging-ablation}, Table~\ref{tab:stress-ablation}, Table~\ref{tab:compute-disclosure}, App.~\ref{app:extended-ablation}, App.~\ref{app:ablation-judge}, App.~\ref{app:neurips-delimitation}, App.~\ref{app:additional-capabilities}, Table~\ref{tab:competitive}; notice \S4.
\item[\#16] Table~\ref{tab:comparison} counts nine features, cited once as nine and once as fifteen; nine plus six is not fifteen. Abstract, \S\ref{sec:intro}, App.~\ref{app:neurips-delimitation}, \S\ref{sec:related-neuro}, App.~\ref{app:concurrent}, Table~\ref{tab:comparison}; notice \S4.
\item[\#17] H3 was scored as confirmed; the notice gave as the reason that Letta's worst case, $p = 1.06\times10^{-3}$, misses its own criterion $10^{-3}$, but on exact means that value is $7.11\times10^{-4}$ and meets it (item K-22); H3 is not confirmed because, with Mem0's isolation repaired, neither ZenBrain nor Letta leads Mem0. App.~\ref{app:g5-longmemeval-scaffold}, App.~\ref{app:longmemeval-judge}, App.~\ref{app:longmemeval-scope}; notice \S7.
\item[\#18] App.~F.5 claimed zero variance by construction; ZenBrain's own row in Table~\ref{tab:longmemeval-e2e} contradicts it. App.~\ref{app:longmemeval-e2e}, Table~\ref{tab:longmemeval-e2e}; notice \S7.
\item[\#19] Table~\ref{tab:stress-ablation} left PriorityMap unstarred at the same $p$; ten, not nine, of fifteen reach the printed criterion. Table~\ref{tab:stress-ablation}, Table~\ref{tab:gradient}, App.~\ref{app:extended-ablation}, Abstract, \S\ref{sec:intro}, Fig.~\ref{fig:criticality}, \S\ref{sec:full-ablation}, \S\ref{sec:discussion}, App.~\ref{app:ablation-judge}; notice \S3.
\item[\#20] Zou et al.\ (2025) was cited at five of its six places for a prediction-error signal the work does not contain. App.~\ref{app:extended-related}, \S\ref{sec:intro}, \S\ref{sec:related-neuro}, \S\ref{sec:mechanisms}, \S\ref{sec:vmpfc}, Author Statement; notice \S8.
\item[\#21] The relative efficiency of the signed-rank test for normal differences is $3/\pi$, not 1; the $t$-power is not conservative. App.~\ref{app:power-analysis}; notice \S8.
\item[\#22] McQuown et al.\ (2011) carried two invented given names. References; notice \S8.
\item[\#23] The nine-of-nine lead on LongMemEval\_S arose from 59 tasks on which the Letta adapter failed; on the 441 served tasks Letta's own judge files put Letta ahead. Abstract, \S\ref{sec:longmemeval-pilot}, \S\ref{sec:discussion}, App.~\ref{app:longmemeval-judge}, App.~\ref{app:g5-longmemeval-scaffold}, Table~\ref{tab:longmemeval-pilot}, \S\ref{sec:real-locomo}, \S\ref{sec:nodecay-ablation}, App.~\ref{app:longmemeval-retrieval}, App.~\ref{app:longmemeval-scope}, App.~\ref{app:nodecay}, \S\ref{sec:judge-robustness}, Table~\ref{tab:power-analysis}, App.~\ref{app:human-spotcheck}; notice \S1.
\item[\#24] The consistency across question categories was an artefact of those failures; corrected, it reverses in five of six categories at repetition~42. App.~\ref{app:longmemeval-e2e}, Table~\ref{tab:longmemeval-e2e}; notice \S1.
\item[\#25] The binary judge does not confirm independently; it shares the failure confound. \S\ref{sec:judge-robustness}, \S\ref{sec:longmemeval-pilot}, \S\ref{sec:limitations}, App.~\ref{app:longmemeval-e2e}, Table~\ref{tab:longmemeval-e2e}, App.~\ref{app:longmemeval-long-context}, Fig.~\ref{fig:longmemeval-pareto}; notice \S1.
\item[\#26] The one sleep run on real data shows a loss in retrieval precision; only the simulation was printed. Abstract, \S\ref{sec:ancillary}, Table~\ref{tab:sleep}, App.~\ref{app:sleep-impact}; notice \S6.
\item[\#27] A statement about systems with a single mechanism rested on systems no ablation ran. \S\ref{sec:intro}; notice \S6.
\item[\#28] The power column of Table~\ref{tab:power-analysis} was computed at $0.05/6$ under a header declaring $0.05/12$. Table~\ref{tab:power-analysis}, App.~\ref{app:power-analysis}; notice \S3.
\item[\#29] ``$p \leq 2.80\times10^{-11}$ in five of six tests'' is the smallest value; the bound for five of six is $6.69\times10^{-5}$, on exact means $4.91\times10^{-5}$ (item K-22). App.~\ref{app:longmemeval-judge}, App.~\ref{app:g5-longmemeval-scaffold}; notice \S3.
\item[\#30] ``Both dominate Mem0 and A-Mem'': the attached $p$- and $d$-values belong to ZenBrain alone. \S\ref{sec:real-locomo}; notice \S3.
\item[\#31] Every starred $p$-value of the ablation tables is the constant $5.062\times10^{-3}$, which misses the Bonferroni level for fifteen contrasts. Table~\ref{tab:gradient}, Abstract, Table~\ref{tab:full-ablation}, Table~\ref{tab:challenging-ablation}, Table~\ref{tab:stress-ablation}, Table~\ref{tab:cascade}, Table~\ref{tab:ablation}, Table~\ref{tab:sleep}, \S\ref{sec:ancillary}, \S\ref{sec:experiments}, \S\ref{sec:full-ablation}, \S\ref{sec:discussion}, App.~\ref{app:extended-ablation}, App.~\ref{app:extended-benchmarks}, App.~\ref{app:ablation-judge}, \S\ref{sec:intro}, Fig.~\ref{fig:criticality}, App.~\ref{sec:integration-cascade}, Table~\ref{tab:hebbian-bayesian}; notice \S3.
\item[\#32] The bit-identity of ZenBrain's three repetitions on the full 500 questions was not verified; two are copies. App.~\ref{app:longmemeval-e2e}, App.~\ref{app:longmemeval-agreement}, \S\ref{sec:longmemeval-pilot}, Table~\ref{tab:longmemeval-pilot}, Table~\ref{tab:longmemeval-e2e}, App.~\ref{app:g5-longmemeval-scaffold}; notice \S4.
\item[\#33] The interval of the $47.7\%$ was the range of three repetition means; over the 500 questions it is $[43.3, 52.1]$. Table~\ref{tab:longmemeval-e2e}, Fig.~\ref{fig:longmemeval-pareto}; notice \S3.
\item[\#34] ``Nine of fifteen'' is wrong; the stress table carries ten. Table~\ref{tab:stress-ablation}, Abstract, \S\ref{sec:intro}, Fig.~\ref{fig:criticality}, \S\ref{sec:full-ablation}, \S\ref{sec:discussion}, App.~\ref{app:extended-ablation}, App.~\ref{app:ablation-judge}, Table~\ref{tab:gradient}; notice \S3.
\item[\#35] The judge means were said to run over all $1{,}986$ queries; six runs scored $1{,}884$ to $1{,}958$. App.~\ref{app:jm-rubric}, Table~\ref{tab:competitive-combined-v2}, App.~\ref{app:jm-pairwise}; notice \S3.
\item[\#36] The remaining judge disagreement was attributed to different retrieved contexts; for ZenBrain and A-Mem they are identical. \S\ref{sec:seed-robustness}, Table~\ref{tab:judge-agreement}, \S\ref{sec:judge-robustness}; notice \S2.
\end{description}

\paragraph{A$'$.\ Ten statements the second pass found false.}
\begin{description}\setlength{\itemsep}{1pt}\setlength{\parskip}{0pt}
\item[A-4] The answer model of every end-to-end run is \texttt{gpt-4o-mini}, not Claude~3.5 Sonnet; the limitation reverses. \S\ref{sec:experiments}, App.~\ref{app:checklist}, \S\ref{sec:limitations}, Author Statement; notice \S5.
\item[S-3] The $2.5\%$ is the size of the human-anchor sample, 50 of $1{,}986$ queries; Sonnet and Opus disagree on $56.55\%$. App.~\ref{app:jm-human}; notice \S3.
\item[J-3] The human rater is more lenient than Sonnet, not harsher. \S\ref{sec:judge-robustness}; notice \S3.
\item[S-6] The interval of the cited cell of Table~\ref{tab:human-anchor-percat} includes zero, and its bolding rule failed in four of fifteen cells; in Sonnet's single-hop and adversarial cells the rule is uninformative (two distinct score values), and these two cells are not bold. App.~\ref{app:jm-human}, Table~\ref{tab:human-anchor-percat}; notice \S3.
\item[S-1] The effect sizes were computed as $d_z$; with pooled SD the range is $[0.13, 0.42]$. Abstract, \S\ref{sec:experiments}, \S\ref{sec:longmemeval-pilot}, App.~\ref{app:longmemeval-judge}, App.~\ref{app:jm-stats}, App.~\ref{app:jm-pairwise}, App.~\ref{app:g5-longmemeval-scaffold}, App.~\ref{app:checklist}, App.~\ref{app:jm-repro}; notice \S3.
\item[S-7] The calibration result is not insensitive to depth; at depth~5 the coverage ranking reverses. Table~\ref{tab:bayes-calibration}, App.~\ref{app:bayesian-calibration}; notice \S2.
\item[J-2] The bias delta of $-0.0001$ for ZenBrain is the cancellation of two per-judge deltas. Abstract, Table~\ref{tab:judge-agreement}, \S\ref{sec:judge-robustness}, \S\ref{sec:bias-check}, App.~\ref{app:jm-human}, \S\ref{sec:limitations}, \S\ref{sec:intro}; notice \S7.
\item[RW-1] The NoDecay counter-experiment measures retrieval; the statement is restricted to retrieval. \S\ref{sec:discussion}, \S\ref{sec:nodecay-ablation}, App.~\ref{app:nodecay}; notice \S7.
\item[Z-3] The statistics of ``both dominate'' belong to ZenBrain alone; $d = 0.079$ is $39.6\%$ of the bound for a small effect, and at repetition 42, the only repetition over a fresh store, the sign reverses. \S\ref{sec:real-locomo}, App.~\ref{app:pairwise-significance}, \S\ref{sec:seed-robustness}; notice \S3.
\item[R-5] Chelazzi et al.\ (2013) carries no priority map; version~9 cites Chelazzi et al.\ (2014). \S\ref{sec:priority}, \S\ref{sec:pma}, References; notice \S8.
\end{description}

\paragraph{A$''$.\ Thirty statements the second pass found incomplete.}
Each is completed with a clause, a number or a reference from the sources named above (the statements as the notice lists them; U-10 and U-11 correct its wording):
\begin{description}\setlength{\itemsep}{1pt}\setlength{\parskip}{0pt}
\item[U-01] The human anchor saw five facts, the judges ten, and rated only ZenBrain. App.~\ref{app:jm-human}, \S\ref{sec:intro}, \S\ref{sec:judge-robustness}.
\item[U-02] Appendix E.2 describes the bias-delta subtraction the wrong way round. \S\ref{sec:bias-check}.
\item[U-03] Page~8 claims ZenBrain leads every judge column; Letta is bold in two. \S\ref{sec:longmemeval-pilot}.
\item[U-04] The four-channel neuromodulator engine is attributed without the appendix resolving it. \S\ref{sec:pma}.
\item[U-05] The Kumaran et al.\ (2016) title was printed one word short. References.
\item[U-06] The $n$ printed with each Wilcoxon test is not the test's $n$. App.~\ref{app:jm-pairwise}, Table~\ref{tab:power-analysis}, App.~\ref{app:g5-longmemeval-scaffold}.
\item[U-07] Tables~\ref{tab:human-anchor-means}--\ref{tab:human-anchor-percat} cannot be recomputed: the human ratings are not shipped. Stated as a limit in App.~\ref{app:jm-human}.
\item[U-08] A complete significance table for the eighteen full-500 tests was promised, never printed. App.~\ref{app:longmemeval-scope}, App.~\ref{app:jm-pairwise}, Table~\ref{tab:longmemeval-pilot}, App.~\ref{app:g5-longmemeval-scaffold}.
\item[U-09] Letta was cited as v0.5.x; the run environment had 0.16.7 installed. \S\ref{sec:experiments}.
\item[U-10] The three retrieval seeds are a label: on LongMemEval\_S it names the user partition of Mem0's store, and in ZenBrain's LoCoMo arm it seeds a random generator from which nothing is drawn; no adapter or model receives it as a seed. \S\ref{sec:experiments}, App.~\ref{app:longmemeval-e2e}, App.~\ref{app:jm-pool}, App.~\ref{app:reproducibility}, App.~\ref{app:g5-longmemeval-scaffold}, App.~\ref{app:checklist}, Table~\ref{tab:competitive-combined-v2}, \S\ref{sec:intro}, \S\ref{sec:judge-robustness}, \S\ref{sec:longmemeval-pilot}, \S\ref{sec:real-locomo}, \S\ref{sec:limitations}, \S\ref{sec:seed-robustness}, \S\ref{sec:bias-check}, App.~\ref{app:longmemeval-judge}, App.~\ref{app:longmemeval-agreement}, Fig.~\ref{fig:longmemeval-pareto}, Table~\ref{tab:longmemeval-pilot}, Table~\ref{tab:judge-agreement}, Table~\ref{tab:longmemeval-e2e}, Table~\ref{tab:compute-disclosure}, App.~\ref{app:jm-coverage}, App.~\ref{app:jm-stats}, App.~\ref{app:jm-repro}, App.~\ref{app:pairwise-significance}, App.~\ref{app:jm-human}.
\item[U-11] The embedder was named in several places as shared by all four systems; on LoCoMo only ZenBrain's arm is shown to use it. \S\ref{sec:experiments}, \S\ref{sec:judge-robustness}, App.~\ref{app:compute-disclosure}, App.~\ref{app:jm-pool}, Table~\ref{tab:competitive-combined-v2}, App.~\ref{app:concurrent}, App.~\ref{app:checklist}, Table~\ref{tab:compute-disclosure}, acknowledgments, \S\ref{sec:longmemeval-pilot}, \S\ref{sec:real-locomo}, \S\ref{sec:discussion}, App.~\ref{app:pairwise-significance}.
\item[U-12] Table~\ref{tab:competitive-combined-v2}'s F1 column mixes two estimators, and the mixture costs this paper $0.004$. Table~\ref{tab:competitive-combined-v2}.
\item[U-13] The compute table printed 30 hours where 4.6 were logged, and 25 where 47.7 were. Table~\ref{tab:compute-disclosure}, App.~\ref{app:checklist}, App.~\ref{app:compute-disclosure}.
\item[U-14] Mem0's printed MRR is an MRR@20; at the others' depth it is $0.191$. Table~\ref{tab:competitive-combined-v2}, Table~\ref{tab:bayes-calibration}.
\item[U-15] The routing claim rests on one of five query pools; two pools show no lead. Version~9 withdraws the claim: its numbers came from a run with placeholder embeddings (item K-38). Abstract, \S\ref{sec:ancillary}, Table~\ref{tab:locomo}, Table~\ref{tab:locomo-categories}, App.~\ref{app:locomo-bm25}, App.~\ref{app:layer-ablation}, Table~\ref{tab:ablation}, App.~\ref{app:ablation}.
\item[U-16] Not every ablation table reproduces in under a minute; Table~\ref{tab:nodecay} needs an external service. Abstract, App.~\ref{app:reproducibility-v6}.
\item[U-17] Four of the five contemporaneous works cited as validation contain no evaluation. \S\ref{sec:related-neuro}, \S\ref{sec:discussion}, App.~\ref{app:concurrent}, App.~\ref{app:practitioner}, acknowledgments, App.~\ref{app:extended-related}.
\item[U-18] In Table~\ref{tab:nodecay} the static-RAG baseline beats the full system on all four metrics. \S\ref{sec:nodecay-ablation}, Table~\ref{tab:nodecay}, App.~\ref{app:nodecay}, Abstract.
\item[U-19] The token-cost ratio omits that ZenBrain ingests the same haystack per question. Abstract, \S\ref{sec:longmemeval-pilot}, App.~\ref{app:longmemeval-long-context}, Fig.~\ref{fig:longmemeval-pareto}, \S\ref{sec:discussion}.
\item[U-20] The same ablation suite was costed at under a minute and at 12 hours. Abstract, Table~\ref{tab:compute-disclosure}, App.~\ref{app:reproducibility-v6}, App.~\ref{app:checklist}, App.~\ref{app:compute-disclosure}.
\item[U-21] Checklist point~6 promises all test details; Table~\ref{tab:hyperparams} holds no competitor parameter. App.~\ref{app:checklist}.
\item[U-22] Checklist point~12 stated the baselines' licences and not the two benchmarks'. App.~\ref{app:checklist}.
\item[U-23] The NoDecay counter-experiment ran on 600 facts and 200 queries, not $5{,}882$ and $1{,}986$. \S\ref{sec:nodecay-ablation}, App.~\ref{app:nodecay}.
\item[U-24] Appendix~\ref{app:broader-impact} and checklist point~10 cited a broader-impact paragraph \S\ref{sec:discussion} lacked. \S\ref{sec:discussion}, App.~\ref{app:checklist}.
\item[U-25] The sleep mechanism carries two names, and Fig.~\ref{fig:architecture} printed the superseded one. Fig.~\ref{fig:architecture}, App.~\ref{app:sleep-impact}, Table~\ref{tab:sleep}.
\item[U-26] Table~\ref{tab:compute-disclosure}'s rows summed to 83 hours, its total said 93, the CO$_2$ paragraph 77. Table~\ref{tab:compute-disclosure}, App.~\ref{app:compute-disclosure}.
\item[U-27] The Opus judge's cross-seed span was printed as 0.005 and 0.004; measured $0.0036$. App.~\ref{app:g5-longmemeval-scaffold}.
\item[U-28] The $+6$ to $+16$ judge-mean lead mixed two aggregation levels; one lower bound is 5.4. \S\ref{sec:nodecay-ablation}, App.~\ref{app:nodecay}.
\item[U-29] Three cross-references went nowhere, one of them pointing at its own section. \S\ref{sec:related-benchmarks}, \S\ref{sec:full-ablation}, Table~\ref{tab:longmemeval-pilot}.
\item[U-30] H1, H2 and H3 named different hypotheses in App.~F than in the preregistration. App.~\ref{app:longmemeval-judge}, App.~\ref{app:longmemeval-scope}, App.~\ref{app:g5-longmemeval-scaffold}.
\end{description}

\paragraph{B.\ Further changes.}
\begin{description}\setlength{\itemsep}{1pt}\setlength{\parskip}{0pt}
\item[Deep copy] The deep copy of TripleCopyMemory saturates with age (Eq.~\ref{eq:deep}); Table~\ref{tab:retention-pma} is recomputed under it (7, 14 and 30~days: $0.576$, $0.789$, $0.899$), and the retention figure of $91.2\%$ is withdrawn at every place. Abstract, \S\ref{sec:pma}, \S\ref{sec:ancillary}, App.~\ref{sec:triple}, App.~\ref{app:retention}, App.~\ref{app:pma-suite-results}, Table~\ref{tab:retention-pma}, App.~\ref{app:locomo-bm25}.
\item[Copies] In none of the intervals of Table~\ref{tab:retention-pma} (1~hour to 30~days) does the fast copy lead; by Eqs.~\ref{eq:fast}--\ref{eq:medium} it leads up to about 54~minutes. The medium copy carries the composite from 1~hour to 3~days, the deep copy from 7~days. App.~\ref{app:pma-suite-results}.
\item[Sleep] The sleep-loop figure left the abstract; App.~\ref{app:sleep-impact} and Table~\ref{tab:sleep} name the generator, the label ``simulation on synthetic facts'' and the run on real data. \S\ref{sec:ancillary}.
\item[Stand-ins] The ablation switches stand-ins acting on simulated values, and four of the fifteen are constant factors on every similarity; the stand-in of vmPFC-FSRS is a constant factor on decay ($\times0.7$) and measures neither the prediction error nor Eq.~\ref{eq:vmpfc-interval}; in the reported runs, where every query follows the whole aging and raises the access counter of every fact, the row of Two-Factor Hebbian measures its interaction with the neuromodulation, the condition of the spectral KG health stand-in never holds during the aging, and the gain from removing PriorityMap comes from the arousal term of its stand-in. Abstract, \S\ref{sec:experiments}, Fig.~\ref{fig:criticality}, \S\ref{sec:discussion}, App.~\ref{sec:remaining-algos}, App.~\ref{app:extended-ablation}, App.~\ref{app:checklist}, \S\ref{sec:intro}, \S\ref{sec:full-ablation}, Table~\ref{tab:gradient}, Table~\ref{tab:challenging-ablation}, Table~\ref{tab:stress-ablation}, App.~\ref{app:reproducibility}, \S\ref{sec:nodecay-ablation}, App.~\ref{sec:integration-cascade}, App.~\ref{app:nodecay}, Table~\ref{tab:nodecay}.
\item[Stress load] Version~8 described the stress level as raising decay and horizon alone; it also enlarges the corpus from 300 to 500 facts and the queries from 100 to 150. Abstract, Fig.~\ref{fig:criticality}, \S\ref{sec:intro}.
\item[Parameters] Table~\ref{tab:hyperparams} gives the parameters version~8 printed for the described architecture, not those of the reported runs; the seven rows that differ from the implementation are marked, and the equation of App.~\ref{sec:simsel} and Algorithm~\ref{alg:sleep} name the TAG weights as those of version~8. App.~\ref{app:hyperparams}, App.~\ref{app:checklist}; notice \S4.
\item[PMA code] The six PMA components run in a private production system; the public repository simulates them in its reproduction package. App.~\ref{app:reproducibility}, Abstract, \S\ref{sec:intro}, \S\ref{sec:related-neuro}, Fig.~\ref{fig:architecture}.
\item[Library] App.~\ref{app:reproducibility} states, measured at version 0.5.1, what the open library does: no stored entry decays, nothing is replayed, and the one strength written back is the FSRS update of \texttt{recordReview}; the lock-score gate of the Two-Factor update is an optional interface, off without a StabilityProtector, and the reproduction package passes none (App.~\ref{sec:twofactor}). Abstract, Fig.~\ref{fig:architecture}, \S\ref{sec:architecture}, App.~\ref{app:algorithms}.
\item[Fig.~3] Redrawn from measured points; the repaired Letta run uses $1{,}049.8$ tokens per question, not the $921.1$ of the published run that version~1 of the notice paired with the repaired accuracy; the non-dominated arms are Mem0, ZenBrain, Letta and the full-context baseline. Fig.~\ref{fig:longmemeval-pareto}, \S\ref{sec:longmemeval-pilot}, App.~\ref{app:longmemeval-long-context}.
\item[Tokens] Version~8 put ZenBrain at $1/106$ of the full-context baseline's tokens (about $1{,}000$ against about $105{,}600$); the history block of the baseline has $105{,}577.9$ tokens per question on average, and counted alike on the whole answer prompts the baseline uses $105{,}669.0$ and ZenBrain $963.2$, $1/109.7$. Abstract, \S\ref{sec:longmemeval-pilot}, \S\ref{sec:discussion}, App.~\ref{app:longmemeval-long-context}, Fig.~\ref{fig:longmemeval-pareto}.
\item[Intervals] Tables~\ref{tab:competitive-combined-v2} and~\ref{tab:longmemeval-e2e} apply one interval procedure to all four arms and name it. \S\ref{sec:experiments}, \S\ref{sec:real-locomo}, App.~\ref{app:longmemeval-e2e}, App.~\ref{app:jm-stats}, App.~\ref{app:checklist}, Fig.~\ref{fig:longmemeval-pareto}.
\item[Test count] The ``11{,}589 CI tests'' of the abstract and \S\ref{sec:intro} were the test count of the production system, not of the library; version~9 does not print the number.
\item[PE signal] \S\ref{sec:mechanisms} and App.~\ref{sec:vmpfc} called the vmPFC prediction-error signal neuromodulation-derived; it is a knowledge-graph-derived signal of our own construction.
\item[Arm] The arm on LoCoMo is the test-harness class, on LongMemEval\_S its Python port; the equality with plain cosine search on the same embedder is reported in \S4 of the notice. \S\ref{sec:limitations}, Table~\ref{tab:competitive-combined-v2}, Abstract.
\item[Package] App.~\ref{app:reproducibility-v6} names the reproduction package at version 0.5.1, its scope (Tables~\ref{tab:full-ablation}--\ref{tab:stress-ablation}) and the run time its record states; Table~\ref{tab:compute-disclosure} carries the logged hours. Abstract, App.~\ref{app:reproducibility}, App.~\ref{app:checklist}.
\item[References] Besides \#20--\#22: Chelazzi et al.\ (2014) for the priority map; LongMemEval at ICLR~2025; six web sources carry the page title of their last archived capture before the access date. Of the six entries the notice could not check, the four web pages have such captures (the other two are an arXiv preprint and our own filing); two of them carried descriptive titles and are corrected. References.
\item[Calibration] App.~\ref{app:bayesian-calibration} calibrates the judge score against gold retrieval and is titled accordingly; no calibration of $P(f)$ is measured, and in the open library the propagation is a heuristic update with a heuristic interval width. \S\ref{sec:mechanisms}, App.~\ref{sec:bayes-formula}, App.~\ref{app:bayesian-calibration}, \S\ref{sec:nodecay-ablation}, App.~\ref{app:nodecay}.
\item[Judges] Version~8 called the comparisons of the three LLM judges comparisons of answer quality and made them the headline; the judges grade the retrieved context, and answers are graded only by the binary judge of the end-to-end runs. Abstract, \S\ref{sec:intro}, \S\ref{sec:longmemeval-pilot}, \S\ref{sec:discussion}, App.~\ref{app:longmemeval-retrieval}, App.~\ref{app:longmemeval-judge}, App.~\ref{app:longmemeval-scope}, App.~\ref{app:longmemeval-e2e}, App.~\ref{app:human-spotcheck}, App.~\ref{app:power-analysis}, App.~\ref{app:checklist}, \S\ref{sec:limitations}, \S\ref{sec:real-locomo}.
\item[StabilityProtector] The implementation (the private production system of item \emph{PMA code}; the test of the reproduction package carries the same cap and threshold rule but neither clamp) caps the access count at $10$ and clamps the lock score and the prediction error to $[0,1]$; computed from the printed equations under these clamps, no update passes once $L\rho > 5/3$, which for a fact with $a \geq 10$ and $c = 1$ holds from day~$239$ if it is a core fact and from day~$1824$ otherwise. The $28.8\%$ of the PMA suite is the result of a test whose criterion is $<$\,30\% and whose sample stays below 400~days. \S\ref{sec:ancillary}, App.~\ref{sec:recon}, App.~\ref{sec:stability}, App.~\ref{app:pma-suite-results}.
\item[Equation 5] Eq.~\ref{eq:vmpfc-interval} has the form of the code, which lengthens the interval at low and shortens it at high prediction error, as the text says; version~8 printed the opposite sign and twice the amplitude. The prediction error is clamped to $[0,1]$. App.~\ref{sec:vmpfc}.
\item[Attribution] Iatropoulos et al.\ (2025) model a synapse as a product of subsynaptic factors; the weight--variance form of the Two-Factor model and its EWC reading are our own construction, and version~9 cites the work as inspiration; Zenke et al.\ (2017), which version~8 cited with it for two-factor rules, is cited for importance-weighted consolidation. \S\ref{sec:intro}, \S\ref{sec:related-neuro}, \S\ref{sec:mechanisms}, App.~\ref{app:extended-related}, App.~\ref{sec:twofactor}, Author Statement.
\item[Hybrid baseline] The reason given for not running a BM25$+$dense hybrid baseline assumed a lexical channel in the measured arm, which is dense-only; the hybrid baseline remains an open check. \S\ref{sec:architecture}, App.~\ref{app:locomo-bm25}.
\item[Clusters] The per-query tests of the four-system comparisons on LoCoMo treat the queries as independent although they are grouped by conversation; no inference with the conversations as clusters is reported. \S\ref{sec:experiments}, App.~\ref{app:jm-pairwise}.
\item[Full-context baseline] The arm whose answer model reads the question's entire history is named the full-context baseline; version~8, the correction notice and the record of version~8's judged outputs (\anon{10.5281/zenodo.22161978}{[DOI redacted for anonymous review]}) call it the oracle, a name that LongMemEval, the benchmark used here, gives to the condition with only the evidence sessions. Abstract, \S\ref{sec:experiments}, \S\ref{sec:longmemeval-pilot}, \S\ref{sec:discussion}, App.~\ref{app:longmemeval-long-context}, Fig.~\ref{fig:longmemeval-pareto}, App.~\ref{app:reproducibility}.
\item[EWC] The Two-Factor penalty has the quadratic form of EWC with the heuristic importance $1/\sigma^2_{ij}$ in place of the Fisher information; version~8 called the two equivalent. The novelty term $N_e$ of the sleep loop counts linked entities. \S\ref{sec:mechanisms}, App.~\ref{sec:twofactor}, App.~\ref{sec:simsel}, App.~\ref{app:algorithms}.
\item[Lock score] Version~8 stated that the learning rate of the Two-Factor update is gated by the StabilityProtector lock score $L$ of App.~\ref{sec:stability}; in the implementation the gate is optional, uses a lock score that the StabilityProtector computes from the edge's variance, $L_\sigma = 1 - \min(\sigma^2_{ij}, 1)$, and acts only when a StabilityProtector is passed; the reproduction package, whose only call site of the update passes none, computes the ungated form. App.~\ref{sec:twofactor}.
\item[Sleep selection] The selection of the sleep loop is binary, LTP above the threshold and LTD otherwise; no candidate is skipped. App.~\ref{sec:simsel}.
\item[Cascade] Table~\ref{tab:cascade} was produced by the cascade suite of the reproduction package, at version~0.5.1; its results file sits in the private repository (App.~\ref{app:reproducibility}), not in the package, and no script of the package writes it. The suite never retrieves, so PriorityMap, MetacognitiveMonitor, compositional context, iMAD and HyperAgent never act in it and the condition of spectral KG health never holds; in the arm without the PMA components three of the nine mechanisms act; the caption and the text of version~8 credited the six PMA components with the backbone, and the text called the full system's advantage an emergent synergy of both groups, although the suite runs no arm with the PMA components alone. The table printed 300 facts, the corpus of the moderate ablation, for the suite's 200, the interval of the full system from an earlier run of the suite next to two from the later one, and $+0.150$ for $0.149$; the suite draws its bootstrap intervals without a seed, its means are deterministic. That the Fiedler value rises by at least 30\% after the suite's sleep step is fixed by construction (every existing edge times 1.3; the edges the step adds can only raise it); version~8 read the rise as a measure of consolidation. \S\ref{sec:experiments}, \S\ref{sec:full-ablation}, Table~\ref{tab:cascade}, App.~\ref{sec:integration-cascade}, App.~\ref{app:reproducibility-v6}, App.~\ref{app:checklist}.
\end{description}

\paragraph{B$'$.\ Statements a reading of every sentence against its source found false or overstated.}
\begin{description}\setlength{\itemsep}{1pt}\setlength{\parskip}{0pt}
\item[K-01] Zep was described as a single-layer knowledge graph without episodic memory and counted with one memory type; it has three tiers of subgraphs (episode, semantic entity, community), by its authors' account the episode tier mirrors episodic memory, and Table~\ref{tab:comparison} counts three, as for MemGPT. \S\ref{sec:related}, Table~\ref{tab:comparison}.
\item[K-02] LoCoMo was said to have all questions share one conversational haystack; it consists of ten conversations, only our four-system comparisons and the ablation cross-validation query one pool of $5{,}882$ facts built from all ten, and in the routing comparisons each question retrieves from its own conversation. \S\ref{sec:related-benchmarks}.
\item[K-03] The three copies of TripleCopyMemory were attributed to the complementary-learning-systems work cited as Schapiro et al.\ (2017) and Kumaran et al.\ (2016); they come from the Basel study, Kveim et al.\ (2024), now cited, and the two remain as motivation. \S\ref{sec:pma}, App.~\ref{sec:triple}, References.
\item[K-04] ``None incorporate cognitive-neuroscience findings'' was said of existing agent memory systems; Table~\ref{tab:comparison} itself credits LightMem with two such features, and the statement holds for the systems cited in the sentence. \S\ref{sec:intro}.
\item[K-05] A superlative and a citation count for Reflexion were printed without index and date; the sentence keeps its point without them. \S\ref{sec:related}.
\item[K-06] Each of the seven layers was said to correspond to an established cognitive construct; five do, and core and cross-context memory are engineering layers. \S\ref{sec:intro}, \S\ref{sec:architecture}.
\item[K-07] All five mechanisms of \S\ref{sec:mechanisms} were said to extend a neuroscience basis without a documented counterpart; four of them do, the set is that of \S\ref{sec:related-neuro}, and query-aware cross-layer retrieval cites no such basis. \S\ref{sec:mechanisms}.
\item[K-08] The Two-Factor model was said to make mature edges robust to catastrophic overwriting; what holds is that mature edges decay more slowly and that the penalty has the form of EWC, which the weight update does not apply and only a test computes; no run measures overwriting. \S\ref{sec:mechanisms}, App.~\ref{sec:twofactor}.
\item[K-09] LightMem and SleepGate were said to use heuristic replay selection; LightMem updates its long-term memory offline from the most similar later entries, without RL scoring or counterfactual candidates, and SleepGate trains a forgetting gate that evicts or compresses stale entries of the KV cache. \S\ref{sec:mechanisms}, App.~\ref{sec:simsel}.
\item[K-10] The retrieval experiments were said to embed with \texttt{text-embedding-3-small} at 1536 dimensions; the generator of the routing and layer runs requests 768 and falls back to a fake embedding provider without a key, and NoDecay and the sleep run on real data embed with \texttt{nomic-embed-text}. \S\ref{sec:experiments}, App.~\ref{app:locomo-bm25}.
\item[K-11] NeuromodulatorEngine was said to drift $6.2\%$ from baseline, confirming homeostatic stability; $0.469$ is the mean largest deviation of a channel from the baseline $0.5$, whose maximum is $0.5$, so the run does not show homeostatic stability. \S\ref{sec:ancillary}, App.~\ref{app:pma-suite-results}.
\item[K-12] NeuromodulatorEngine was counted as always critical; by the criterion of Table~\ref{tab:gradient} it is progressive (unstarred under moderate load), which makes six progressive components and one always critical, and the legends follow. Table~\ref{tab:gradient}, Fig.~\ref{fig:criticality}, \S\ref{sec:full-ablation}.
\item[K-13] The synthetic sleep gain and the loss on real data were presented as one procedure; the gain ($+37\%$ stability, $47.4\%$ lower storage estimate) is the three-phase sleep cycle of the harness (SWS, REM, SHY), the loss that of the Simulation-Selection stand-in. \S\ref{sec:ancillary}, Abstract, App.~\ref{app:sleep-impact}, Table~\ref{tab:sleep}, App.~\ref{app:locomo-bm25}.
\item[K-14] Ebbinghaus was said to have demonstrated exponential decay; he fitted a logarithmic function, and $R = e^{-t/S}$ is the form of the scheduling model. App.~\ref{app:extended-related}, App.~\ref{sec:vmpfc}.
\item[K-15] The efficiency badges were said to use a 30-day sliding window; they compare the last 7~days with the entries older than 90~days (threshold $10\%$), and the 30-day window belongs to the bias tracking. App.~\ref{sec:metacog}.
\item[K-16] Stickgold and Walker (2013) were said to have shown the pruning of weak connections; their perspective piece argues for a triage (stabilize, integrate, generalize, forget). App.~\ref{app:extended-related}.
\item[K-17] The modulation parameters were said to be consumed by the ReconsolidationEngine, the PriorityMap and the sleep loop; none of the four reaches a consumer (no production code requests them, and the memory coordinator's context pruning, which accepts the attention ratio as an optional input, is called without it), the two engines take channel levels as optional inputs, and the sleep cycle pulls the stored levels back toward the baseline. \S\ref{sec:pma}, App.~\ref{sec:neuromod}.
\item[K-18] The time constants of TripleCopyMemory were called empirical hyperparameters calibrated to retention experiments; they are set by hand, and no calibration run exists. App.~\ref{sec:triple}.
\item[K-19] Type-specific resistance thresholds were said to prevent casual overwrites; they are defined but not read in the update path. App.~\ref{sec:recon}.
\item[K-20] Art.~50 of the AI Act was cited for duties of memory-bearing systems; it sets four duties (disclosing the interaction, marking synthetic content machine-readably, informing persons exposed to emotion recognition or biometric categorisation, disclosing deep fakes and generated text published on matters of public interest), and deletion falls under Art.~17 GDPR. App.~\ref{app:broader-impact}.
\item[K-21] Two appendices said that all their numbers are reproduced verbatim from one run; App.~\ref{app:extended-longmemeval} also carries the runs with repaired adapters and recomputations, App.~\ref{app:extended-locomo-judge} recomputations from the files of its run. App.~\ref{app:extended-locomo-judge}, App.~\ref{app:extended-longmemeval}.
\item[K-22] The signed-rank tests over means of three repetitions were computed on floating-point means, whose rounding residues add non-zero pairs and split ties; on exact means the nine $p$-values of App.~\ref{app:longmemeval-judge} are larger (ZenBrain--Letta under Sonnet~4.5 $2.56\times10^{-4}$, not $1.46\times10^{-6}$) and all still pass the corrected level, Letta's six change as well (against A-Mem under Sonnet~4.5 $7.11\times10^{-4}$, not $1.06\times10^{-3}$, so they meet the criterion of H3), and on LoCoMo printed values change, Letta--Mem0 under Sonnet~4.5 no longer passing the corrected level. \S\ref{sec:real-locomo}, App.~\ref{app:longmemeval-judge}, App.~\ref{app:longmemeval-scope}, App.~\ref{app:jm-pairwise}, App.~\ref{app:g5-longmemeval-scaffold}, Abstract, \S\ref{sec:longmemeval-pilot}, Table~\ref{tab:longmemeval-pilot}, \S\ref{sec:discussion}.
\item[K-23] The answer check was called the official protocol of LongMemEval; the templates follow \texttt{get\_anscheck\_prompt()} verbatim only for the standard case, are rewritten for temporal reasoning and preference, and the abstention variant was never drawn ($30$ abstention questions use the template of their category). \S\ref{sec:judge-robustness}, \S\ref{sec:longmemeval-pilot}, App.~\ref{app:longmemeval-e2e}, Table~\ref{tab:longmemeval-e2e}, App.~\ref{app:longmemeval-long-context}, Fig.~\ref{fig:longmemeval-pareto}, \S\ref{sec:discussion}.
\item[K-24] MemPalace, Mem0 and Mastra were said to perform full-context memory consolidation; MemPalace's $96.6\%$ comes from raw semantic search without an LLM, Mem0 extracts with an LLM and retrieves on its own, and only Mastra consolidates in the model's context. \S\ref{sec:intro}, \S\ref{sec:limitations}, App.~\ref{app:longmemeval-e2e}, Table~\ref{tab:longmemeval-e2e}, App.~\ref{app:checklist}.
\item[K-25] The MetacognitiveMonitor was said to detect and surface an agent's confirmation and recency biases, following \citet{fleming2012}; it can compute both from how a user accepts or rejects suggestions, as scores of our own construction that the cited work does not define, and raise an alert only for the asymmetry, no production code calls either, no run concerns agents, and the precision and recall printed for its bias detection come from a function of the test that scores the acceptance asymmetry. \S\ref{sec:pma}, \S\ref{sec:ancillary}, App.~\ref{sec:metacog}, App.~\ref{app:broader-impact}, App.~\ref{app:pma-suite-results}.
\item[K-26] The StabilityProtector was said to prevent casual overwrites and to lower the risk of injection attacks; its rule blocks small updates of existing memories, but in the production system no code writes the lock records it reads, so it blocks none, and no run concerns attacks. \S\ref{sec:pma}, App.~\ref{sec:stability}, App.~\ref{app:broader-impact}.
\item[K-27] Mem0's drop from its published $93.4\%$ under our protocol, on the same system judged by the same template, was called the strongest evidence that the setting rather than the system drives the absolute level; our Mem0 runs with its LLM extraction switched off, at $k = 5$, with our own templates, and how much of the gap the settings explain is not measured. App.~\ref{app:longmemeval-e2e}, Table~\ref{tab:longmemeval-e2e}.
\item[K-28] A-Mem's $k$-ablation was said to plateau at $k$ in $[20, 30]$, making $k = 5$ a standard lower bound; its paper retrieves more than five notes per query in its main setting and names no such range, and what $k = 5$ does to the comparison is not measured. App.~\ref{app:longmemeval-e2e}.
\item[K-29] The full-context baseline was said to use the same answer template; it has its own (the whole history), and answer model and judge are the same. App.~\ref{app:longmemeval-long-context}.
\item[K-30] The caption of Fig.~\ref{fig:retention} credited Two-Factor, vmPFC-FSRS and Simulation-Selection, sleep-driven stabilization and fewer confabulations, and called the reviews of the FSRS-only arm optimally timed, and the text a selective retention of important memories that matches the human forgetting curve; the run computes Ebbinghaus decay with an emotion factor, a review rule with FSRS-style stability, the library's basic Hebbian module and its sleep module, mainly the emotional factor decides what is retained, the FSRS-only arm reviews each fact when it falls due, and no run compares confabulation, human forgetting or the timing of reviews. Fig.~\ref{fig:retention}, App.~\ref{app:retention}.
\item[K-31] Table~\ref{tab:sleep} and its text credited the synthetic sleep gain to the Simulation-Selection loop, with LTP and LTD counts, RL-driven pruning and counterfactual candidates; the run is the three-phase sleep cycle of the harness (per run SWS replays $14.0$ memories, REM $126.0$, $245$ ripples, $24.2$ new edges from REM interleaving). App.~\ref{app:sleep-impact}, Table~\ref{tab:sleep}, Abstract.
\item[K-32] Part A of Table~\ref{tab:hebbian-bayesian} was credited to Two-Factor dynamics with importance-weighted edges and emergent clusters; the run uses the library's basic Hebbian module, and the cluster structure is planted by the co-activations; Part B's separation was read as confirming the contradiction relations, which the generator plants between false and true facts. \S\ref{sec:related-neuro}, \S\ref{sec:experiments}, \S\ref{sec:ancillary}, App.~\ref{app:extended-benchmarks}, App.~\ref{app:twofactor-bayesian}, Table~\ref{tab:hebbian-bayesian}, App.~\ref{app:checklist}.
\item[K-33] $-0.130$ was printed as a correlation with $p < 0.01$; it is the mean product of the dopamine and serotonin deviations that the built-in coupling of $-0.3$ produces, and no test was run. App.~\ref{app:pma-suite-results}.
\item[K-34] Four values carried a wrong last digit; they read $0.214$, $0.767$, $0.510$ and $-16.8\%$. Table~\ref{tab:stress-ablation}, Table~\ref{tab:nodecay}, Table~\ref{tab:retention-pma}.
\item[K-35] NoDecay was said to disable only the decay step; each simulated day ends once the age counter advances, so the daily mechanisms after it do not run either, and at the strength cap, where every fact starts, they could not raise a strength. \S\ref{sec:real-locomo}, App.~\ref{app:nodecay}, Table~\ref{tab:nodecay}.
\item[K-36] Simple Memory was said to lose about $52\%$ of its facts; its generator loses $56\%$. App.~\ref{app:long-horizon}.
\item[K-37] The NoDecay $p = 0.043$ was computed over ten copies of one run (the seeds feed a random draw that is never taken), and the results file carries no per-query values; no $p$-value is reported, and $\Delta P@5 = 0.002$ and $|d| = 0.015$ remain. Abstract, \S\ref{sec:real-locomo}, App.~\ref{app:nodecay}, Table~\ref{tab:nodecay}.
\item[K-38] The routing lead of version~8 ($+20.7\%$ F1 over Flat Store, carried by the temporal questions) and its layer drops came from a run with placeholder embeddings: run without a key, the generator gives every mean of version~8's layer ablation and the F1 of the ZenBrain row of Table~\ref{tab:locomo}; in a later run with \texttt{text-embedding-3-small} the flat variant (F1 $0.059$) lies above the full system ($0.058$) and both above BM25-only, which uses no embedder ($0.052$ in the placeholder run, on the same subsamples), and Table~\ref{tab:ablation} shows both runs, with the Core-Memory spread of the placeholder run ($0.001$; version~8 printed $0.000$). Abstract, \S\ref{sec:intro}, \S\ref{sec:experiments}, \S\ref{sec:real-locomo}, \S\ref{sec:ancillary}, App.~\ref{app:locomo-bm25}, Table~\ref{tab:locomo}, App.~\ref{app:layer-ablation}, Table~\ref{tab:ablation}, App.~\ref{app:ablation}, Table~\ref{tab:locomo-categories}.
\item[K-39] Short-term and cross-context memory were read as layers without measurable impact on LoCoMo, and the full variant as ``all 7 layers active''; they are not layers of the evaluated class, and their rows are $0.0\%$ by construction. App.~\ref{app:layer-ablation}, App.~\ref{app:ablation}.
\item[K-40] Reconsolidation's mode accuracy of $\geq 95\%$ was read as correct classification; it matches 3 of 8 hand-labelled pairs, the 92 random pairs per seed test its threshold rule against itself, and contradiction detection uses five fixed pairs. \S\ref{sec:ancillary}, App.~\ref{app:pma-suite-results}.
\item[K-41] The gap of the long-horizon archetypes was put down to algorithmic protection; the archetypes also differ in initial strength, in how strength enters the score and in the threshold $0.1$ that only Simple Memory has. \S\ref{sec:full-ablation}, App.~\ref{app:long-horizon}.
\item[K-42] PriorityMap's NDCG@10 of $0.997$ was measured against a known ground-truth importance; the test's reference is its own weighting of the same four inputs ($0.15/0.3/0.25/0.3$ against the priority weights $0.2/0.25/0.25/0.3$). \S\ref{sec:ancillary}, App.~\ref{sec:priority}, App.~\ref{app:pma-suite-results}.
\item[K-43] F1 and task success were said to come from LoCoMo and from synthetic tasks; both are taken on the same LoCoMo questions (a stratified $80\%$ subsample per seed), and the definition of task success is given. App.~\ref{app:ablation}, App.~\ref{app:locomo-bm25}, Table~\ref{tab:locomo}.
\item[K-44] The CO$_2$ estimate rested on about $25$M judge tokens; the runs logged no token counts, and a count from the result files of the four-system runs gives $66{,}888$ scored LoCoMo judge calls and $29$ to $41$M tokens (at $5$ to $3.5$ characters per token), so $29$ to $41$\,kg for these calls alone. App.~\ref{app:compute-disclosure}.
\item[K-45] The embedding cache was given as about $293$\,MB, $5{,}882$ calls and \$1; the last listing shows $2.7$\,GB, $7{,}868$ texts (about $0.21$M tokens) are embedded, about \$$0.004$ at \$$0.02$ per million tokens. App.~\ref{app:compute-disclosure}.
\item[K-46] Algorithms~1 and~2 describe the architecture as version~8 gave it; their captions say which steps the code we examined performs, which it does not, and what the library's coordinator does at version~0.5.1. App.~\ref{app:algorithms}.
\item[K-47] Algorithm~\ref{alg:sleep} printed a sorted selection with a fixed LTP increment, pruning and the test $\geq$; it now follows the code, and App.~\ref{app:broader-impact} no longer counts LTD-based edge pruning among the forgetting mechanisms. \S\ref{sec:mechanisms}, App.~\ref{sec:simsel}, App.~\ref{app:broader-impact}, Algorithm~\ref{alg:sleep}.
\item[K-48] Table~\ref{tab:compute-disclosure} was called the total compute of all reported experiments; it names what it leaves out, among them the $7{,}779$ scored calls of the reference judge on LoCoMo. acknowledgments, App.~\ref{app:compute-disclosure}, Table~\ref{tab:compute-disclosure}, App.~\ref{app:checklist}.
\item[K-49] \texttt{claude-sonnet-4-6} was called a rolling alias; it is a pinned snapshot like \texttt{claude-opus-4-6} (Anthropic, Models overview, retrieved 18~October~2026). Table~\ref{tab:competitive-combined-v2}, App.~\ref{app:jm-coverage}, App.~\ref{app:checklist}.
\item[K-50] The judge scripts were said to issue the retrieval calls; the lists of Mem0, Letta and A-Mem were written by \texttt{run\_baselines.py} and ZenBrain's by a separate run of its arm, and the judge scripts only call the judges. App.~\ref{app:jm-repro}.
\item[K-51] That single-hop is the weakest category for Opus and GPT-4o was explained by a near-binary correctness signal; the single-hop scores of GPT-4o spread across the scale. App.~\ref{app:jm-human}.
\item[K-52] ZenBrain's higher Brier and ECE were put down to depth and score granularity; recomputed at depth~5 from the stored judge files, depth explains the gap to Letta, not that to A-Mem, and the granularity holds for all systems alike. App.~\ref{app:bayesian-calibration}.
\item[K-53] The failure analysis was said to take the 20 lowest-scoring queries; it took 20 of the 682 queries scored 0, the first by identifier among those without a gold fact in the top five, all from one conversation. App.~\ref{app:failure-modes}, Table~\ref{tab:failure-modes}.
\item[K-54] ``13 of 20'' queries in the four named clusters are 17 of 20, the clusters follow the rules of the classifier, not a check of the retrieved entity, and the entity-aware reranking said to target the largest cluster directly was not run on these queries. App.~\ref{app:failure-modes}.
\item[K-55] The ablations of the five most critical components were read as a judge-confirmed degradation; the reference $0.380$ holds for all questions with another embedder and depth, on the same 200 questions it is $0.424$, the ablation of \texttt{two\_factor\_hebbian} leaves 199 of 200 retrieved lists unchanged and scores $0.308$, and the other four lie within $0.012$ of it. App.~\ref{app:ablation-judge}, Table~\ref{tab:ablation-judge}.
\item[K-56] The checklist cited repositories that the references did not give; they are now cited at their entries. References.
\item[K-57] The checklist named Cohen's $d$ for all effect sizes; the simulation, sleep and NoDecay runs and repetition 42 alone report Cohen's $d$ with pooled SD. \S\ref{sec:experiments}, \S\ref{sec:real-locomo}, App.~\ref{app:jm-stats}, App.~\ref{app:pairwise-significance}, App.~\ref{app:checklist}, App.~\ref{app:g5-longmemeval-scaffold}, App.~\ref{app:jm-repro}.
\item[K-58] App.~\ref{app:broader-impact} counted vmPFC-FSRS decay among the forgetting mechanisms that give users control and named three building blocks for Article~50; the vmPFC coupling has no caller in the production system and only rescales review intervals, only calendar entries carry an \texttt{ai\_generated} flag, assistant responses none, none of the trace events App.~\ref{app:broader-impact} named exists, and the deletion endpoints erase memory per context, layer and learned fact but do not reach several stores, among them the table into which the assistant stores facts, while the assistant's deletion tools only deactivate a fact. App.~\ref{app:broader-impact}, App.~\ref{app:checklist}.
\item[K-59] App.~\ref{app:broader-impact} named four mitigations---human approval for sensitive memory operations, reconsolidation logged for forensic rollback, the ablation registry as a kill-switch in production and the absence of any cross-user data sharing---and \S\ref{sec:pma} and App.~\ref{sec:recon} the rollback, with the logged snapshot as a safety mechanism found in none of the surveyed systems; in the production system the governance layer asks for approval of proactive and social actions only under a stored policy or at high or critical risk, sending e-mail and memory operations do not pass through it, the engine's insert names columns that the migration of its event table does not define, so against that schema no reconsolidation event is stored, the stored fact is written outside the engine and the rollback restores nothing, no production code reads the registry, which switches components off in the evaluation harness, the table into which the assistant stores facts is read by context, not by user, and the personal profile has no user key; Mem0's open code keeps the previous value of every update and deletion. \S\ref{sec:pma}, \S\ref{sec:discussion}, App.~\ref{sec:recon}, App.~\ref{app:broader-impact}, App.~\ref{app:checklist}.
\item[K-60] The checklist credited App.~\ref{app:broader-impact} with transparent emotional tagging, an opt-in Ebbinghaus decay and governance policies for emotional memory, with deletion tools as a positive impact, with the risks of manipulation via emotional weighting and of privacy leakage across contexts, and with per-context schema isolation to prevent cross-context leakage; App.~\ref{app:broader-impact} names none of the first three, its deletion tools only deactivate a fact, its risks are behavioral profiling, bias reinforcement, amplified affective responses and overwriting through reconsolidation, and the nightly consolidation and the personal profile bridge the isolation. App.~\ref{app:checklist}, App.~\ref{app:broader-impact}.
\item[K-61] \S\ref{sec:architecture} called the merging of the cross-context layer privacy-aware with configurable access controls; no access control gates it. \S\ref{sec:architecture}.
\item[K-62] App.~\ref{app:compute-disclosure} named Sonnet~4.5, Opus~4.6 and GPT-4o for the two keys; the runs also call Sonnet~4.6, \texttt{gpt-4o-mini} and \texttt{text-embedding-3-small}. App.~\ref{app:compute-disclosure}, acknowledgments, App.~\ref{app:checklist}, Table~\ref{tab:compute-disclosure}.
\item[K-63] App.~\ref{app:neurips-delimitation} said the ReconsolidationEngine modifies already-stored memories, and App.~\ref{sec:recon} printed its mode rule without the session penalty and its raw PE with a contradiction bonus; in the production system the engine modifies no stored memory, no caller triggers the bonus, and the implementation subtracts the penalty of $0.1$ before it selects the mode. App.~\ref{app:neurips-delimitation}, App.~\ref{sec:recon}.
\item[K-64] App.~\ref{app:pma-suite-results}, \S\ref{sec:pma} and \S\ref{sec:ancillary} named the PMA components as the subject of the PMA suite; the suite runs its own re-implementations of them, imports none of their classes and detects contradictions with its own PE function. \S\ref{sec:pma}, \S\ref{sec:ancillary}, App.~\ref{app:pma-suite-results}.
\item[K-65] App.~\ref{sec:remaining-algos} gave the Metacognitive HyperAgent its described role, a meta-policy over MetacognitiveMonitor, and said that it, iMAD and spectral KG health trigger in deployment on rarer events; the module of this name in the production system is a self-improvement agent over its own strategies, does not read the monitor and never reports a strategy, because its daily worker passes each with one attempt and it evaluates only those with at least five; no code outside its own two files calls iMAD; and spectral KG health runs as stage~9 of the sleep cycle but reads a relation table that no migration creates, so against that schema it never reports fragmentation. App.~\ref{sec:remaining-algos}.
\item[K-66] The statement of competing interests did not name the author's affiliation as the organization that develops the open library and operates the private production system, a sole proprietorship of the author that also builds \anon{the products ZenAI and Eigenraum}{two products} on the ZenBrain architecture; neither product is sold at the time of submission, a later commercial use of the architecture, for example to finance the research, is not excluded, and no concrete steps toward it are planned. Competing interests.
\item[K-67] No surveyed system was said to integrate more than two of the compared mechanisms, and none of the three NeurIPS contributions more than one; the comparison table scores eight systems, none of which has more than two of its nine features, and the other systems are described, not scored. Abstract, \S\ref{sec:intro}, \S\ref{sec:related-neuro}, App.~\ref{app:concurrent}, App.~\ref{app:neurips-delimitation}.
\item[K-68] The F1 on LoCoMo was described as the benchmark's public, substring-based evaluation; our harness scores retrieval with a token F1 against the gold answers. Abstract, \S\ref{sec:intro}, \S\ref{sec:real-locomo}, App.~\ref{app:locomo-bm25}.
\item[K-69] Nine of the fifteen mechanisms were called explicit instantiations of prior literature, among them Two-Factor synaptic edges, vmPFC-FSRS and Simulation-Selection sleep; seven of the nine foundational algorithms name an origin in the literature, and those three are variants of the evaluation harness motivated by recent work. \S\ref{sec:intro}, Author Statement, App.~\ref{app:checklist}.
\item[K-70] Webb et al.\ (2025) were cited as independent convergent validation of the neuro-inspired methodology; their result concerns their own architecture, not this one. \S\ref{sec:related-neuro}, acknowledgments, App.~\ref{app:practitioner}, App.~\ref{app:extended-related}.
\item[K-71] Mem0's spread across the three LoCoMo repetitions and its lower intra-judge agreement were read as a seed instability of Mem0, not of our measurement, and its lead in P@5, R@5 and F1 was put down to a permissive recall budget; its store was not reset between the repetitions, a harness defect of ours, and its raw figures count identifiers repeated in that store (deduplicated P@5 $0.071$). \S\ref{sec:judge-robustness}, \S\ref{sec:real-locomo}, Table~\ref{tab:competitive-combined-v2}, \S\ref{sec:seed-robustness}, App.~\ref{app:jm-coverage}.
\item[K-72] The 59 Letta failures on LongMemEval\_S were put down to the Docker server and a pilot-scale transient, and the halving retry was described as shared preprocessing; the Letta adapter lacked the halving fallback of the ZenBrain and A-Mem adapters, a defect of ours. \S\ref{sec:longmemeval-pilot}, Table~\ref{tab:longmemeval-pilot}, App.~\ref{app:longmemeval-retrieval}, App.~\ref{app:longmemeval-scope}.
\item[K-73] Letta's retrieval lead over ZenBrain on the 441 LongMemEval\_S tasks its published adapter served was put at about $2$--$3$ points on every metric; the harness counts an identifier repeated within a list twice, the published Letta lists repeat one on 21 of those tasks, and on distinct facts the lead is between $0.41$ and $1.32$ points. \S\ref{sec:longmemeval-pilot}, Table~\ref{tab:longmemeval-pilot}, App.~\ref{app:longmemeval-retrieval}, App.~\ref{app:longmemeval-scope}.
\item[K-74] The Author Statement and the checklist counted a reasoning backend for agent-level experiments among the methodological uses of LLMs; no result file contains such a role. Author Statement, App.~\ref{app:checklist}.
\item[K-75] Hebb (1949) was quoted with ``neurons that fire together wire together'', and Pimsleur (1967) was said to have introduced spaced repetition; the phrase is a later slogan that summarizes Hebb's proposal, and Pimsleur introduced a graduated-interval recall schedule. App.~\ref{app:extended-related}.
\item[K-76] The MetacognitiveMonitor was said to detect urgency signals, to boost encoding in its novelty windows and to produce efficiency badges that surface in the user interface; in the production system no code calls the urgency detection, the novelty boost or the badge. App.~\ref{sec:metacog}.
\item[K-77] App.~\ref{app:longmemeval-agreement} said that Letta's agreement statistics are quoted only for the tasks on which it returned a retrieval; none is reported for LongMemEval\_S. App.~\ref{app:longmemeval-agreement}.
\item[K-78] The caption of Table~\ref{tab:longmemeval-e2e} called the numbers reported for MemPalace, Mem0 and Mastra the public LongMemEval leaderboard; they are vendor self-reports, not entries on a common leaderboard. Table~\ref{tab:longmemeval-e2e}.
\item[K-79] In App.~\ref{app:longmemeval-long-context} the per-category differences for multi-session and single-session-preference carried a wrong last digit; they read $+6.5$ and $+5.6$. App.~\ref{app:longmemeval-long-context}.
\item[K-80] The growing gap between emotional and neutral retention in the cascade suite was said to confirm the amygdala-mediated consolidation effect of McGaugh (2004); the model was built to encode that effect, and the suite reproduces it. App.~\ref{sec:integration-cascade}.
\item[K-81] App.~\ref{app:retention} called the 30-day value of the PMA suite the strength composite of a single TripleCopy trace; the suite averages over initial strengths drawn at random with mean $0.912$. App.~\ref{app:retention}, Table~\ref{tab:retention-pma}, Abstract, \S\ref{sec:pma}, \S\ref{sec:ancillary}, App.~\ref{sec:triple}, App.~\ref{app:pma-suite-results}.
\item[K-82] Stickgold and Walker (2013) and O'Neill et al.\ (2010) were cited for the observation that sleep preferentially consolidates at-risk memories; they describe selection by salience and the reactivation of waking experience. App.~\ref{app:sleep-impact}.
\item[K-83] The PMA suite was said to show the amygdala fast-path elevating high-emotion items above the priority floor; its test raises a high-emotion item to the floor, $P = 0.5$. App.~\ref{app:pma-suite-results}.
\item[K-84] App.~\ref{app:compute-disclosure} said the LoCoMo dump was used in all reported experiments; the runs on LongMemEval\_S, the simulation experiments, the routing comparisons and the sleep run on real data do not use it. App.~\ref{app:compute-disclosure}.
\item[K-85] The four experiment suites of the reproduction package were said to produce JSON output, which a helper script captures; they print their results to standard output, and a script of the package writes only those of the ablation suite as JSON. App.~\ref{app:reproducibility-v6}.
\item[K-86] App.~\ref{app:jm-repro} said that all numbers of the LoCoMo comparison and of the seed-robustness analysis can be re-derived from one analysis file; the pairwise statistics of the LoCoMo comparison came from \texttt{g5-judge-significance.json}. App.~\ref{app:jm-repro}.
\item[K-87] A cross-reference of the checklist sent the BM25 framing of LoCoMo to the discussion, which does not contain it; it is in the LoCoMo comparison. App.~\ref{app:checklist}.
\item[K-88] App.~\ref{app:additional-capabilities} said the STC rescue module rescues fading memories when a nearby strongly-consolidated memory is activated within the STC window; it is written to rescue a weakly encoded memory when a strongly encoded memory of the same topic forms within that window, and no code of the production system calls it. App.~\ref{app:additional-capabilities}.
\item[K-89] App.~\ref{app:additional-capabilities} said that eight specialist modules of the Global Workspace Theory context assembler compete for a shared workspace and that the winning coalition is broadcast to all layers; ten modules compete beside an always-included core-memory module, the content of the winners is assembled into the context passed to the model, and no code of the production system calls the assembler. App.~\ref{app:additional-capabilities}.
\item[K-90] App.~\ref{app:additional-capabilities} said that intention triggers fire when a designated retrieval cue is detected and that time-based ones are handled by FSRS; the module checks time, activity and context triggers itself, and no code of the production system calls these checks. App.~\ref{app:additional-capabilities}.
\item[K-91] App.~\ref{app:additional-capabilities} said that the curiosity engine prioritizes regions of the KG with high uncertainty and positive recent learning slope; it ranks the topics of a chat query log, its learning-progress bonus favours domains whose error rises because the code reads the newest error first, and no migration creates the query log it reads. App.~\ref{app:additional-capabilities}.
\item[K-92] App.~\ref{app:additional-capabilities} said that none of its four mechanisms is documented in the surveyed memory systems; Vestige documented synaptic tagging and prospective memory in its README as archived on 7~April 2026. App.~\ref{app:additional-capabilities}.
\item[K-93] \S\ref{sec:related-neuro} named quantified knowledge-gap detection among the mechanisms we did not find in the surveyed systems without saying where the paper describes it; the paper describes gap detection only for the curiosity engine of App.~\ref{app:additional-capabilities}, which reads a query log that no migration creates. \S\ref{sec:related-neuro}.
\item[K-94] App.~\ref{app:locomo-bm25} called memory lifecycle management (retention and sleep consolidation) ZenBrain's primary contributions; the retention figure of TripleCopyMemory is a property of its formula (item \emph{Deep copy}), and the sleep gain is that of the harness, with a loss on real data (K-13). App.~\ref{app:locomo-bm25}.
\item[K-95] \S\ref{sec:pma} and App.~\ref{sec:triple} described TripleCopyMemory as storing each event in three copies combined by their maximum; in the production system the store writes only the fast and the medium copy, and its insert names a column that the migration of its table does not define, so against that schema no copy is stored; no code writes the deep copy or calls the retrieval, which takes the first copy with strength at least $0.3$ in the order fast, medium, deep. \S\ref{sec:pma}, App.~\ref{sec:triple}.
\item[K-96] \S\ref{sec:related-neuro} and App.~\ref{app:practitioner} described Anthropic's \emph{Dreams} as memory consolidation on a schedule and cited its documentation page; the page describes an asynchronous job that an API call creates and names no schedule; the schedule is that of the announcement of 6~May 2026. \S\ref{sec:related-neuro}, App.~\ref{app:practitioner}, References.
\end{description}

\paragraph{C.\ The correction notice.}
Version~2 of the notice corrects its own statements where version~1
deviated from the audit or from later measurements; its section
``Changes in version~2'' lists them. Where the notice and the audit
differed, version~9 follows the audit.

\vskip 0pt plus 4\baselineskip\penalty-200\vskip 0pt plus -4\baselineskip
\paragraph{D.\ Versions of this work.}
\begin{center}
\begin{tabular}{lll}
\toprule
Date & Place & Identifier \\
\midrule
2026-03-31 & Zenodo, version 1 & \anon{10.5281/zenodo.19353664}{[redacted]} \\
2026-04-01 & Zenodo, version 2 & \anon{10.5281/zenodo.19369852}{[redacted]} \\
2026-04-01 & TDCommons & \anon{dpubs\_series/9683}{[redacted]} \\
2026-04-03 & Zenodo, version 3 & \anon{10.5281/zenodo.19410934}{[redacted]} \\
2026-04-04 & Zenodo, version 4 & \anon{10.5281/zenodo.19413933}{[redacted]} \\
2026-04-06 & TDCommons & \anon{dpubs\_series/9720}{[redacted]} \\
2026-04-07 & Zenodo, version 5 & \anon{10.5281/zenodo.19450421}{[redacted]} \\
2026-04-07 & TDCommons (``v5'') & \anon{dpubs\_series/9732}{[redacted]} \\
2026-04-09 & Zenodo, version 6 & \anon{10.5281/zenodo.19481262}{[redacted]} \\
2026-04-26 & arXiv, v1 & arXiv:2604.23878v1 \\
2026-05-02 & arXiv, v2 & arXiv:2604.23878v2 \\
2026-05-03 & Zenodo, version 7 & \anon{10.5281/zenodo.19482704}{[redacted]} \\
2026-05-05 & TDCommons (``v7'') & \anon{dpubs\_series/10021}{[redacted]} \\
2026-08-09 & Zenodo, version 8 & \anon{10.5281/zenodo.21858218}{[redacted]} \\
2026-08-09 & arXiv, v3 & arXiv:2604.23878v3 \\
2026-09-18 & Correction notice, version 1 & \anon{10.5281/zenodo.22831087}{[redacted]} \\
\bottomrule
\end{tabular}
\end{center}
The Zenodo versions share the concept
\anon{10.5281/zenodo.19353663}{[redacted]}; versions 5 and 6 carry the MIT
licence, all others CC BY 4.0. Dates are submission dates for arXiv and
publication dates otherwise. The correction notice, cited in this paper by
its concept \anon{10.5281/zenodo.22831086}{[redacted]}, appears here with
its version~1.
}

\section{Paper Checklist}
\label{app:checklist}

\begin{enumerate}

\item \textbf{Claims.}
Do the main claims made in the abstract and introduction accurately reflect the paper's contributions and scope?
\answerYes{The abstract and Introduction state four explicitly bounded
claims (\S\ref{sec:intro}, ``Scope and claims''): (i)~the answer-level
claim is end-to-end accuracy under a binary judge on LongMemEval\_S, while
the three LLM judges grade the retrieved context as a diagnostic;
(ii)~the architecture integrates nine foundational algorithms, seven of
which name an origin in the literature, and six PMA components;
(iii)~the LLM judges are a calibrated proxy, checked by agreement between
two raters, a cross-provider bias check and a single-rater human
spot-check; (iv)~no comparison against MemPalace or Mastra in their tuned
configurations. Experimental validation:
\S\ref{sec:experiments}.}

\item \textbf{Limitations.}
Does the paper discuss the limitations of the work performed by the authors?
\answerYes{See \S\ref{sec:limitations} (dedicated Limitations
paragraph): ablations on synthetic corpora with stand-ins acting on
simulated values, the LLM judges as a diagnostic of the retrieved context,
a single answer model (\texttt{gpt-4o-mini}; transfer to other models
untested), no comparison against MemPalace or Mastra in their tuned
configurations, and the
measured arm as a test-harness class and its port rather than the
described architecture; plus
\S\ref{sec:real-locomo} (LoCoMo BM25 framing) and
App.~\ref{app:broader-impact} (broader impact).}

\item \textbf{Theory assumptions and proofs.}
\answerNA{Empirical systems paper.}

\item \textbf{Experimental result reproducibility.}
Does the paper fully disclose all the information needed to reproduce the main experimental results of the paper to the extent that it affects the main claims and/or conclusions of the paper?
\answerYes{The npm packages
(\anon{\texttt{@zensation/algorithms@0.2.0} and \texttt{@zensation/core@0.2.0}}{[package names redacted for anonymous review]})
and the reproduction package of
Tables~\ref{tab:full-ablation}--\ref{tab:stress-ablation} are public, and
those tables regenerate from it without API keys
(App.~\ref{app:reproducibility-v6}); the answer-level comparison with Letta on
LongMemEval\_S recomputes from the per-question files shipped with the
correction notice (App.~\ref{app:reproducibility}). The arm measured as
ZenBrain in the four-system comparisons is a test-harness class and its
Python port, not the packages, and the harness that ran those comparisons
is private; the LoCoMo pool is a derived dump whose hash cannot be
produced from the public release (App.~\ref{app:compute-disclosure}).
Seeds of the simulation experiments are documented in
App.~\ref{app:reproducibility}; the statistical protocol is in
Section~\ref{sec:experiments} and App.~\ref{app:judge-methodology}.}

\item \textbf{Open access to data and code.}
Does the paper provide open access to the data and code, with sufficient instructions to faithfully reproduce the main experimental results?
\answerNo{The npm packages and the reproduction package of
Tables~\ref{tab:full-ablation}--\ref{tab:stress-ablation} are public under
Apache-2.0 (App.~\ref{app:reproducibility-v6}). The harness of the
four-system comparisons and its result files are kept in a private
repository; published are the judged outputs and analysis scripts of the
LongMemEval\_S comparison as a measurement package, the per-question
outcomes and judge scores of the comparison with Letta, shipped with the correction notice, and the LoCoMo
retrieval metrics as a data record (App.~\ref{app:reproducibility}).}

\item \textbf{Experimental setting/details.}
Does the paper specify all the training and test details necessary to understand the results?
\answerYes{Section~\ref{sec:experiments}, Appendix~\ref{app:hyperparams};
Table~\ref{tab:hyperparams} lists the parameters version~8 gave for the
described architecture, not the 33 constants of the class behind the
evaluated arm (App.~\ref{app:hyperparams}). The
effective settings of the three compared systems are given with their
versions in \S\ref{sec:experiments}, with their embedding backbones in
\S\ref{sec:judge-robustness} and their ingest caps in
\S\ref{sec:longmemeval-pilot}, and with their retrieval depths
in Table~\ref{tab:bayes-calibration}.}

\item \textbf{Experiment statistical significance.}
Does the paper report error bars suitably and correctly defined or other appropriate information about the statistical significance of the experiments?
\answerYes{95\% intervals (a percentile bootstrap over questions, or over the ten seeds in Table~\ref{tab:cascade}, or the range of three repetitions where a caption says so), Wilcoxon signed-rank tests with Bonferroni correction, and effect sizes $d_z$ on paired differences
for the judged system comparisons and Cohen's $d$ with pooled SD for repetition 42 alone, the
simulation runs, the sleep run on real data and the NoDecay counter-experiment.}

\item \textbf{Experiments compute resources.}
Does the paper report the computational resources consumed?
\answerYes{All experiments run on a single Apple M-series laptop
(16\,GB RAM), serving \texttt{nomic-embed-text} (768-dim) via Ollama for
the arms configured for it (\S\ref{sec:judge-robustness}). The retrieval
stage is not free of external calls: A-Mem, and Mem0 on LongMemEval\_S, are
configured with \texttt{gpt-4o-mini} (Mem0 on LoCoMo with its library's
default model), but neither calls an LLM in the reported runs (Mem0 stores
with \texttt{infer=False}, A-Mem's LLM-driven note evolution is switched
off), and the only OpenAI calls inside the two pipelines are those of
Mem0's embedder on LoCoMo; the ZenBrain arm embeds with OpenAI in the ablation
cross-validation and in the later routing-ablation run; their cost, and that
of the \texttt{gpt-4o-mini} answer and judge calls of the end-to-end runs,
is not reported. Simulation-only components (retention, sleep,
Hebbian KG, Bayesian propagation) complete in $<$\,5 minutes, and the
ablation suite of Tables~\ref{tab:full-ablation}--\ref{tab:stress-ablation}
in 13\,s according to the record of its reproduction package
(App.~\ref{app:reproducibility-v6}). The four-system runs took $4.6$\,h on
LoCoMo and $47.7$\,h on LongMemEval\_S (logged); the LLM-as-Judge grading
(Sonnet~4.5, Opus~4.6, GPT-4o via Anthropic/OpenAI APIs) amounts to
$<$\,\$200 in the judge rows of Table~\ref{tab:compute-disclosure}
(\anon{matches Funding-paragraph total; }{}per-judge breakdown there), which
leave out the judge calls of the LongMemEval\_S runs with the Letta and Mem0
adapters repaired, those of the ablation cross-validation
(App.~\ref{app:ablation-judge}) and those of the reference judge
\texttt{claude-sonnet-4-6}.}

\item \textbf{Code of ethics.}
Does the research conform to an ethics review?
\answerYes{The benchmarks evaluate AI system outputs, not human
cognition; no IRB protocol was required. A limited human evaluation
($n{=}50$ disagreement-stratified LoCoMo queries, single independent
rater, blind to LLM-judge scores) was conducted as a robustness check
for the LLM-as-Judge methodology
(\S\ref{sec:judge-robustness}, App.~\ref{app:jm-human});
informed consent was obtained. Privacy implications of memory-bearing
AI systems are discussed in Broader Impact (App.~\ref{app:broader-impact}).}

\item \textbf{Broader impacts.}
Does the paper discuss both potential positive and negative societal impacts?
\answerYes{Section~\ref{sec:discussion} carries a compact broader-impact
paragraph that Appendix~\ref{app:broader-impact} expands, enumerating
positive impacts (deletion endpoints, the open release),
risks (behavioral profiling, bias reinforcement, amplified affective
responses, overwriting through reconsolidation), and what the production
system provides toward mitigation.}

\item \textbf{Safeguards.}
Does the paper describe safeguards for responsible use?
\answerYes{GDPR-aligned forgetting through deletion endpoints per context, layer
and learned fact, which miss several stores, among them the assistant's own fact
table (Appendix~\ref{app:broader-impact}), and, in the private
production system, an opt-in retention period with automatic deletion,
scoped API keys and per-context schema isolation, which the
nightly consolidation bridges for preference, goal, basic-information and
personality facts and the personal profile also crosses.}

\item \textbf{Licenses.}
Are the creators of assets used in the paper properly credited and
the license terms respected?
\answerYes{All baselines (Mem0 Apache-2.0, Letta Apache-2.0, A-Mem
MIT) and the two evaluation benchmarks are cited with their original
papers and repositories; LoCoMo is released under CC BY-NC 4.0 and was
used for non-commercial research, LongMemEval\_S under the MIT license.}

\item \textbf{New assets.}
Are new assets introduced in the paper well documented and available?
\answerYes{ZenBrain is released as Apache-2.0-licensed npm packages
(\anon{\texttt{@zensation/algorithms}, \texttt{@zensation/core}}{package names redacted for anonymous review}) with
full API documentation, usage examples, and adapter templates.}

\item \textbf{Crowdsourcing and human subjects.}
\answerYes{Limited human evaluation only: the $n{=}50$ human-anchor
spot-check (\S\ref{sec:judge-robustness},
App.~\ref{app:jm-human}) used a single independent
rater, blinded to LLM-judge scores, scoring queries on the same 0--5
rubric used by the three LLM judges. No crowdsourcing platform was
used; rating took ${\approx}90$ minutes; rater is the paper author's
contact; compensation: none (informal collaboration); informed
consent: yes. The evaluation targets AI system outputs, not human
cognition or behavior, so no IRB approval was required.}

\item \textbf{IRB approvals.}
\answerNA{Evaluation targets AI system outputs (LLM-as-Judge
robustness check), not human cognition. Consult Item 14 above for
the rater protocol.}

\item \textbf{Use of LLMs.}
Does the paper disclose the use of large language models in the
research itself (beyond incidental writing assistance)?
\answerYes{LLMs are used or configured in six disclosed roles:
(a)~coding assistant for implementation scripts (Author Statement);
(b)~writing aid for drafting and editing prose (Author Statement);
(c)~literature-search and synthesis assistant for related-work
retrieval (Author Statement); (d)~LLM-as-Judge graders of the retrieved
context, Sonnet~4.5, Opus~4.6 and GPT-4o at temperature~0 (on LoCoMo,
Sonnet~4.5 and GPT-4o over three repetitions labelled 42/123/456 and
Opus~4.6 over three for Mem0 and one for the other systems; on
LongMemEval\_S, each judge three times for ZenBrain and once for each
competitor; Sonnet~4.5 also in the ablation cross-validation of
App.~\ref{app:ablation-judge}), with \texttt{claude-sonnet-4-6}, a pinned
snapshot like \texttt{claude-opus-4-6}, as a reference only
(App.~\ref{app:judge-methodology});
(e)~\texttt{gpt-4o-mini} as answer model and binary judge of the
end-to-end runs on LongMemEval\_S; and (f)~the OpenAI models of Mem0 and
A-Mem and the model of Letta's agent, part of the compared systems: A-Mem,
and Mem0 on LongMemEval\_S, are
configured with \texttt{gpt-4o-mini}, Mem0 on LoCoMo with its library's
default model, but neither calls an LLM in the reported runs (Mem0 stores
with \texttt{infer=False}, A-Mem's LLM-driven note evolution is switched
off), Letta's agent registers a model that its adapter never calls (on
LongMemEval\_S the handle \texttt{letta/letta-free}), and the only OpenAI
calls inside the pipelines of Mem0 and A-Mem are those of
Mem0's embedder on LoCoMo (\S\ref{sec:judge-robustness}). Version~8 named a further role, a
reasoning backend for agent-level experiments; no result file contains
it (App.~\ref{app:changes-v9}, item K-74). Roles
(a)--(c) are non-methodological aids; roles (d)--(f) are
methodological components and are described in
Section~\ref{sec:experiments}. No LLM is used to generate, rewrite,
or filter ground-truth labels, and the human author verified all
scientific claims, citations, and statistical analyses.}

\end{enumerate}

\end{document}